\documentclass[journal]{IEEEtran}
\usepackage{cite}
\ifCLASSINFOpdf
   \usepackage[pdftex]{graphicx}
\else
\fi
\usepackage{amsmath}
  \usepackage[caption=false,font=normalsize,labelfont=sf,textfont=sf]{subfig}
\usepackage{multirow}
\usepackage{amsfonts}
\usepackage{algorithm}
\usepackage{stackengine}
\usepackage{algorithmicx}
\usepackage[noend]{algpseudocode}% http://ctan.org/pkg/algorithmicx
\usepackage{balance}
\usepackage[absolute]{textpos}
\begin{document}

\begin{textblock}{12}(2,0.4)
	\noindent Please cite as follows:
	K. Malialis, C. G. Panayiotou and M. M. Polycarpou, "Online Learning With Adaptive Rebalancing in Nonstationary Environments," in IEEE Transactions on Neural Networks and Learning Systems, doi: 10.1109/TNNLS.2020.3017863.
\end{textblock}

\title{Online Learning with Adaptive Rebalancing\\in Nonstationary Environments}

\author{
	Kleanthis~Malialis,
	Christos G.~Panayiotou,
	and~Marios M.~Polycarpou% <-this % stops a space

	\thanks{Authors are with the KIOS Research and Innovation Center of Excellence (K.M., C.G.P., M.M.P.) and the Department of Electrical and Computer Engineering (C.G.P., M.M.P.), University of Cyprus, Cyprus. Contact: \{malialis.kleanthis, christosp, mpolycar\}@ucy.ac.cy. ORCID: \{0000-0003-3432-7434, 0000-0002-6476-9025, 0000-0001-6495-9171\}}% <-this % stops a space
}

\maketitle

\begin{abstract}
An enormous and ever-growing volume of data is nowadays becoming available in a sequential fashion in various real-world applications. Learning in nonstationary environments constitutes a major challenge, and this problem becomes orders of magnitude more complex in the presence of class imbalance. We provide new insights into learning from nonstationary and imbalanced data in online learning, a largely unexplored area. We propose the novel Adaptive REBAlancing (AREBA) algorithm that selectively includes in the training set a subset of the majority and minority examples that appeared so far, while at its heart lies an adaptive mechanism to continually maintain the class balance between the selected examples. We compare AREBA with strong baselines and other state-of-the-art algorithms and perform extensive experimental work in scenarios with various class imbalance rates and different concept drift types on both synthetic and real-world data. AREBA significantly outperforms the rest with respect to both learning speed and learning quality. Our code is made publicly available to the scientific community.
\end{abstract}


% \fi
\IEEEpeerreviewmaketitle

\section{Introduction}\label{sec:intro}
\IEEEPARstart{E}{fficient} and effective analysis methods for the ever-increasing volume of sequential data in a wide range of applications is of paramount importance. In practical applications, data is evolving or drifting over time, i.e., data is drawn from nonstationary distributions. Various factors can trigger a nonstationarity effect or concept drift, for example, seasonality or periodicity effects, changes in users' habits, interests or preferences, and hardware or software faults \cite{ditzler2015learning}. Learning in nonstationary environments constitutes a major challenge. In such environments, a classifier with learning capabilities is of vital importance as it will provide an adaptive behaviour and help maintain optimal performance. The problem becomes significantly more complex if class imbalance co-exists with concept drift. In this case, class imbalance refers to sequential data that have skewed distributions and is a difficult problem as it causes a traditional learning algorithm to be ineffective because of its poor generalisation ability and its weak prediction power for the minority class examples\cite{he2008learning}.

Learning from nonstationary and imbalanced data has been studied separately but several key challenges remain open when the joint problem is considered. The majority of existing work, focuses on \textit{batch} (or \textit{chunk-by-chunk}) learning i.e. when examples arrive in batches (or chunks). In this paper we address the combined challenges of drift and imbalance in \textit{online} (or \textit{one-by-one}) learning, i.e., when a single example arrives at each step. The design of batch learning algorithms differs significantly from that of online learning and, therefore, the majority are typically unsuitable for online learning tasks \cite{wang2018systematic}. Addressing these key challenges can have a significant impact in various applications areas, e.g., in critical infrastructure systems, smart buildings, finance and banking, security and crime, healthcare and environmental sciences \cite{wang2018systematic, ditzler2013incremental, kyriakides2014intelligent, dal2015calibrating}.

The desired properties of an online classifier learning from nonstationary and imbalanced data are \cite{ditzler2013incremental, gama2014survey}: (1) \textit{Learning new knowledge}: The classifier should learn novel knowledge as new data is arriving. (2) \textit{Preserving previous knowledge}: Being able to preserve previous knowledge relies on the ability of the classifier to determine what previous knowledge is still relevant (and hence to preserve it) and what has now become irrelevant (and hence to discard or ``forget'' it). (3) \textit{High performance}: The classifier should obtain high performance on both the majority and minority classes. (4) \textit{Fast operation}: The classifier should operate in less than the example (or batch) arrival time. (5) \textit{Fixed storage}: The classifier should use no more than a fixed amount of memory for any storage; ideally, it should be capable of \textit{incremental} learning i.e. when learning occurs on a single instance (or batch) without considering (and hence storing) previous data. Balancing the trade-offs between the aforementioned properties is a challenging task.
%For instance, the first two properties are referred to as the \textit{stability - plasticity dilemma}, where stability refers to preserving previous knowledge and plasticity refers to learning new knowledge \cite{carpenter1991artmap}. Completely ``forgetting'' previous knowledge gives rise to what is known as \textit{catastrophic forgetting} \cite{french1991using}, therefore, a classifier should aim to balance the two. The two properties rely on the classifier's ability to detect drift, distinguish it from noise, and, if needed, adapt to changes while being robust with respect to noise \cite{gama2014survey}.

The contributions made are: \textit{(i)} We provide new insights into learning from nonstationary and imbalanced data, a largely unexplored area which focuses on the combined challenges of class imbalance and concept drift in online learning. \textit{(ii)} We propose the novel \textit{Adaptive REBAlancing (AREBA)} algorithm that maintains the aforementioned desired properties. \textit{AREBA} selectively includes in the training set a subset of the positive and negative examples that appeared so far, while at its heart lies an adaptive rebalancing mechanism to continually maintain class balance. \textit{(iii)} We compare \textit{AREBA} to strong baselines and state-of-the-art algorithms and perform an extensive experimental work in scenarios with various imbalance rates and different drift types on both synthetic and real-world data. \textit{AREBA} significantly outperforms the rest with respect to both learning speed and learning quality. \textit{(iv)} To our knowledge, this paper is one of the very few studies that examines online imbalance learning under each type of drift independently. For reproducibility of our results we make the datasets used and our code publicly available to the community\footnote{https://github.com/kmalialis/areba/}.

The organisation of this paper is as follows. Section~\ref{sec:background} provides the background material necessary to understand the contributions made. Section~\ref{sec:related_work} provides an in-depth review of related work. \textit{AREBA} is presented in Section~\ref{sec:proposed_method}. Our experimental setup is described in Section~\ref{sec:exp_setup}. An analysis of the proposed method is given in Section~\ref{sec:analysis}, followed by a comprehensive comparative study in Section~\ref{sec:study}. We conclude in Section~\ref{sec:discussion} where we discuss some important remarks, the pros and cons of \textit{AREBA} and pointers for future work.

\section{Background}\label{sec:background}
We consider a data generating process that provides at each time step $t$ a sequence of examples or instances $S^t = \{(x^t_i,y^t_i)\}^M_{i=1}$ from an unknown probability distribution $p^{t}(x,y)$, where $x^t \in \mathbb{R}^d$ is a $d$-dimensional input vector belonging to input space $X \subset \mathbb{R}^d$, $y^t \in Y$ is the class label where $Y=\{0,1\}$ and $M$ is the number of instances arriving at each step. The focus of this paper is on binary classification and, as a convention, the positive class represents the minority class. When the observed sequence $S^t$ consists only of a single instance (i.e. $M=1$), it is termed \textbf{online} (or \textit{one-by-one}) learning, otherwise it is termed \textbf{batch} (or \textit{chunk-by-chunk}) learning \cite{ditzler2015learning}. The design of batch learning algorithms differs significantly from that of online learning as they are designed to process batches of data, possibly by utilising an offline learning algorithm \cite{wang2018systematic}. Therefore, the majority of batch learning algorithms are typically not suitable for online learning tasks \cite{wang2018systematic}. This work focuses on online learning.

An online classifier receives a new example $x^t$ at time step $t$ and makes a prediction $\hat{y}^t$ based on a concept $h: X \to Y$ such that $\hat{y}^t = h(x^t)$. The classifier receives the true label $y^t$, its performance is evaluated using a loss function and is then trained, i.e., its parameters are updated accordingly based on the loss incurred. This process is repeated at each time step. Depending on the application, new examples do not necessarily arrive at regular and pre-defined intervals.

If data is sampled from a long, potentially infinite, sequence which is typically the case for big data applications, it is unrealistic to expect that all the previously observed data will always be available. If learning occurs on the most recent single instance only without taking into account previously observed data, it is termed \textbf{incremental} learning \cite{ditzler2015learning}. For online and incremental learning, the cost at time $t$ is calculated using the loss function $l$ as follows $J=l(y^t,\hat{y}^t)$.

This framework is suitable for \textit{human-in-the-loop} learning. As mentioned, the label becomes available as the next example arrives i.e. \textit{verification latency} does not exist. Algorithms of this framework, including \textit{AREBA}, are typically trained from user interaction by domain experts. Various and widely studied domains exist that fit into this framework, and hence, this assumption is satisfied e.g. in financial fraud \cite{dal2015calibrating}, a banker can provide every few minutes if a credit card transaction is fraudulent. For rain prediction \cite{ditzler2013incremental} an expert can provide every few hours if rain precipitation was observed. In healthcare applications \cite{fernandes2017transfer}, a doctor can provide the x-ray result as soon as it is completed. The framework may not be ideal in some cases (e.g. for real-time applications). Relaxing this assumption will be part of our future work, but we take a first step towards this direction and examine the robustness of all algorithms under conditions where this assumption is violated.

According to the Bayesian decision theory, a classification can be described by the prior probability $p(y)$ and the class conditional probability or likelihood $p(x|y)$ for all classes $y$ \cite{gama2014survey}. The classification decision is made according to the posterior probability, which for class $y$, is expressed as:
\begin{equation}
	p(y|x) = \frac{p(x|y)p(y)}{p(x)}
\end{equation}
\noindent where $p(x) = \sum_{y=\{0,1\}}{p(x|y)p(y)}$.

\textbf{Class imbalance} \cite{he2008learning} is a key challenge in learning and occurs when at least one data class is under-represented, thus constituting a minority class. For a binary classification problem, class imbalance at time $t$ occurs if:
\begin{equation}
	p^t(y = 0) >> p^t(y = 1)
\end{equation}

\noindent where class $0$ (negative) and $1$ (positive) represents the majority and minority class respectively.

\textbf{Concept drift} represents a change in the joint probability and the drift between time step $t_0$ and $t_1$ is defined as follows:
\begin{equation}
	\exists x \quad p^{t_0}(x,y) \neq p^{t_1}(x,y)
\end{equation}
Concept drift can occur in three forms: \textit{(i)} a change in prior probability $p(y)$ \textit{(ii)} a change in class-conditional probability or likelihood $p(x|y)$ and \textit{(iii)} a change in posterior probability $p(y|x)$. In real-world applications, the three forms can appear together. A change in posterior probability $p(y|x)$ which may or may not be due to a change in $p(x)$, is known as \textit{real} drift because the \textit{true} decision boundary is changed. A change in the distribution of the incoming data $p(x)$ without affecting $p(y|x)$ is known as \textit{virtual} drift because the \textit{true} decision boundary remains unchanged, however, the classifier's \textit{learnt} decision boundary may drift away from the \textit{true} one.

While other drift characteristics \cite{webb2016characterizing} are important, our focus is on the drift type. As mentioned, along with \cite{wang2018systematic}, this paper is one of the few studies that examines online imbalance learning under each type of concept drift independently.

\section{Related Work}\label{sec:related_work}
This section provides an in-depth review of related work and describes the state-of-the-art methods. For exhaustive surveys, the interested reader is directed towards these excellent papers: \cite{gama2014survey, ditzler2015learning} for drift methods, \cite{he2008learning} for imbalance methods, \cite{wang2018systematic} for methods that address both, and \cite{krawczyk2017ensemble} for online ensembles.

\subsection{Concept drift}
Concept drift algorithms are classified as \textit{memory-based}, \textit{change detection-based} and \textit{ensembling} \cite{gama2014survey}.

\subsubsection{\textbf{Memory-based}}\label{sec:drift_memory}
algorithms typically employ a sliding window approach to maintain a set of recent examples that a classifier is trained on; a representative algorithm of this category is \textit{FLORA} \cite{widmer1996learning}. A key challenge is to determine a priori the window size as a larger window is better suited for a gradual drift, while a smaller window is suitable for an abrupt drift. To address this, methods use an adaptive sliding window \cite{widmer1996learning} or multiple sliding windows \cite{lazarescu2004using}. This is also known as an \textit{abrupt forgetting} approach because the examples that fall outside the window are immediately dropped out of memory. Alternatively, \textit{gradual forgetting} approaches employ the full memory but the influence of older examples is deteriorating, e.g., using an exponential decay weighting strategy \cite{klinkenberg2004learning}.

\subsubsection{\textbf{Change detection-based}}\label{sec:drift_change_detection}
algorithms that employ explicit mechanisms to detect concept drift.  These methods are based on sequential analysis and control charts, e.g., \textit{Page-Hinkley (PH)} test \cite{page1954continuous}, \textit{cumulative sum (CUSUM)} \cite{page1954continuous}, \textit{just-in-time (JIT)} classifiers \cite{alippi2008justI, alippi2008justII} and on monitoring two distributions, e.g., \textit{adaptive windowing (ADWIN)} \cite{bifet2007learning}. These approaches are also known as \textit{active} detectors and are generally suitable for detecting abrupt concept drift but may fail to work well in prediction settings with gradual or recurring concept drift \cite{ditzler2015learning}, although recently \textit{JIT} classifiers have been extended to address recurring concept drift \cite{alippi2013just}.

\subsubsection{\textbf{Ensembling}}\label{sec:drift_ensembling}
an ensemble of classifiers can improve performance and provide the flexibility of injecting new data by adding classifiers or ``forgetting'' irrelevant data by removing or updating existing classifiers \cite{brzezinski2018ensemble}. It can be computationally costly; recall that one of the desired properties of a classifier is to be able to operate fast in less than the example arrival time. Popular methods are the \textit{streaming ensemble algorithm (SEA)} \cite{street2001streaming}, \textit{Learn++.NSE} \cite{elwell2011incremental}, \textit{diversity for dealing with drifts (DDD)} \cite{minku2011ddd} and \textit{online bagging (OB)} \cite{oza2005online}. Another method is the \textit{accuracy updated ensemble (AUE)} \cite{brzezinski2013reacting} which combines accuracy-based weighting mechanisms known from chunk-based ensembles with the incremental nature of Hoeffding Trees. Its follow-up work \textit{Online AUE (OAUE)} is provided in \cite{brzezinski2014combining}. The interested reader is directed towards \cite{krawczyk2017ensemble} for a survey on ensemble learning for data streams.

None of the aforementioned approaches consider class imbalance. Below we discuss how these are combined with class imbalance methods to address the joint problem.

\subsection{Class imbalance}
\textit{Cost-sensitive learning} and \textit{resampling} algorithms have recently shown particular success in this area \cite{wang2018systematic}.

\subsubsection{\textbf{Cost-sensitive learning}}\label{sec:imbalance_cs}
The \textit{cost-sensitive online gradient descent method (CSOGD)} uses this loss function:
\begin{equation}\label{eq:cs_cost}
J = ( I_{y^t=0} + I_{y^t=1} \frac{c_p}{c_n} ) ~ l(y^t, \hat{y}^t),
\end{equation}
\noindent where $I_{condition}$ is the indicator function that returns 1 if $condition$ is satisfied and 0 otherwise, $c_p, c_n \in [0,1]$ and $c_p + c_n = 1$ are the misclassification costs for positive and negative classes respectively \cite{wang2014cost}. The authors use the perceptron classifier and stochastic gradient descent, and apply the cost-sensitive modification to the hinge loss function, achieving excellent results. The downside of this method is that the costs need to be pre-defined, however, the extent of the class imbalance may not be known in advance. Moreover, in nonstationary environments, it cannot cope with imbalance changes (i.e. $p(y)$ drift) as the pre-defined costs remain static.

This issue can be resolved by introducing an adaptive cost strategy. One way to achieve an adaptive cost strategy is by using \textit{class imbalance detection (CID)} \cite{wang2013learning} to determine the imbalance rate in an online manner. The authors define a time-decayed class size metric, where for each class $k$, its size $s_k$ is updated at each time $t$ according to the following equation:
\begin{equation}
s^t_k = \theta s^{t-1}_k + I_{y^t = k} (1 - \theta)
\label{eq:cid}
\end{equation}
\noindent  where $0 < \theta < 1$ is a pre-defined time decay factor that gives less emphasis to older data. This metric can determine the imbalance rate at any given time, for instance, for a binary classification problem where the positive class is the minority ($s^t_p < s^t_n$), the imbalance rate at time $t$ is given by $s^t_n / s^t_p$.

Another method that uses an adaptive cost strategy with a perceptron-based classifier is \textit{RLSACP} \cite{ghazikhani2013recursive}. \textit{EONN} \cite{ghazikhani2013ensemble} uses an ensemble of cost-sensitive online neural networks to cope with drift and imbalance. As with \textit{CSOGD}, the costs are pre-defined, thus limiting its adaptability to evolving data. 

\subsubsection{\textbf{Resampling}}\label{sec:imbalance_resampling}
Traditionally, in \textit{offline} learning, resampling techniques alter the training set to deal with the skewed data distribution, specifically, oversampling techniques ``grow'' the minority class while undersampling techniques ``shrink'' the majority class. The simplest and most popular technique is random oversampling (or undersampling) where data examples are randomly added (or removed) respectively \cite{zhou2006training}. More sophisticated techniques exist, for example, the use of \textit{Tomek links} \cite{tomek1976two} discards borderline examples while the \textit{SMOTE} \cite{chawla2002smote} algorithm generates new minority class examples based on the similarities to the original ones. Recently, Generative Adversarial Networks (GANs) have been used to approximate the distribution and generate data for the minority class \cite{douzas2018effective}.

Resampling has been demonstrated to be a powerful technique for addressing online imbalance learning problems as well. \textit{Uncorrelated bagging (UCB)} \cite{gao2008classifying} is an ensembling technique that is trained on all the minority examples observed so far, plus a subset of the most recent majority examples. This technique has two drawbacks; it assumes that the distribution of the minority class is stationary and it does not handle the accumulated minority class examples for lifelong learning. \textit{SERA} \cite{chen2009sera} and \textit{REA} \cite{chen2011towards} are based on \textit{UCB} and use more intelligent oversampling techniques. Other notable examples are the \textit{Learn++CDS} and \textit{Learn++NIE} \cite{ditzler2013incremental} methods where both of them use the aforementioned \textit{Learn++NSE} method (Section~\ref{sec:drift_ensembling}) to address concept drift. The former combines the \textit{SMOTE} algorithm to address class imbalance while the latter combines a variation of bagging. Despite their effectiveness, all these techniques are only suitable for \textit{batch} learning and not for \textit{online} learning, which is the focus of this paper.

\textit{ESOS-ELM} \cite{mirza2015ensemble} is an ensemble of online sequential extreme learning machines that are trained on balanced subsets of the data stream. It relies, however, on the assumption that drift does not affect the minority class. \textit{Oversampling-based online bagging (OOB)} \cite{wang2015resampling} is an online ensembling method that extends the \textit{OB} method (Section~\ref{sec:drift_ensembling}) which addresses concept drift. It works by adjusting the learning bias from the majority to the minority class adaptively through resampling by utilising the \textit{CID} method (Section~\ref{sec:imbalance_cs}) and its time-decayed class size metric defined in Eq~(\ref{eq:cid}). Its basic idea is as follows. \textit{OOB} updates each classifier of the ensemble $K$ times. If a minority example arrives the value of $K$ increases, otherwise it decreases. The effectiveness of \textit{OOB} has been demonstrated using two types of classifiers, namely, Hoeffiding trees and neural networks. The ensemble size varies for each study e.g. in \cite{wang2015resampling} 50 trees and 50 neural networks were used while in \cite{wang2018systematic} 15 neural networks were used. The approach can be computationally costly and may hinder online learning in high-speed sequential applications for two reasons. The first reason is due to the multiple classifiers (ensembling) and the second is because each classifier gets updated multiple times per time step. Analogous to \textit{OOB}, its authors also introduce \cite{wang2015resampling} the \textit{Undersampling-based online bagging (UOB)} algorithm.

\subsection{Open Challenges}\label{sec:challenges}
Several key challenges still remain open when the joint problem of imbalance and drift is considered. The authors in \cite{krawczyk2017ensemble} state that ``working with class-imbalanced and evolving streams is still in early stages'', while this study \cite{wang2018systematic} ``reveals research gaps in the field of online imbalance learning with concept drift''. Specifically, many existing methods are capable of addressing only a single problem, either imbalance or drift, but not the joint problem. In other methods that address the joint problem, weaknesses are revealed under conditions where one or both the problems become very challenging. For instance, we will demonstrate these weaknesses under conditions where class imbalance is extreme (e.g. $0.1\%$). This paper introduces the concept of maintaining \textit{separate} and \textit{balanced} queues for each class, and has a \textit{dual-nature} as it merges ideas from memory-based and resampling algorithms.

Also, besides its type, drift can be classified by its severity, speed, predictability, frequency and recurrence \cite{minku2010impact}. A more recent study characterises drift by subject, frequency, transition, re-occurrence and magnitude \cite{webb2016characterizing}. Therefore, in practice, it is very difficult to characterise concept drift. Our focus is on \textit{learning the concept drift} without its explicit characterisation and detection. As discussed in Section~\ref{sec:drift_change_detection}, explicit or active drift detectors can perform well under specific drift characteristics. However, no detector can universally perform satisfactory under any combination of drift characteristics \cite{barros2018large}. In fact, detailed characteristics of drift have not been consistently investigated in the literature \cite{krawczyk2017ensemble}. Our proposed approach learns the concept drift without its explicit characterisation and detection, and adapts the classifier continuously.

\section{Proposed Method}\label{sec:proposed_method}
We now introduce the \textit{Adaptive REBAalancing (AREBA)} algorithm. Its central idea is to selectively include in the training set a subset of the positive and negative examples that appeared so far. At its heart lies an adaptive rebalancing mechanism that dynamically modifies the queue sizes to maintain class balance between the selected examples. \textit{AREBA} extends our recently introduced algorithm \textit{queue-based resampling} (\textit{QBR}) \cite{malialis2018queue}.

\begin{figure}[t!]
	\centering
	\includegraphics[scale=0.42]{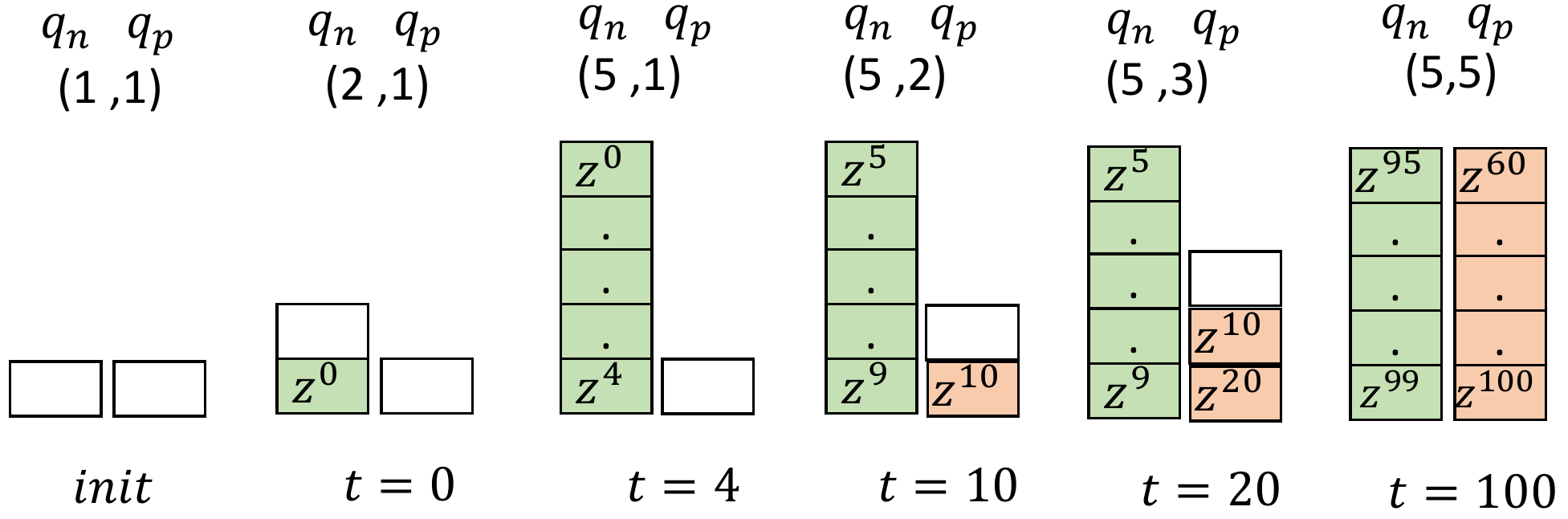}
	\caption{Example of \textit{QBR} for $B=10$. Negative examples are shown in green, positive ones in light red and the minority class is the positive class. It takes 100 time steps for the queues to become balanced.}
	\label{fig:qbr}
\end{figure}

\begin{figure}[t]
	\centering
	\includegraphics[scale=0.37]{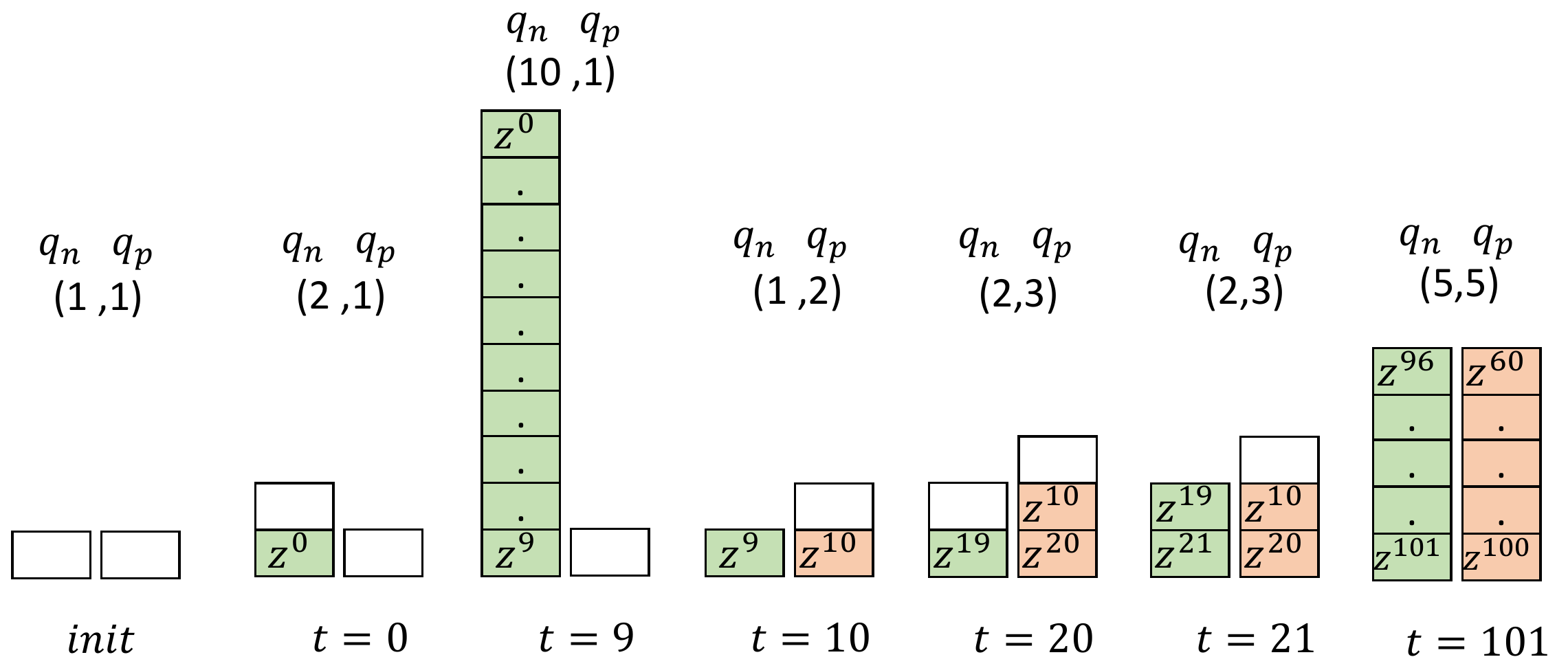}
	\caption{Example of  \textit{AREBA} for $B=10$. Negative examples are shown in green, positive ones in light red and the minority class is the positive class. Rebalancing enforces the queues to remain balanced throughout the time.}
	\label{fig:areba}
\end{figure}

\begin{algorithm}[t]
	\caption{Queue-based Resampling (QBR)}
	\label{alg:qbr}
	\begin{algorithmic}[1]
		
		\State\textbf{Input:}
		\State $f$: classifier
		\State $B$: total storage size ($B \geq 2$)
		
		\State \textbf{Initialisation:}
		\State queues $q^0_p, q^0_n = \{\}$
		\State queue capacities $q^0_p.cap = q^0_n.cap = 1$
		
		\For{each time step $t$}
		
		\State{receive example $x^t \in \mathbb{R}^d$}
		\State{predict class $\hat{y}^t \in \{0,1\}$}
		\State{receive true label $y^t \in \{0,1\}$}
		
		\If{$y^t == 1$}
		\State{$q^t_p = q^{t-1}_p.append\big((x^t,y^t)\big)$}
		\Else
		\State{$q^t_n = q^{t-1}_n.append\big((x^t,y^t)\big)$}
		\EndIf
		
		\If{$q^t_p.is\_full()$}
		\If{$q^t_p.cap < \frac{B}{2}$}
		\State{$q^t_p.cap = q^t_p.cap + 1$}\Comment{increase capacity}
		\ElsIf{$q^t_p.cap == \frac{B}{2}$}
		\State{pass}\Comment{queue no longer grows}
		\EndIf
		\EndIf		
		
		\If{$q^t_n.is\_full()$}
		\If{$q^t_n.cap < \frac{B}{2}$}
		\State{$q^t_n.cap = q^t_n.cap + 1$}
		\ElsIf{$q^t_n.cap == \frac{B}{2}$}
		\State{pass}
		\EndIf
		\EndIf		
		
		\State{prepare the training set $q^t = q^t_p \cup q^t_n$}
		\State{calculate cost $J$ on $q^t$ using Eq~(\ref{eq:cost}})
		\State{update classifier once $f.train()$}
		
		\EndFor
		
	\end{algorithmic}
\end{algorithm}

\begin{algorithm}[t!]
	\caption{Adaptive Rebalancing (AREBA)}
	\label{alg:areba}
	\begin{algorithmic}[1]
		
		\State{\textbf{Input:}\\
			$f$: classifier\\
			$B$: total storage size ($B \geq 2$)\\
			$\theta$: decay factor for class size metrics $s^t_p, s^t_n$
		}
		
		\State{\textbf{Initialisation:}\\
			class sizes $s^0_p, s^0_n = 0$\\
			queues $q^0_p, q^0_n = \{\}$\\
			queue capacities $q^0_p.cap = q^0_n.cap = 1$
		}
		
		\For{each time step $t$}
		
		\State{receive example $x^t \in \mathbb{R}^d$}
		\State{predict class $\hat{y}^t \in \{0,1\}$}
		\State{receive true label $y^t \in \{0,1\}$}
		\State{update positive class size $s^t_p = \theta s^{t-1}_p + I_{y^t = 1} (1 - \theta)$}
		\State{update negative class size $s^t_n = \theta s^{t-1}_n + I_{y^t = 0} (1 - \theta)$}
		\If{$y^t == 1$}
		\State{$q^t_p = q^{t-1}_p.append\big((x^t,y^t)\big)$}
		\Else
		\State{$q^t_n = q^{t-1}_n.append\big((x^t,y^t)\big)$}
		\EndIf
		
		\If{$q^t_p.is\_empty()$}
		\If{$q^t_n.cap < B$}
		\State{$q^t_n.cap = q^t_n.cap + 1$}\Comment{increase capacity}
		\ElsIf{$q^t_n.cap == B$}
		\State pass\Comment{queue no longer grows}
		\EndIf
		
		\ElsIf{$q^t_n.is\_empty()$}
		\If{$q^t_p.cap < B$}
		\State{$q^t_p.cap = q^t_p.cap + 1$}
		\ElsIf{$q^t_p.cap == B$}
		\State pass
		\EndIf
		
		\Else
		
		\If{$s^t_n > s^t_p$}\Comment{if positive class is minority}
		\If{$q^t_p.is\_full()$}
		\If{$q^t_p.cap < \frac{B}{2}$}
		\State{$q^t_p.cap = q^t_p.cap + 1$}
		\State{$q^t_n.cap = q^t_p.cap - 1$}
		\ElsIf{$q^t_p.cap == \frac{B}{2}$}
		\If{$q^t_n.cap \neq q^t_p.cap$}
		\State{$q^t_n.cap = q^t_p.cap$}
		\EndIf
		\EndIf
		\EndIf
		\EndIf
		
		\If{$s^t_n \leq s^t_p$}\Comment{if negative class is minority}
		\If{$q^t_n.is\_full()$}
		\If{$q^t_n.cap < \frac{B}{2}$}
		\State{$q^t_n.cap = q^t_n.cap + 1$}
		\State{$q^t_p.cap = q^t_n.cap - 1$}
		\ElsIf{$q^t_n.cap == \frac{B}{2}$}
		\If{$q^t_p.cap \neq q^t_n.cap$}
		\State{$q^t_p.cap = q^t_n.cap$}
		\EndIf
		\EndIf
		\EndIf
		\EndIf
		
		\EndIf
		
		\State{prepare the training set $q^t = q^t_p \cup q^t_n$}
		\State{calculate cost $J$ on $q^t$ using Eq~(\ref{eq:cost}})
		\State{update classifier once $f.train()$}
		
		\EndFor
		
	\end{algorithmic}
\end{algorithm}

\subsection{Queue-based Resampling (QBR)}
The memory size $B \in 2\mathbb{Z}^+$ (i.e. $B \geq 2$ and even) determines how many previously observed examples can be stored. The selection of the examples is achieved by maintaining at any given time $t$ two separate windows of capacity (maximum length) $\frac{B}{2}$. The windows are implemented using queues i.e. $q^t_n$ and $q^t_p$ contain the negative and positive examples respectively:
\begin{equation}
\begin{aligned}
	q^t_n = \{(x_i,y_i)\}^{|q^t_n|}_{i=1} \\
	q^t_p = \{(x_i,y_i)\}^{|q^t_p|}_{i=1}
\end{aligned}
\end{equation}
\noindent where $|q^t_n|, |q^t_p| \in [0,\frac{B}{2}]$ are the current lengths of the queues. Let $z_i=(x_i,y_i)$, for any two $z_i, z_j \in q^t_n$ (or $q^t_p$) such that $j > i$, $z_j$ arrived more recently in time.

An example showing how \textit{QBR} works when $B=10$ for 100 time steps is shown in Figure~\ref{fig:qbr}. Negative examples are shown in green, positive ones in light red and the minority class is the positive class. The class imbalance is set to $CI=10\%$ i.e. $p(y=1)=0.1$ and for the sake of illustration, positive instances arrive at times multiple of ten ($t=10, 20, ..., 100$). Initially, both queues are empty (shown as empty boxes) but their capacity is set to one (shown in the parenthesis). At $t=0$ a negative example ($z^0$) arrives which is appended to the negative queue and the queue's capacity is incremented by one. At $t=4$ the negative queue is full and has reached the full capacity $\frac{B}{2}$. At $t=10$ the first positive example ($z^{10}$) arrives which is appended to the positive queue and the queue's capacity is incremented. At $t=100$ the positive queue is full and has reached the full capacity $\frac{B}{2}$. Note that at this time both queues contain the most recent $\frac{B}{2}$ examples i.e. $z^{95}, .., z^{99}$ and $z^{60}, .., z^{100}$ for the negative and positive queue respectively.

The union of the two queues is then taken  to form the new training set. The cost function is given by:
\begin{equation}\label{eq:cost}
J = \frac{1}{|q^t|} \sum_{i=1}^{|q^t|} l(y_i, h(x_i)),
\end{equation}
\noindent where $q^t = q^t_p \cup q^t_n$ and $|q^t| \in [1,B]$. At each step the classifier is updated once based on the cost $J$ incurred. \textit{QBR}'s pseudocode is shown in Algorithm~\ref{alg:qbr}. In Lines 5-6 the queues are initially empty with a capacity of one each. In Lines 11-14 a new example is appended in its relevant queue based on its true label. The \textit{append} function behaves exactly as in Fig.~\ref{fig:qbr} i.e. it inserts the most recent example in a queue, while discarding the oldest one. Consider the case of $t=10$ where the most recent example in the negative queue is $z^9$. When the example $z^9$ arrived at $t=9$, the example $z^4$ was discarded from the queue. In Lines 15-24 the capacity of the relevant queue is incremented. In Lines 25-27 the training set is prepared, the cost is calculated and the classifier is updated once.

Of particular importance, is the observation that the original class imbalance problem still persists in the queues for a sustained period of time. Let's revisit time $t=10$ in Figure~\ref{fig:qbr}. While the $q^t_n$ is full, the $q^t_p$ only contains a single example. Recall that to train the classifier we first take the union of the queues and then calculate the cost. Class imbalance is reduced as positive examples arrive and the problem eventually disappears at $t=100$ when the queues become balanced.

\subsection{Adaptive Rebalancing (AREBA)}\label{sec:areba}
Adaptive rebalancing introduces a novel element that dynamically modifies the queue lengths in order to constantly maintain balance between the queues. Without this element, the initial class imbalance problem would still persist in the queue-based system as discussed in the previous section.

We describe in Figure~\ref{fig:areba} how \textit{AREBA} works through the same example as before. Negative examples are shown in green, positive ones in light red and the minority class is the positive class. The class imbalance is set to $CI=10\%$ i.e. $p(y=1)=0.1$ and for the sake of illustration, positive instances arrive at times multiple of ten ($t=10, 20, ..., 100$). Initially, both queues are empty (shown as empty boxes) but their capacity is set to one (shown in the parenthesis). At $t=0$ a negative example ($z^0$) arrives which is appended to the negative queue and the queue's capacity is incremented by one. Contrary to \textit{QBR}, each queue is allowed to have a maximum capacity of $B$ (rather than $\frac{B}{2}$). Since no positive examples are observed in the beginning, at $t=9$ the $q_n$ is full and has reached the maximum capacity $B$.

The first positive example ($z^{10}$) arrives at $t=10$ and is appended to $q_p$. Rebalancing is now initiated and the capacity of $q_n$ and $q_p$ becomes 1 and 2 respectively. The $q_n$ only contains its most recent example ($z^9$), hence, the queues are now balanced. The queues remain balanced until the second positive example ($z^{20}$) arrives at $t=20$. The $q_p$ contains now the two most recent positive examples ($z^{20}$, $z^{10}$) while $q_n$ contains the most recent negative example ($z^{19}$). At $t=20$ the capacity of $q_n$ and $q_p$ becomes 2 and 3 respectively. At $t=21$ another negative example ($z^{21}$) arrives which is appended to the relevant queue, thus the queues are again balanced. At $t=101$ each queue is full and has a capacity of $\frac{B}{2}$.

\textit{AREBA} proposes an adaptive mechanism that dynamically alters the queue sizes to maintain balance between the examples contained in the queues. To achieve this, it is necessary to be able to decide in an online fashion which class is the minority and which the majority. \textit{AREBA} adopts the \textit{CID} method's time-decayed class size metrics defined in Eq~(\ref{eq:cid}).

\textit{AREBA} is shown in Algorithm~\ref{alg:areba}. In Lines 7-8 the capacity of each queue is initialised to 1. In Lines 13 and 14 the class size metrics are updated. The new example is appended in its relevant queue (Lines 15-18). We then check if one of the queues is empty (Lines 19-28). For instance, if the positive queue is empty, we increase the capacity of the negative queue by 1; this corresponds to the cases $t=0$ to $t=9$ in Figure~\ref{fig:areba}. Since the positive class is the minority class, the rest of the cases in Figure~\ref{fig:areba} are captured by Lines 30-37. Line 31 checks if the positive queue is full (in our illustration this occurs at $t=10, 20, ..., 100$) and then we adapt the capacities accordingly (Lines 32-37) depending on whether the capacity of the positive queue has reached $\frac{B}{2}$. Similarly, Lines 38-45 are applicable when the negative class is the majority class.

In summary, \textit{AREBA} introduces the concept of maintaining \textit{separate} and \textit{balanced} queues for each class and its effectiveness is attributed to its \textit{dual-nature} as it combines ideas from memory-based and resampling methods. These and other important remarks are discussed in detail in Section~\ref{sec:discussion}.

\section{Experimental Setup}\label{sec:exp_setup}
This section describes the synthetic and real-world datasets used along with any data pre-processing steps performed. It also describes the baseline and state-of-the-art methods used along with their selected parameters. It further describes the performance metrics and the evaluation method used.

\subsection{Datasets}\label{sec:datasets}

\subsubsection{Synthetic datasets} They provide the flexibility to control the imbalance level, when to introduce drift, and control the drift type. We experiment with imbalance of 10\% (mild), 1\% (severe) and 0.1\% (extreme) and examine each drift type individually to inspect the advantages and limitations of all the compared methods. The three synthetic datasets used are described below where we will use their balanced and imbalanced versions, with and without drift.

\textbf{Circle} \cite{gama2004learning}: It has the two features $x_1, x_2 \in [0, 1]$. The classification function is a circle $(x_1 - x_{1c})^2 + (x_2 - x_{2c})^2 = r_c^2$ where $(x_{1c}, x_{2c})$ is its centre and $r_c$ its radius. The circle with $(x_{1c}, x_{2c}) = (0.4, 0.5)$ and $r_c = 0.2$ is created. Instances inside the circle are classified as positive, otherwise as negative.

\textbf{Sine} \cite{gama2004learning}: It consists of the features $x_1 \in [0,2\pi]$ and $x_2 \in [-1,1]$. The classification function is $sin(x_1)$. Instances below the curve are classified as positive, otherwise as negative. Feature rescaling has been performed so that $x_1, x_2 \in [0, 1]$.

\textbf{Sea} \cite{street2001streaming}: It has the features $x_1, x_2 \in [0, 10]$. Instances that satisfy $x_1 + x_2 \leq 7$ are classified as positive, otherwise as negative. Rescaling has been performed so that $x_1, x_2 \in [0, 1]$.

\subsubsection{Real-world datasets} They are typically more complex than synthetic ones and have a large number of noisy features, but the true nature of concept drift may be unknown. The five datasets used in this paper cover various application domains, specifically, healthcare, security and crime, finance and banking, image classification and environmental sciences.

\textbf{Cervical Cancer} \cite{fernandes2017transfer}: The dataset was collected at the Hospital Universitario de Caracas in Caracas, Venezuela and contains demographic information, habits and historical medical records of 858 patients. The task is to predict the outcome of a biopsy with respect to cervical cancer and each class label was decided by a team of six experts. The dataset is highly imbalanced as 55 out of the 858 cases (6.4\%) correspond to cases of cervical cancer. The number of features is 45. The exact nature of drift is unknown but it is expected to occur due to the fact that cancer cells gain genetic variation over time and also due to changes in patients' habits.

\textbf{Fraud} \cite{dal2015calibrating}: The dataset contains transactions made by credit cards in September 2013 by European cardholders. This dataset presents transactions that occurred in two days, where we have 492 frauds out of 284,807 transactions. The dataset contains 30 features and is severely imbalanced as the positive class (frauds) accounts for 0.172\% of all transactions. The exact nature of concept drift is unknown but it is expected to occur due to the adaptive nature of adversarial actions.

\textbf{Credit Score} \cite{credit_score}: The dataset contains demographic and financial / banking information. The task is to predict the credit score (good, bad) of a customer. The dataset contains 1000 entries, out of which 300 correspond to a bad credit i.e. class imbalance is 30\%. The number of features is 24. The exact nature of concept drift is unknown but may occur due to customer attempts to fake or improve their credit score.

\textbf{MNIST} \cite{lecun1998gradient}: It is a database of handwritten digits (``0''-``9'') where each image is 28x28. Contrary to the rest of the datasets where their data type is numeric, MNIST is commonly used for training image processing systems for visual tasks. The dataset is diverse as the digits were written by $\sim$250 adult and student writers. We selected digit ``7'' to be the majority class with 6000 instances and digit ``2'' to be the minority class with 60 instances, therefore, imbalance is close to $1\%$. Drift can occur as new handwriting styles appear during learning.

\textbf{Forest Cover Type} \cite{blackard1999comparative}: The datasets consists of cartographic information obtained from the US Forest Service. The task is to predict the forest cover type for given 30x30 meter cells from the Roosevelt National Forest in Colorado. We have selected the cover type ``1'' to be the majority class with 200000 instances and type ``4'' to be the minority class with 2000 instances, therefore, imbalance is close to $1\%$. This dataset has been used in many concept drift studies e.g. \cite{brzezinski2013reacting}.

\subsection{Compared methods}\label{sec:methods}
All compared methods share the same base classifier which is a fully-connected neural network of one 8-neuron hidden layer, except for MNIST which consists of two 512-neuron hidden layers to deal with its complex data type (i.e. images). The base classifier is configured as follows: \textit{He Normal} \cite{he2015delving} weight initialisation, the \textit{Adam} \cite{kingma2014adam} optimisation algorithm, \textit{LeakyReLU} \cite{maas2013rectifier} as the activation function of the hidden neurons, sigmoid activation for the output neuron, and the binary cross-entropy loss function. The learning rate is $0.01$ for the synthetic, \textit{Credit Score} and \textit{Forest Cover Type} datasets, $0.1$ for \textit{Cervical Cancer} and $0.0001$ for \textit{Fraud}. For \textit{MNIST} the learning rate is $0.001$ and $L2$ regularisation is set to $0.01$.
%The compared methods are:

\textbf{Baseline.} A baseline algorithm where no mechanisms to address class imbalance or concept drift exist. This baseline method is an \textit{online} and \textit{incremental} learning algorithm.

\textbf{Sliding window.} A \textit{memory-based} method that uses a single sliding window to address drift, but no mechanism to address imbalance exists. This is in contrast to \textit{QBR} which utilises one sliding window per class. The window size is set to $W=100$. It is an \textit{online} but not \textit{incremental} learning algorithm as it requires access to $W-1$ previously observed data examples.

\textbf{Adaptive\_CS.} A state-of-the-art adaptive cost-sensitive learning method. It uses the \textit{CSOGD} cost function defined in Eq~(\ref{eq:cs_cost}) initialised to $c = \frac{c_p}{c_n} = \frac{0.95}{0.05} = 19$ as suggested by its authors. These costs are adapted according to the \textit{CID} approach, where its time-decayed class size metrics are defined in Eq~(\ref{eq:cid}). The time-decayed factor is set to $\theta = 0.99$. To overcome stability issues in performance we set an upper bound to the ratio i.e. $1 \leq c \leq 50$ and we update the costs every $250$ steps. This method is both \textit{online} and \textit{incremental}.

\textbf{OOB.} A state-of-the-art online resampling algorithm (Section~\ref{sec:imbalance_resampling}). We will consider \textit{OOB} with 20 classifiers and the special case \textit{OOB\_single} where only a single classifier is used. The time-decayed factor is set to $\theta = 0.99$. \textit{OOB} is an \textit{online} and \textit{incremental} learning algorithm but unlike other methods, it performs multiple updates of the classifier at each time step.

\textbf{AREBA.} The proposed method whose pseudocode is provided in Algorithm~\ref{alg:areba}. The time-decayed factor is set to $\theta = 0.99$. \textit{AREBA} is an \textit{online} learning algorithm. When $B=2$, the method (referred to as \textit{AREBA\_2}) becomes a \textit{near-incremental} learning algorithm as it requires access only to a single old example. In our study, we will examine various values of memory size, which we will refer to as \textit{AREBA\_B}.

We discuss now some aspects concerning the computational cost of algorithms. For all methods, \textit{one-pass} learning is used i.e. the base classifier is updated once (\textit{\#epochs=1}) at every step, except for \textit{OOB} which is updated $K$ times (\textit{\#epochs=K}). A single batch update is performed within an epoch; this is to allow fast learning in high-speed applications. For \textit{AREBA}, the $q^t$ is the batch while for \textit{Sliding}, the window $W$ is the batch. For the rest, the batch size is 1 as they are incremental algorithms. The same classifier gets updated throughout the duration of an experiment. Even in the presence of drift, we never reset the classifier or introduce any new classifier(s).

\subsection{Performance metrics}
Traditionally, classifiers are evaluated using the overall accuracy metric. When class imbalance exists, accuracy becomes problematic as it is biased towards the majority class. This occurs because any metric that uses values from both columns of the confusion matrix will be inherently sensitive to class imbalance \cite{he2008learning}. Therefore, it is necessary to adopt a performance metric which is not sensitive to class imbalance.

The geometric mean $G\textnormal{-}mean$ \cite{he2008learning} is such a suitable metric that evaluates the degree of inductive bias in terms of a ratio of positive accuracy $Acc^+$ (or recall) and negative accuracy $Acc^-$ (or specificity), as defined below:
 \cite{kubat1997learning}.
\begin{equation}\label{eq:gmean}
\begin{split}
%	G\textnormal{-}mean = & \sqrt{Acc^+ * Acc^-} = \sqrt{\frac{TP}{TP + FN} * \frac{TN}{TN + FP}}
	G\textnormal{-}mean = & \sqrt{Acc^+ * Acc^-} = \sqrt{(TP/P) * (TN/N)}
\end{split}
\end{equation}
\noindent where $TP, P, TN$ and $N$ is the number of true positives, total positives, true negatives and total negatives respectively. $G\textnormal{-}mean$ has some desirable properties as it is high when both $Acc^+$ and $Acc^-$ are high and when their difference is small.

\subsection{Evaluation method}\label{sec:evaluation}
To evaluate, compare and assess predictive sequential learning algorithms we adopt the \textit{prequential error with fading factors} method. It has been proven that for learning algorithms in stationary data this method converges to the Bayes error \cite{gama2013evaluating}. This method does not require a holdout set and the predictive model is always tested on unseen data. In our experimental study the fading factor is set to $\theta = 0.99$.

In all simulations we plot the prequential metric (e.g. \textit{G-mean}) in every step averaged over 50 repetitions, including the error bars displaying the standard error around the mean. Additionally, in all experiments we test for statistical significance using a \textit{one-way repeated measures ANOVA} and then using \textit{post-hoc multiple comparisons tests} with \textit{Fisher's least significant difference} correction procedure to show which of the compared method is significantly different from the others.

\section{Empirical Analysis of AREBA}\label{sec:analysis}
This section presents a two-fold analysis of the proposed method. In particular, we examine and discuss the roles of the adaptive rebalancing mechanism and the memory size.

\begin{figure}[t]
	\centering
	
	\subfloat[$CI=10\%$]{\includegraphics[scale=0.10]{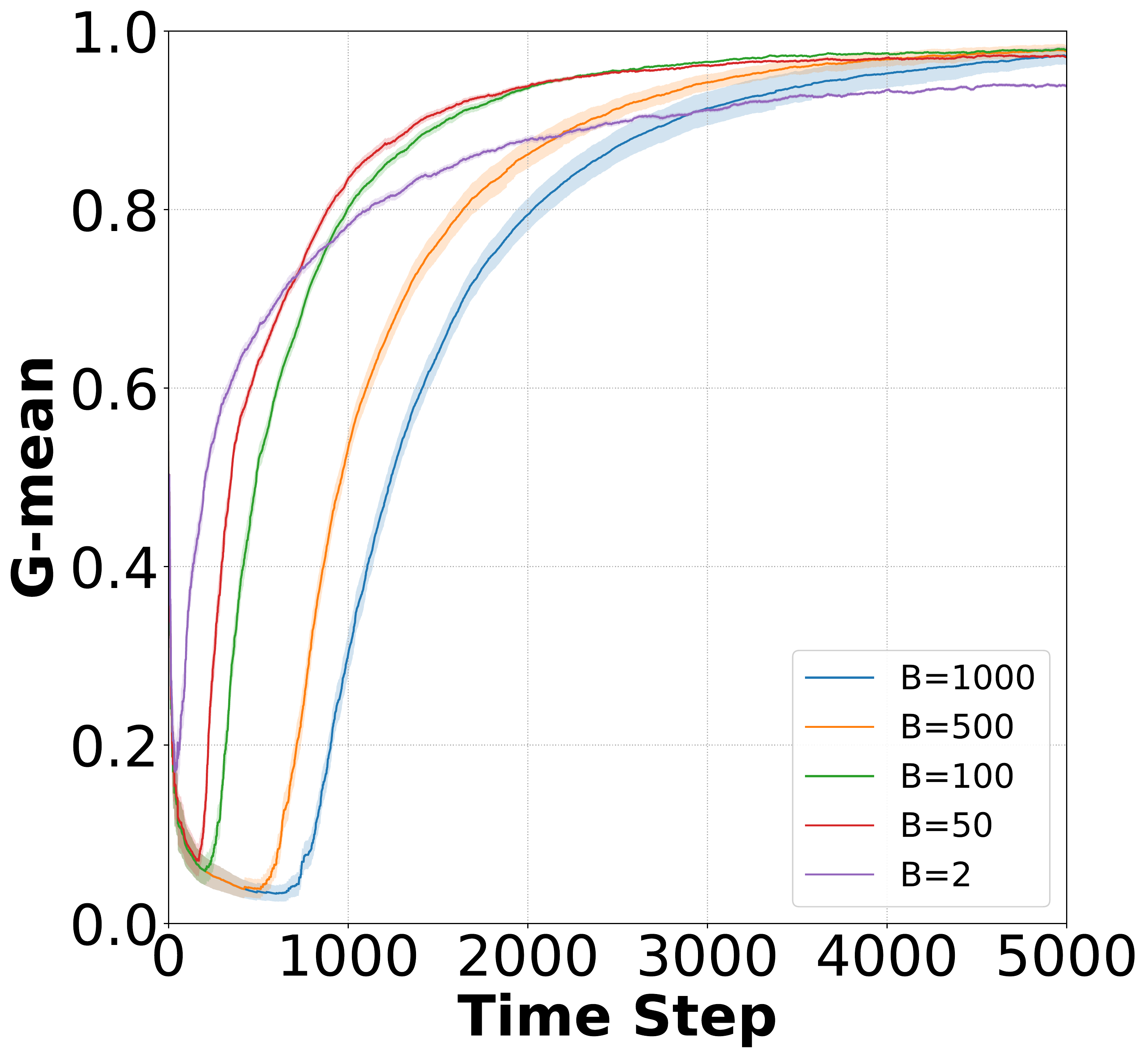}%
		\label{fig:analysis_circle_pp10_qbr}}
	\subfloat[$CI=1\%$]{\includegraphics[scale=0.10]{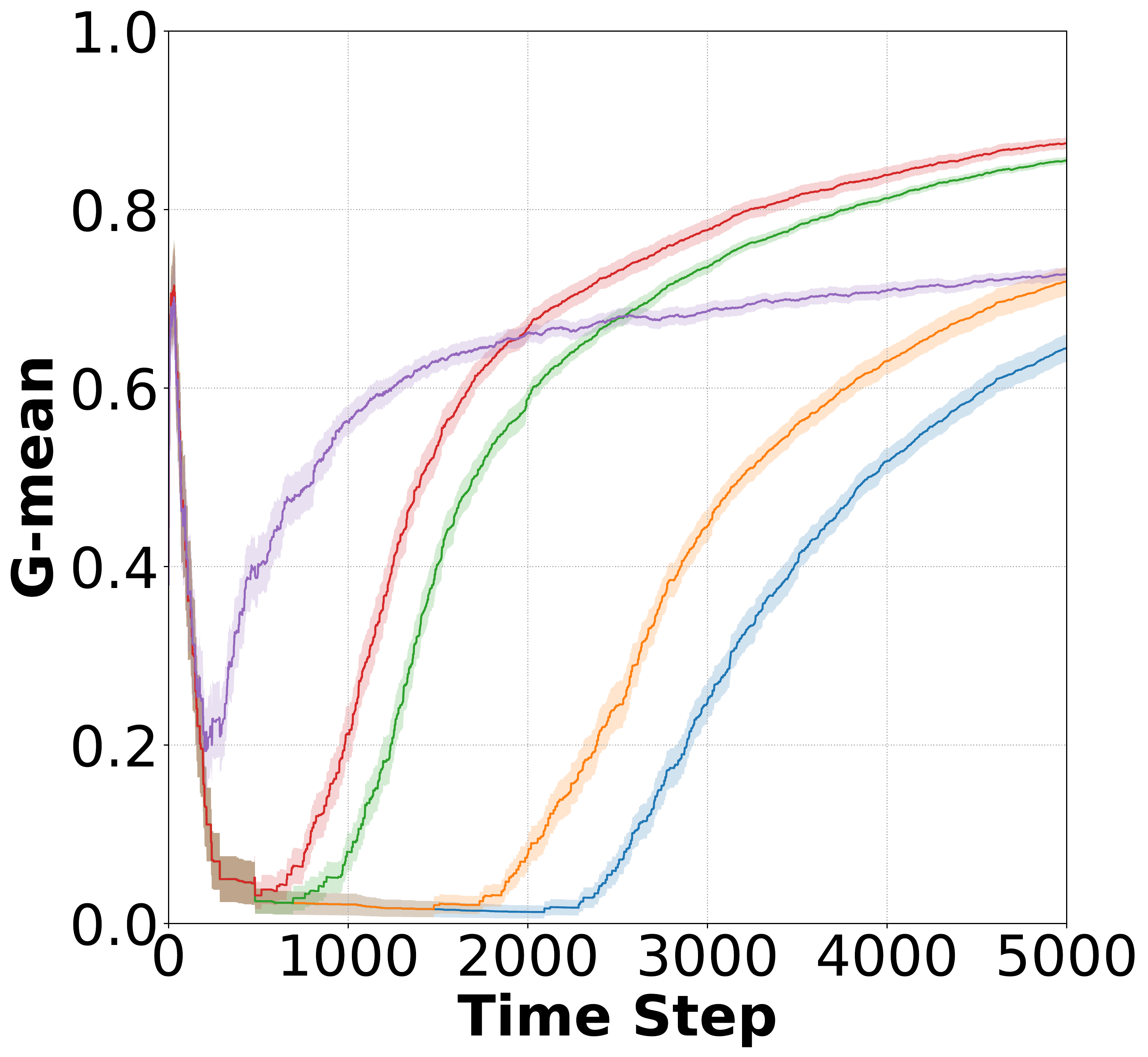}%
		\label{fig:analysis_circle_pp1_qbr}}
	\subfloat[$CI=0.1\%$]{\includegraphics[scale=0.10]{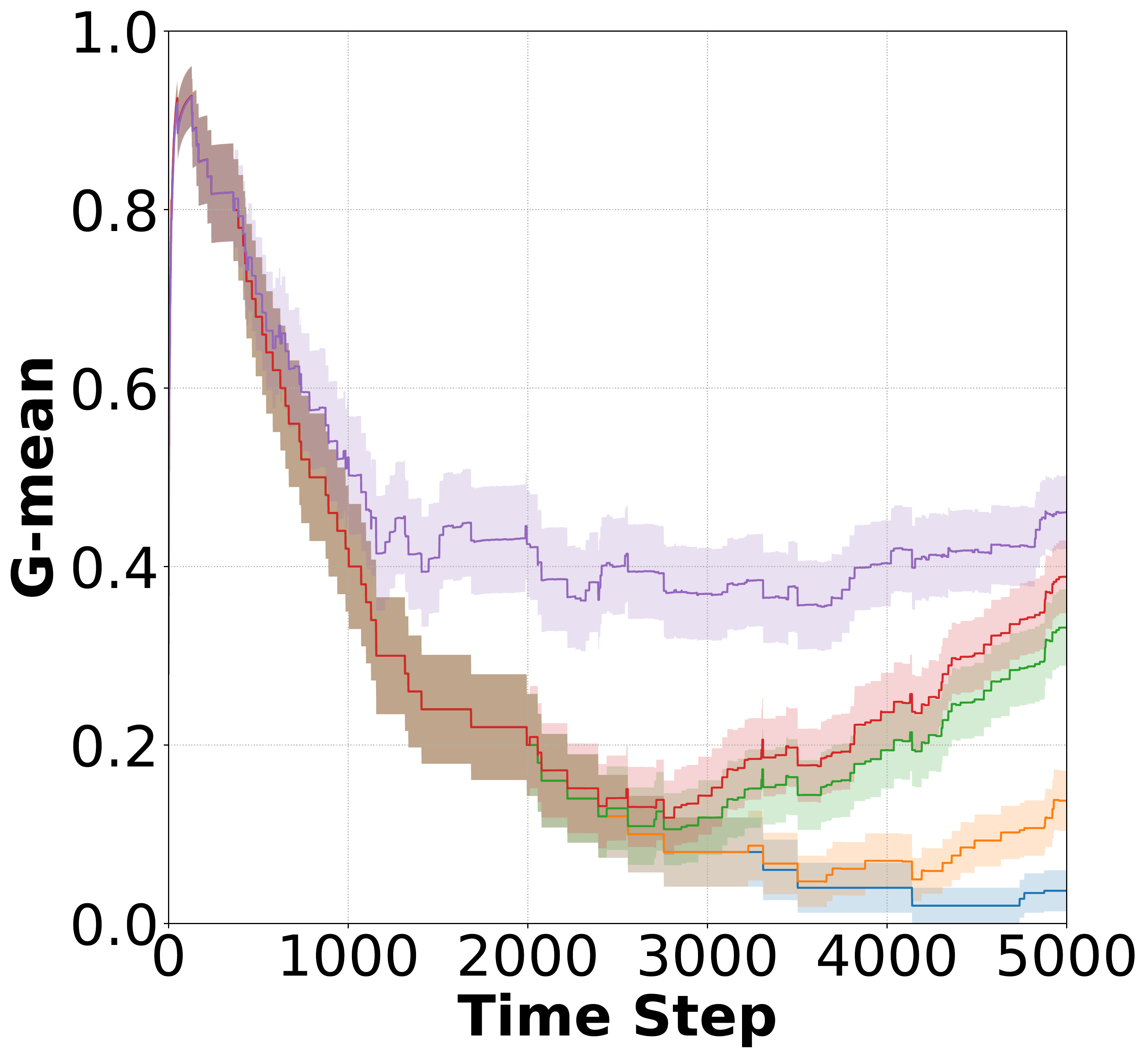}%
		\label{fig:analysis_circle_pp0_qbr}}
	
	\caption{\textit{QBR}'s behaviour for the Circle dataset (stationary) with different memory sizes $B=2, 50, 100, 500, 1000$ and class imbalance rates of $CI = 10\%, 1\%, 0.1\%$. QBR is very sensitive to the memory size as the learning speed and/or the final performance are heavily affected by the choice of B.}
\end{figure}

\subsection{Role of the adaptive rebalancing mechanism}\label{sec:role_areba}
This analysis has been conducted on the stationary \textit{Circle} dataset, i.e., without concept drift. Figures~\ref{fig:analysis_circle_pp10_qbr} - \ref{fig:analysis_circle_pp0_qbr} depict how the memory size affects the learning performance of \textit{QBR}. The left, middle and right columns correspond to experiments with class imbalance of $CI = 10\%, 1\%, 0.1\%$ respectively.

In Fig.~\ref{fig:analysis_circle_pp10_qbr}, where imbalance is mild i.e. $CI=10\%$, the final performance remains unaffected for $B \geq 50$, while for $B < 50$ it performs slightly worse. However, the learning speed is severely affected by the choice of $B$ i.e. it gets slower with an increasing value of $B$. For instance, \textit{QBR} with $B=500$ equalises the performance of $B=2$ at about $t=2000$.

In Fig.~\ref{fig:analysis_circle_pp1_qbr} where class imbalance is severe i.e. $CI=1\%$, both the learning speed and final performance are severely affected by the choice of $B$. The learning speed gets slower with an increasing value of $B$, for instance, \textit{QBR} with $B=500$ equalises the performance of $B=2$ at about $t=5000$. In regard to the final performance, after a certain threshold (here, $B \geq 50$), it significantly deteriorates as the value of $B$ increases. For instance, \textit{QBR} with $B=1000$ does not even equalise the performance of $B=2$ after 5000 time steps.

In Fig.~\ref{fig:analysis_circle_pp0_qbr} where imbalance is extreme ($CI=0.1\%$), the learning speed and final performance are severely affected by $B$, as previously. \textit{QBR} with $B=2$ is by far the best algorithm. \textit{QBR} with $B=50, 100$ only start closing the gap after 5000 time steps while $B=500, 1000$ perform poorly at $t=5000$.

The analogous experiments for \textit{AREBA} are shown in Figures \ref{fig:analysis_circle_pp10_areba} - \ref{fig:analysis_circle_pp0_areba}. Irrespective of the imbalance severity, after a certain threshold (here, $B \geq 50$), all \textit{AREBA} versions behave almost identically. To sum up, these important remarks can be made:
\begin{itemize}
	\item Without the adaptive rebalancing mechanism, \textit{QBR} is very sensitive to the choice of the memory size $B$ and, as a result, the learning speed and/or the final performance are significantly affected. The problem becomes more acute as class imbalance becomes more severe.
	\item The adaptive rebalancing mechanism makes \textit{AREBA} robust to the choice of the memory size $B$, irrespective of the imbalance severity. Specifically, the higher the value of $B$ the better, however, after a (dataset-specific) threshold, the improvement is negligible.
\end{itemize}

\subsection{Role of the memory size}\label{sec:role_B}
This section identifies the role of the memory size in the presence of outdated concepts. We examine both the leaning speed and final performance. To examine the learning speed we present the learning curves; when the curve is steep it means the algorithm is learning faster. The learning curves present the prequential G-mean at every time step. To examine the learning quality we present the final performance i.e. the one obtained at the last time step of the curve.

The analysis is conducted on the synthetic datasets on two settings, based on \cite{wang2018systematic}. The short setting (5000 steps, drift at $t=2500$) allows us to examine \textit{AREBA}'s behaviour immediately after the drift, while the long (20000 steps, drift at $t=10000$) allows us to examine \textit{AREBA}'s long-term behaviour.

\begin{figure}[t]
	\centering
	
	\subfloat[$CI=10\%$]{\includegraphics[scale=0.10]{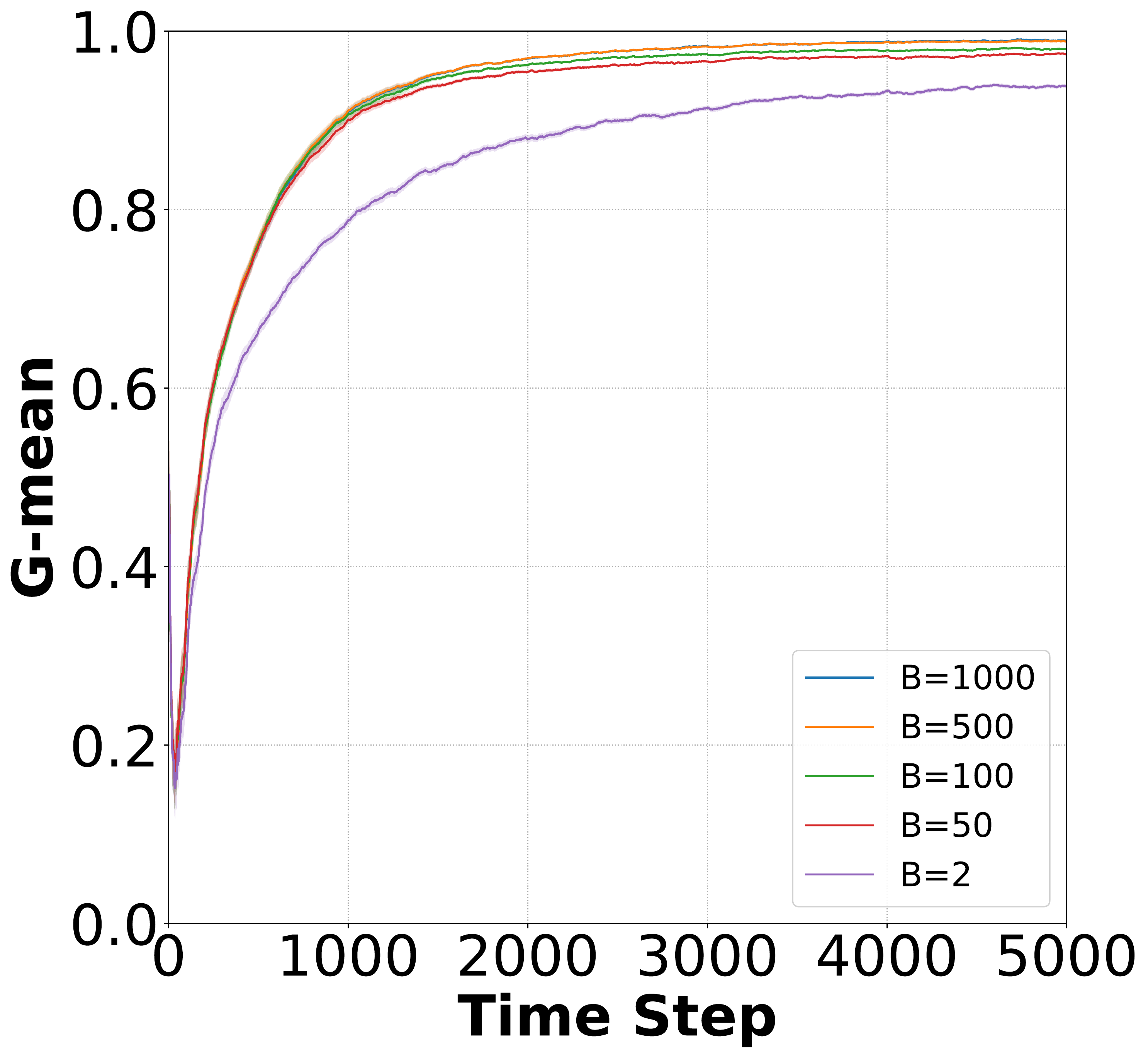}%
		\label{fig:analysis_circle_pp10_areba}}
	\subfloat[$CI=1\%$]{\includegraphics[scale=0.10]{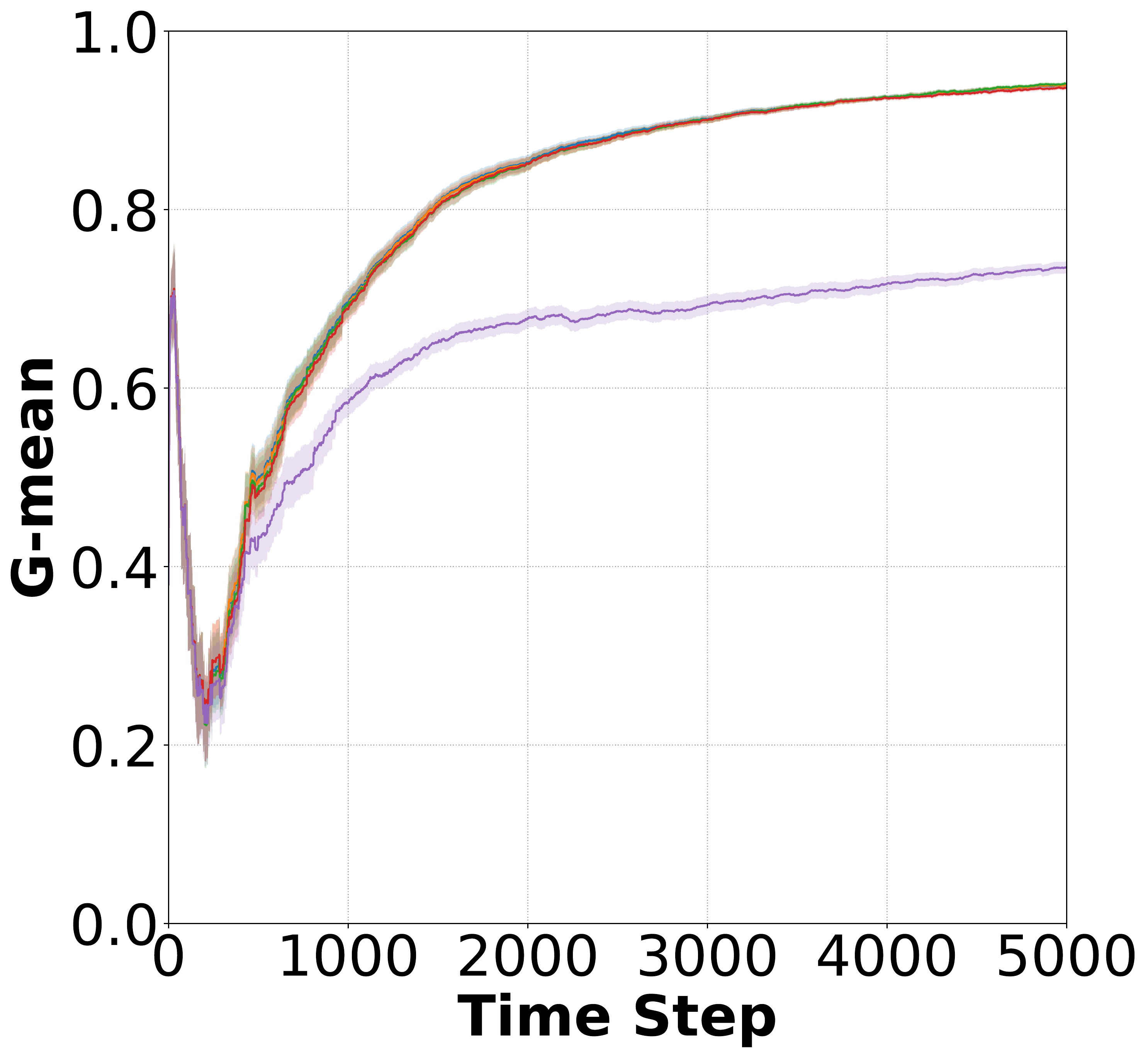}%
		\label{fig:analysis_circle_pp1_areba}}
	\subfloat[$CI=0.1\%$]{\includegraphics[scale=0.10]{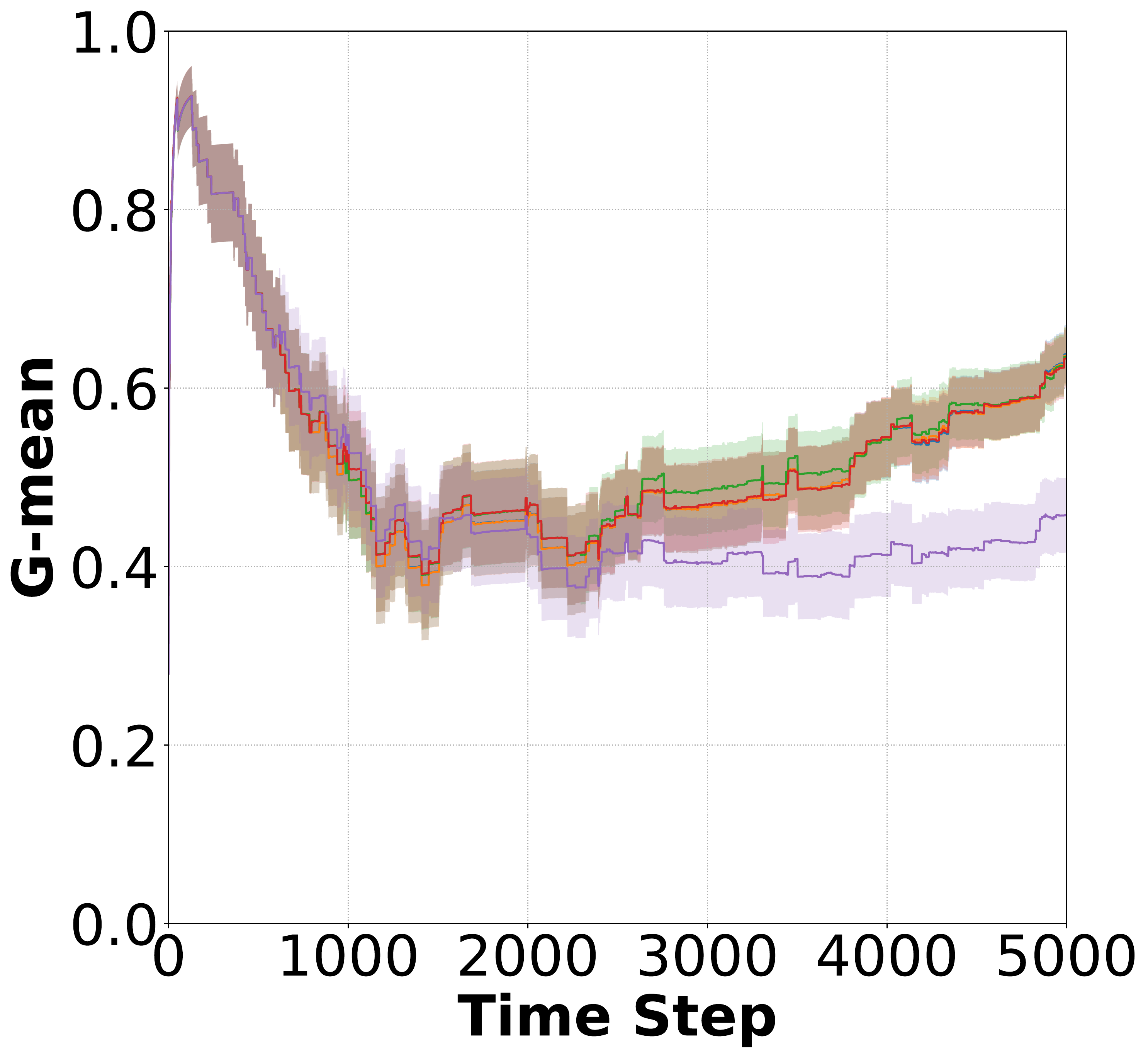}%
		\label{fig:analysis_circle_pp0_areba}}
	
	\caption{Role of the adaptive rebalancing mechanism for the Circle dataset (stationary) with different memory sizes ($B$) and class imbalance ($CI$) rates. The \textit{AREBA} mechanism is robust to different values of the memory size.}
\end{figure}

For the \textit{Circle} dataset the drift is defined as follows:
\begin{equation}\label{eq:posterior_circle}
\begin{aligned}
p(y=1 | x\textnormal{ }inside\textnormal{ }circle) = 1.0 & \longrightarrow 0.0\\
p(y=1 | x\textnormal{ }outside\textnormal{ }circle) = 0.0 & \longrightarrow 1.0\\
p(y=0 | x\textnormal{ }outside\textnormal{ }circle) = 1.0 & \longrightarrow 0.0\\
p(y=0 | x\textnormal{ }inside\textnormal{ }circle) = 0.0 & \longrightarrow 1.0\\
\end{aligned}
\end{equation}

\begin{figure*}[t!]
	\centering
	
	\subfloat[$CI=10\%$]{\includegraphics[scale=0.11]{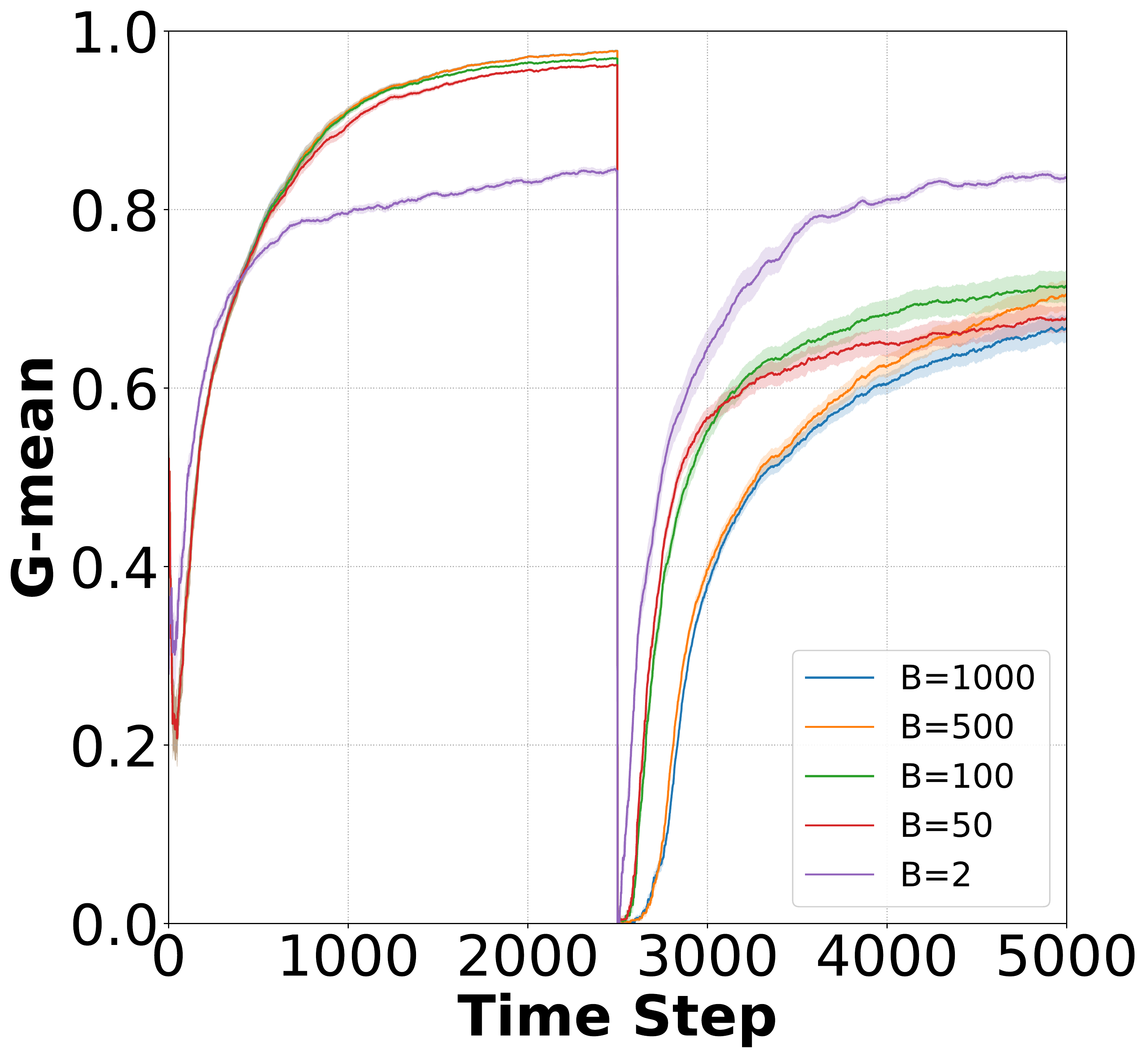}%
		\label{fig:analysis_circle_pp10_posterior_areba}}
	\subfloat[$CI=1\%$]{\includegraphics[scale=0.11]{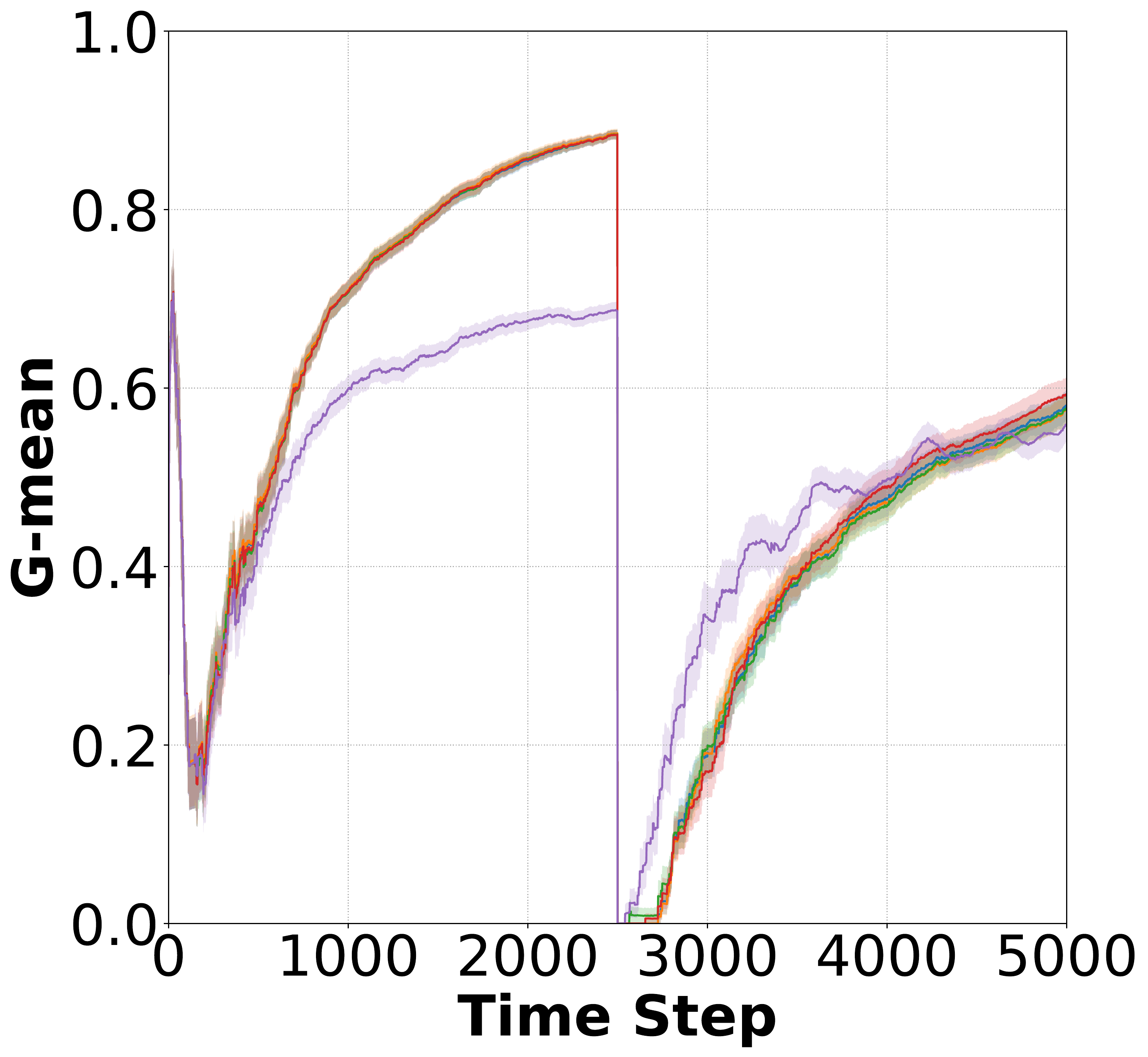}%
		\label{fig:analysis_circle_pp1_posterior_areba}}
	\subfloat[$CI=0.1\%$]{\includegraphics[scale=0.11]{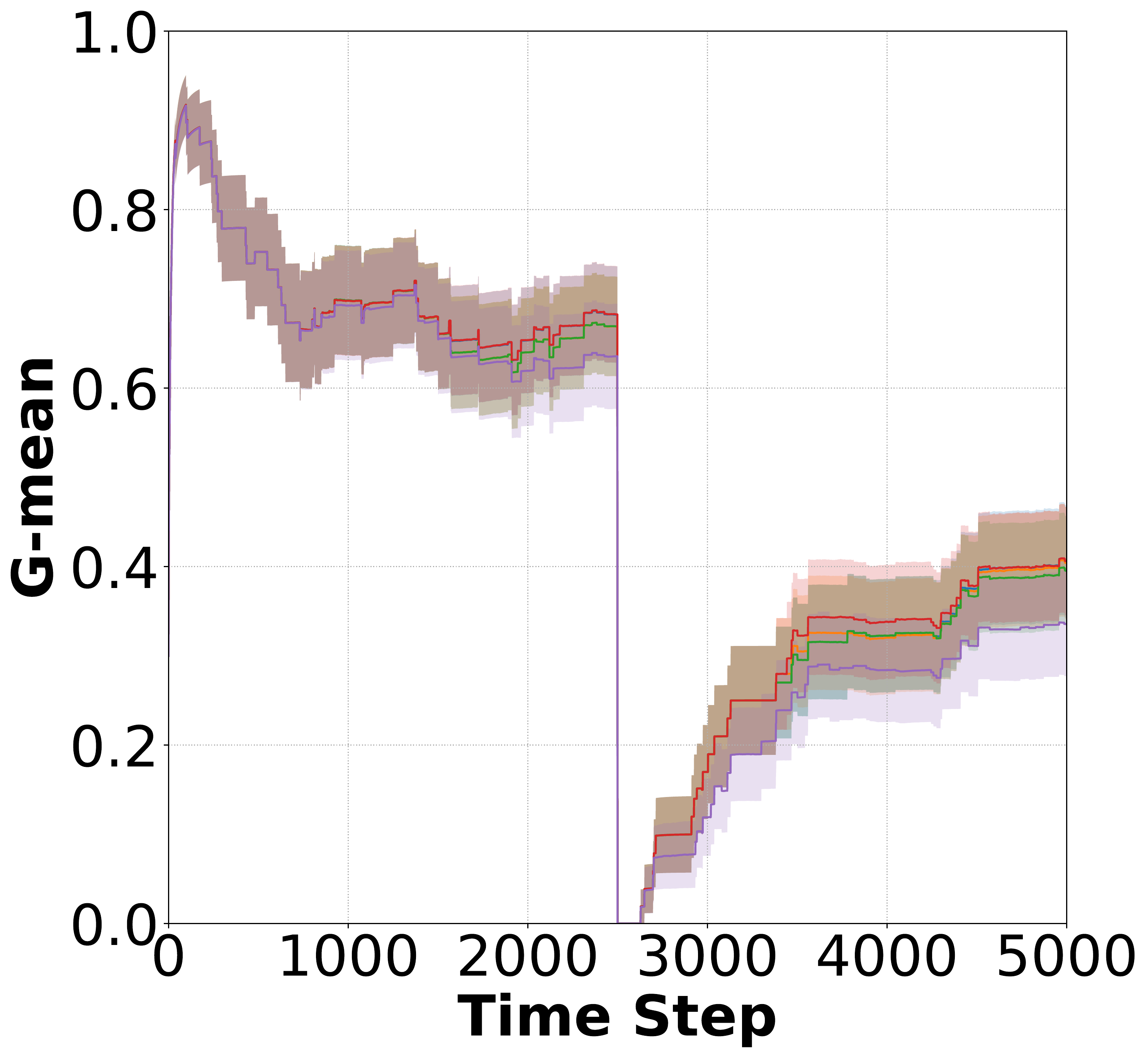}%
		\label{fig:analysis_circle_pp0_posterior_areba}}
	\subfloat[\textit{Pre-drift}]{\includegraphics[scale=0.11]{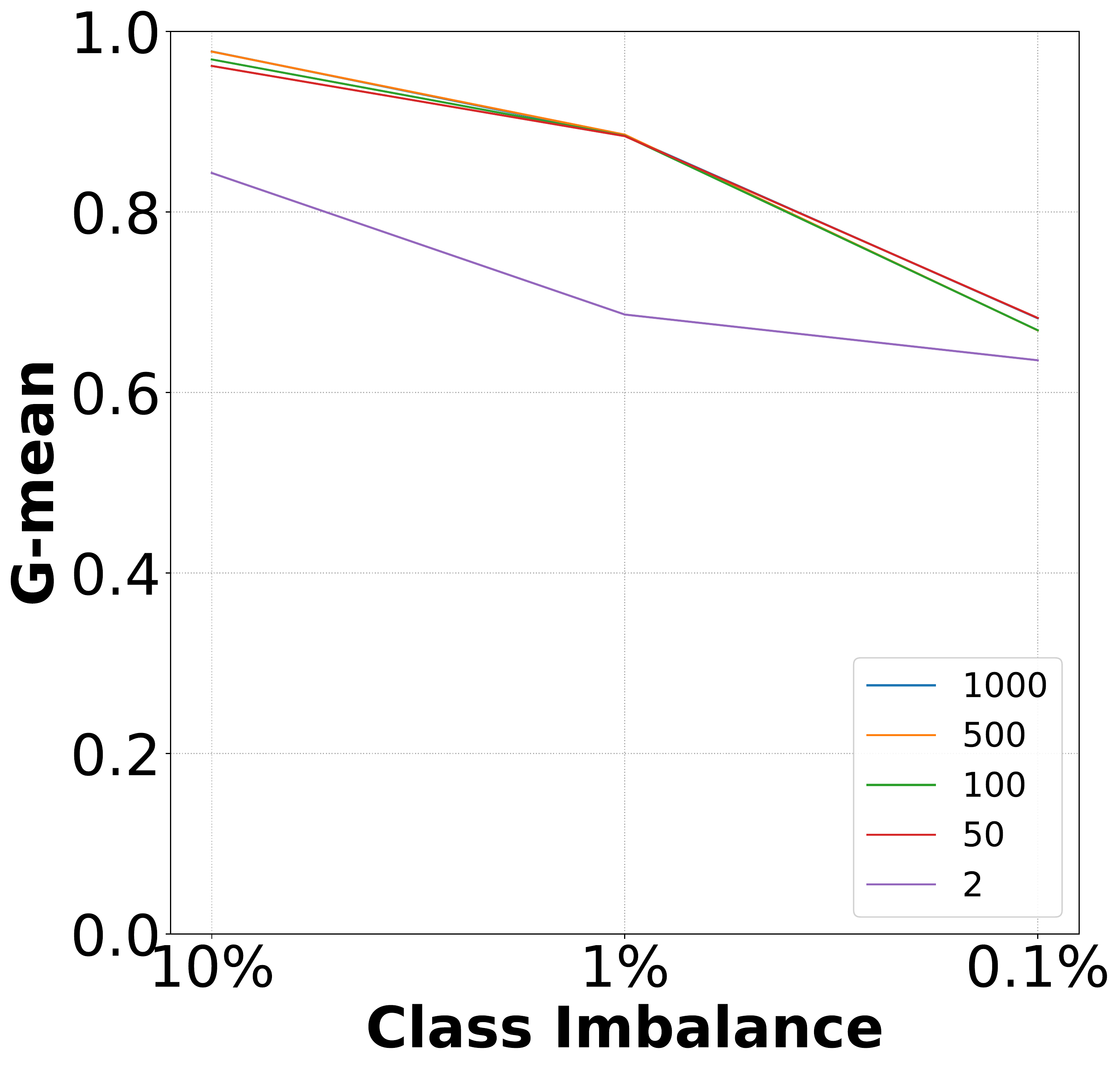}%
		\label{fig:circle_pre_drift}}
	\subfloat[\textit{Post-drift}]{\includegraphics[scale=0.11]{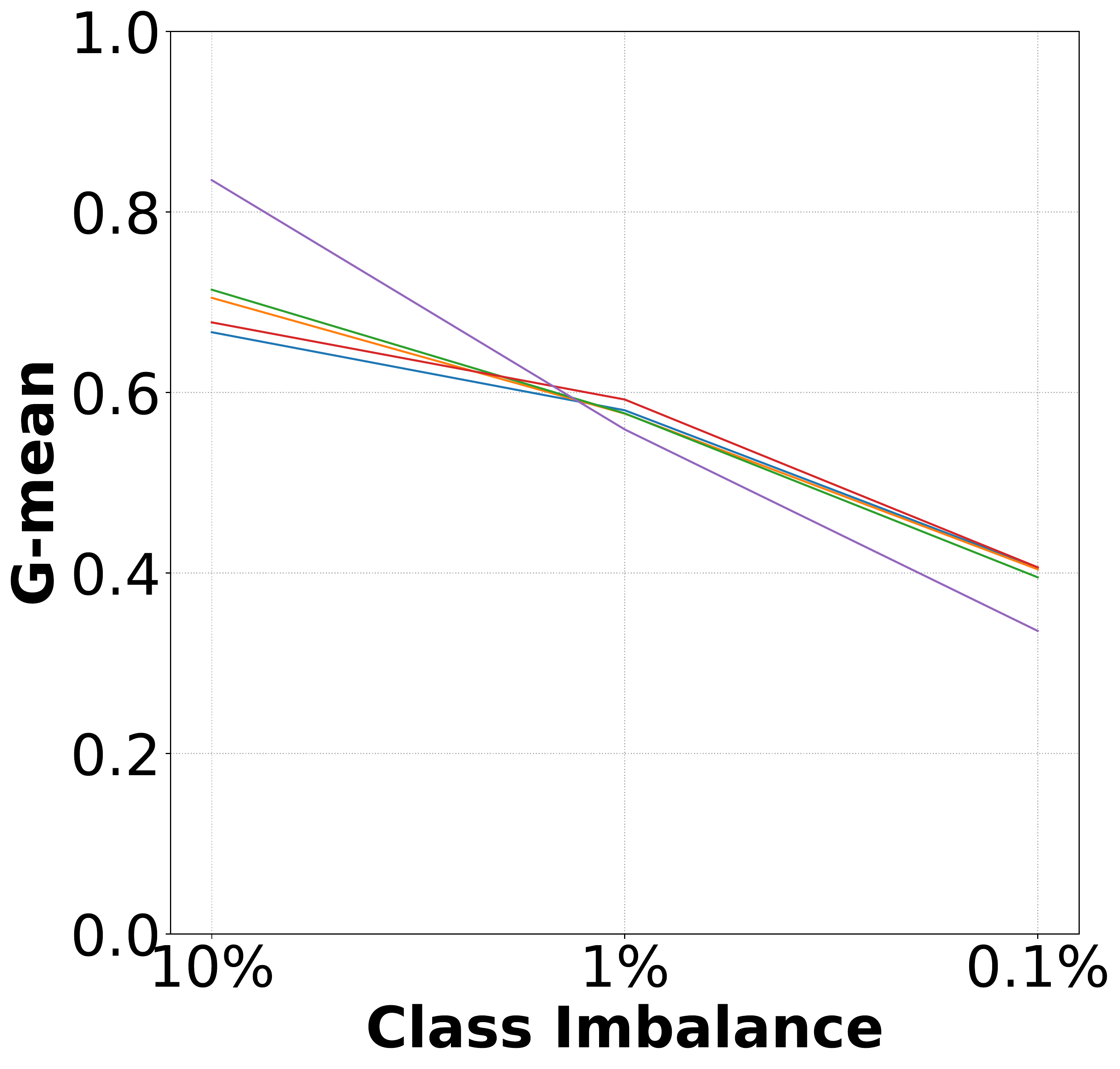}%
		\label{fig:circle_post_drift}}
	
	\caption{Role of the memory size ($B$) for \textit{AREBA} in the \textit{Circle} dataset with posterior probability drift and various class imbalance ($CI$) rates. The last two figures depict the final performance before ($t=2499$) and after ($t=5000$) the drift.} 
	\label{fig:analysis_circle}
\end{figure*}

\begin{figure*}[t]
	\centering
	
	\subfloat[$CI=10\%$]{\includegraphics[scale=0.11]{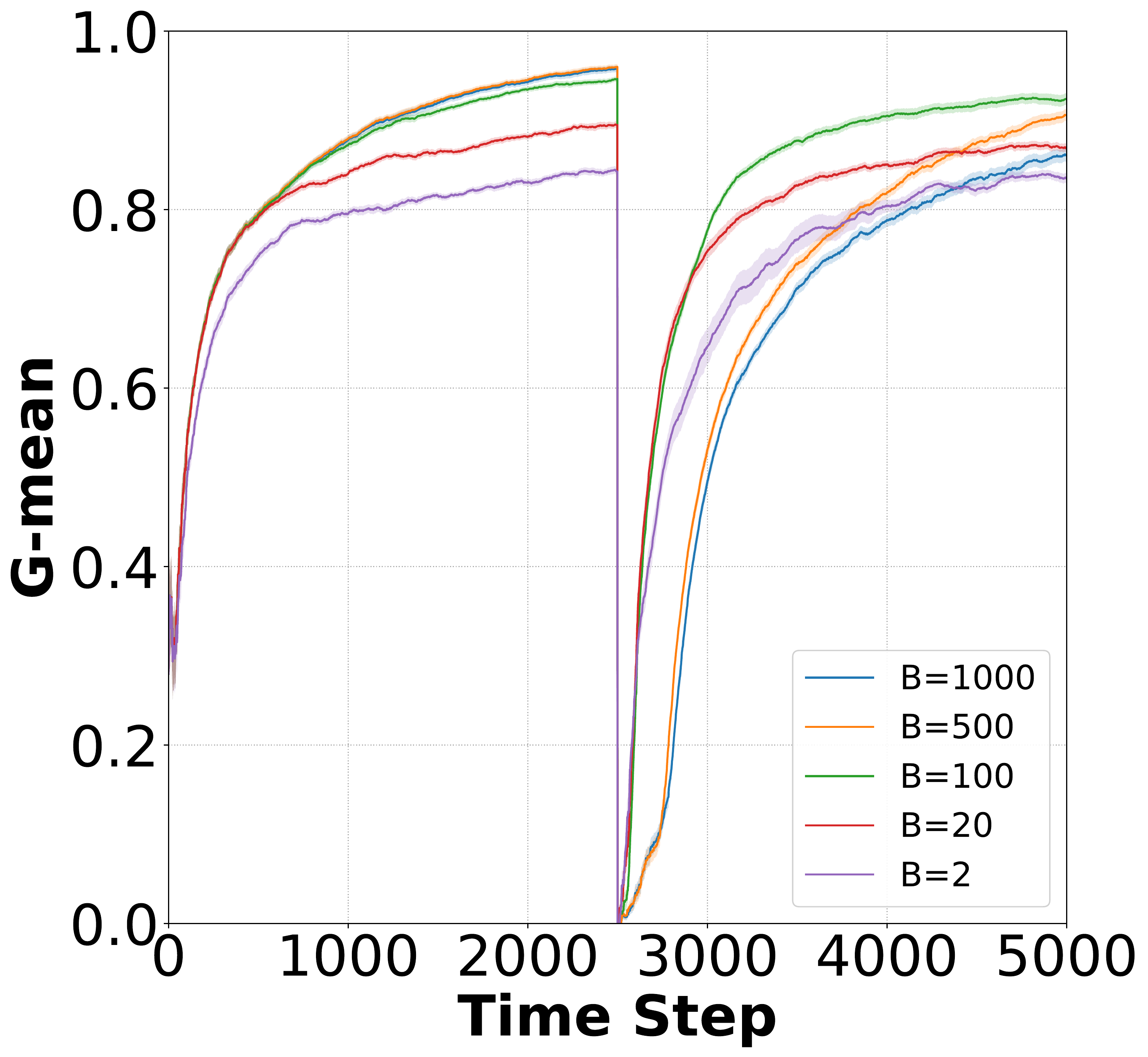}%
		\label{fig:analysis_sine_pp10_posterior_areba}}
	\subfloat[$CI=1\%$]{\includegraphics[scale=0.11]{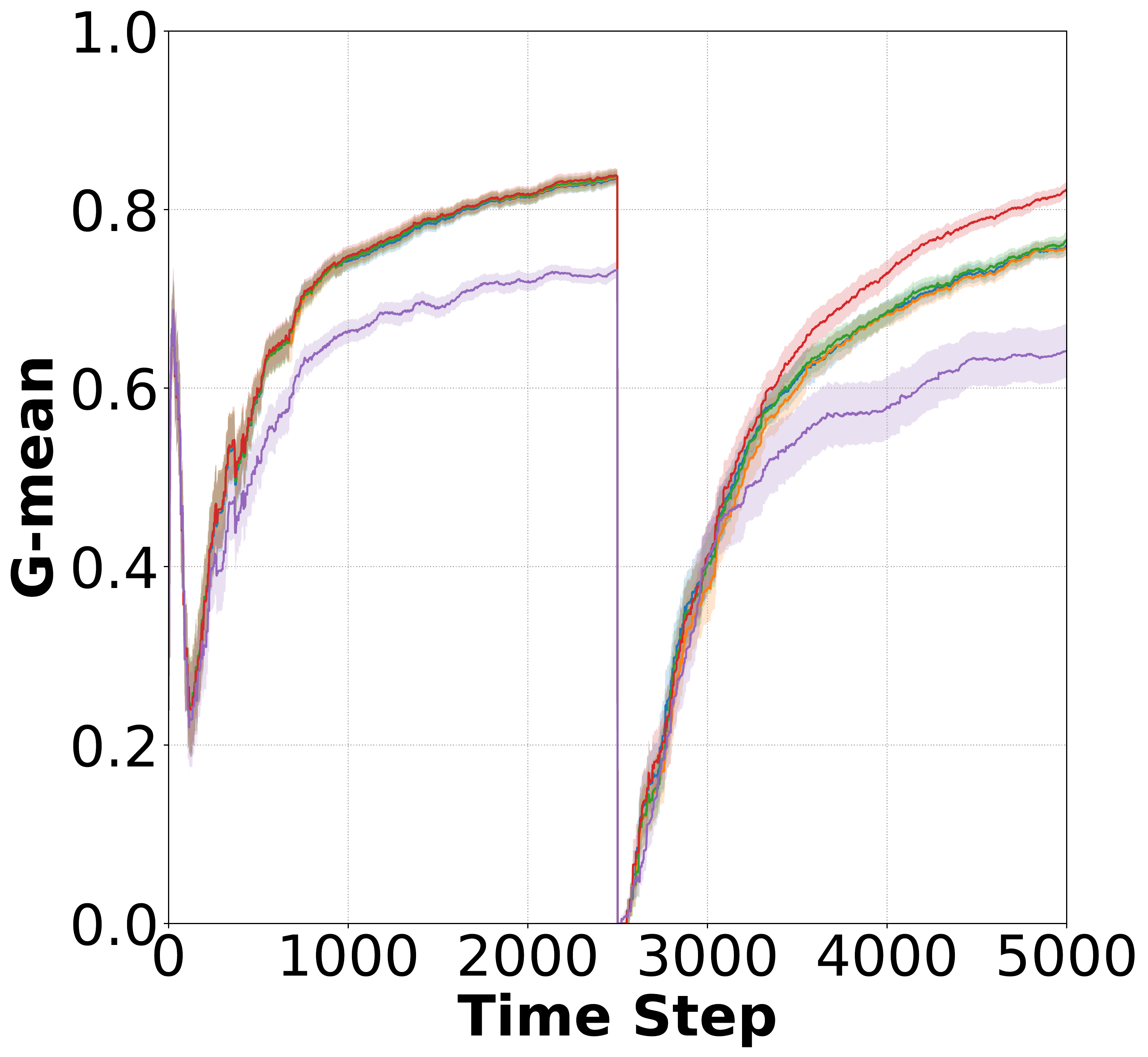}%
		\label{fig:analysis_sine_pp1_posterior_areba}}
	\subfloat[$CI=0.1\%$]{\includegraphics[scale=0.11]{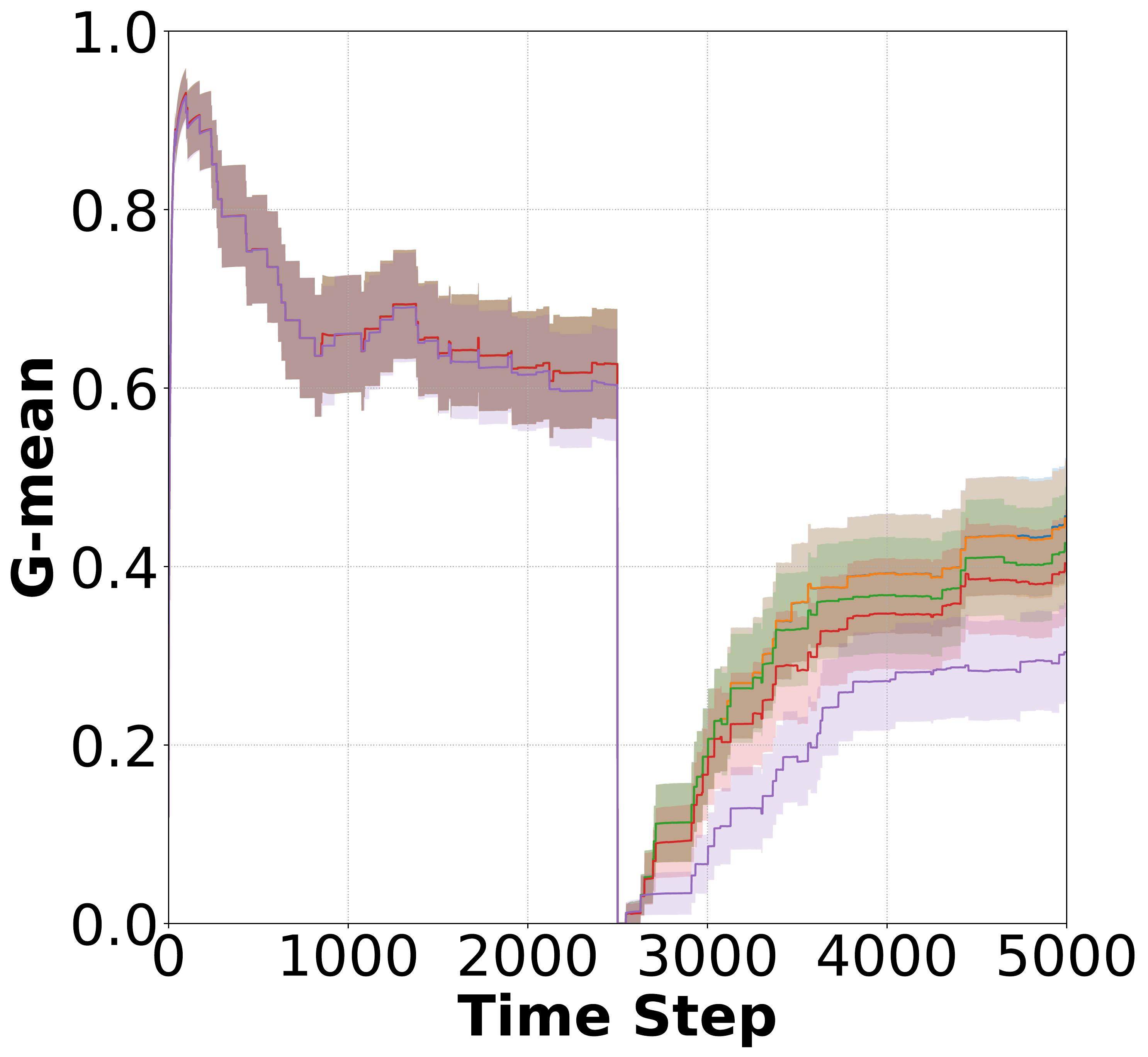}%
		\label{fig:analysis_sine_pp0_posterior_areba}}
	\subfloat[\textit{Pre-drift}]{\includegraphics[scale=0.11]{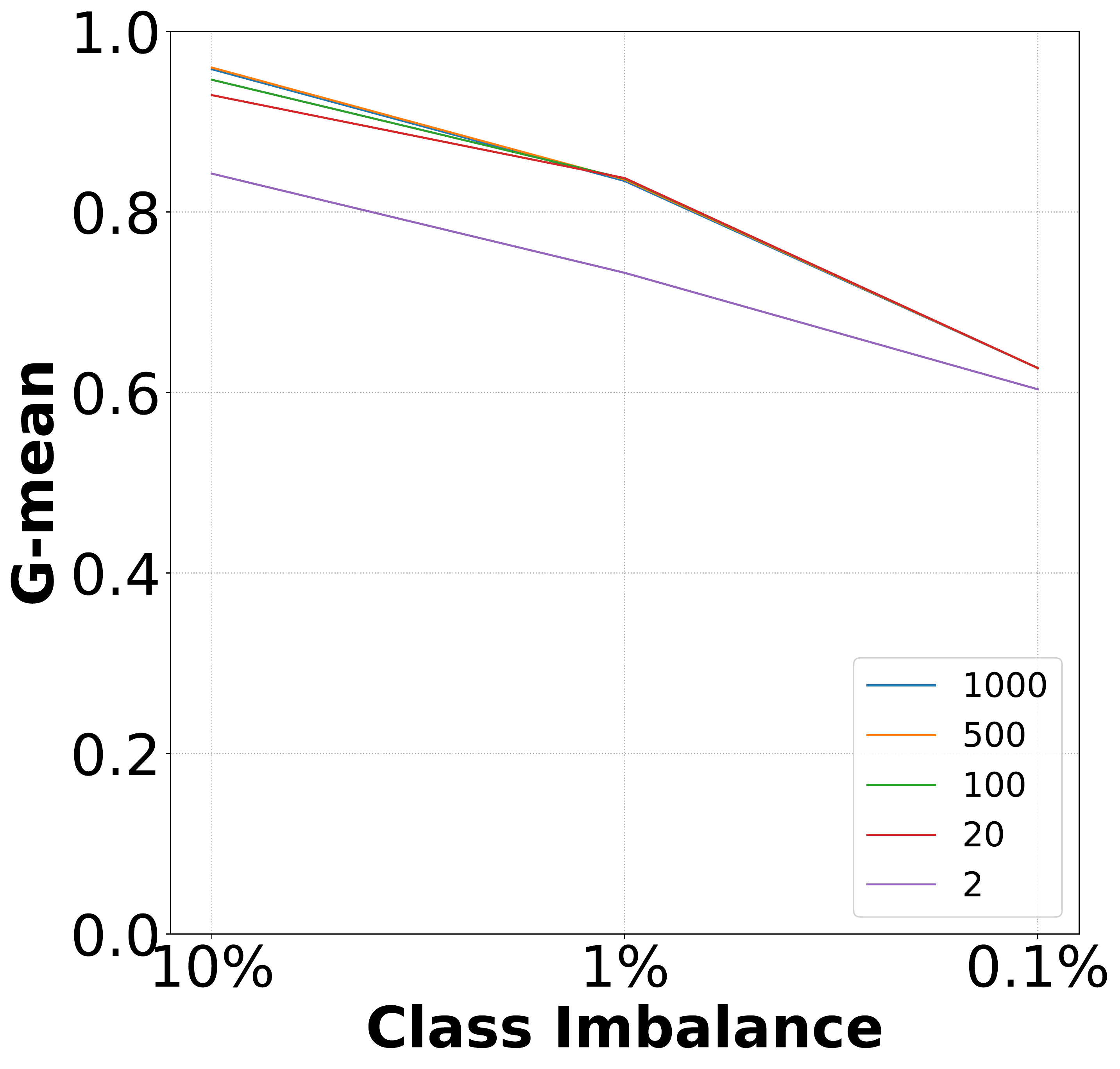}%
		\label{fig:sine_pre_drift}}
	\subfloat[\textit{Post-drift}]{\includegraphics[scale=0.11]{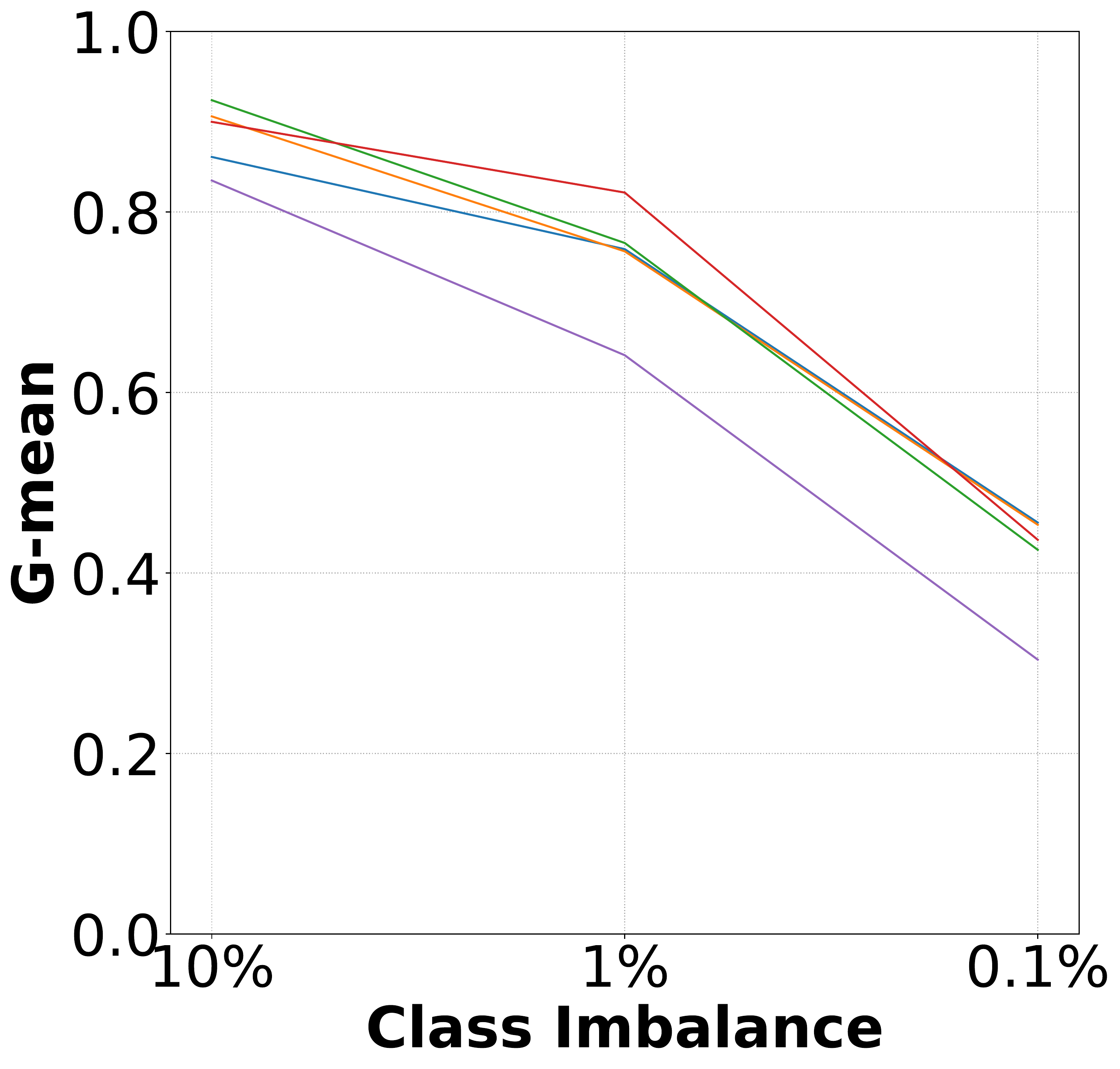}%
		\label{fig:sine_post_drift}}
	
	\caption{Role of the memory size ($B$) for \textit{AREBA} in the \textit{Sine} dataset with posterior probability drift and various class imbalance ($CI$) rates. The last two figures depict the final performance before ($t=2499$) and after ($t=5000$) the drift.} 
	\label{fig:analysis_sine}
\end{figure*}

\begin{figure*}[t]
	\centering
	
	\subfloat[$CI=10\%$]{\includegraphics[scale=0.11]{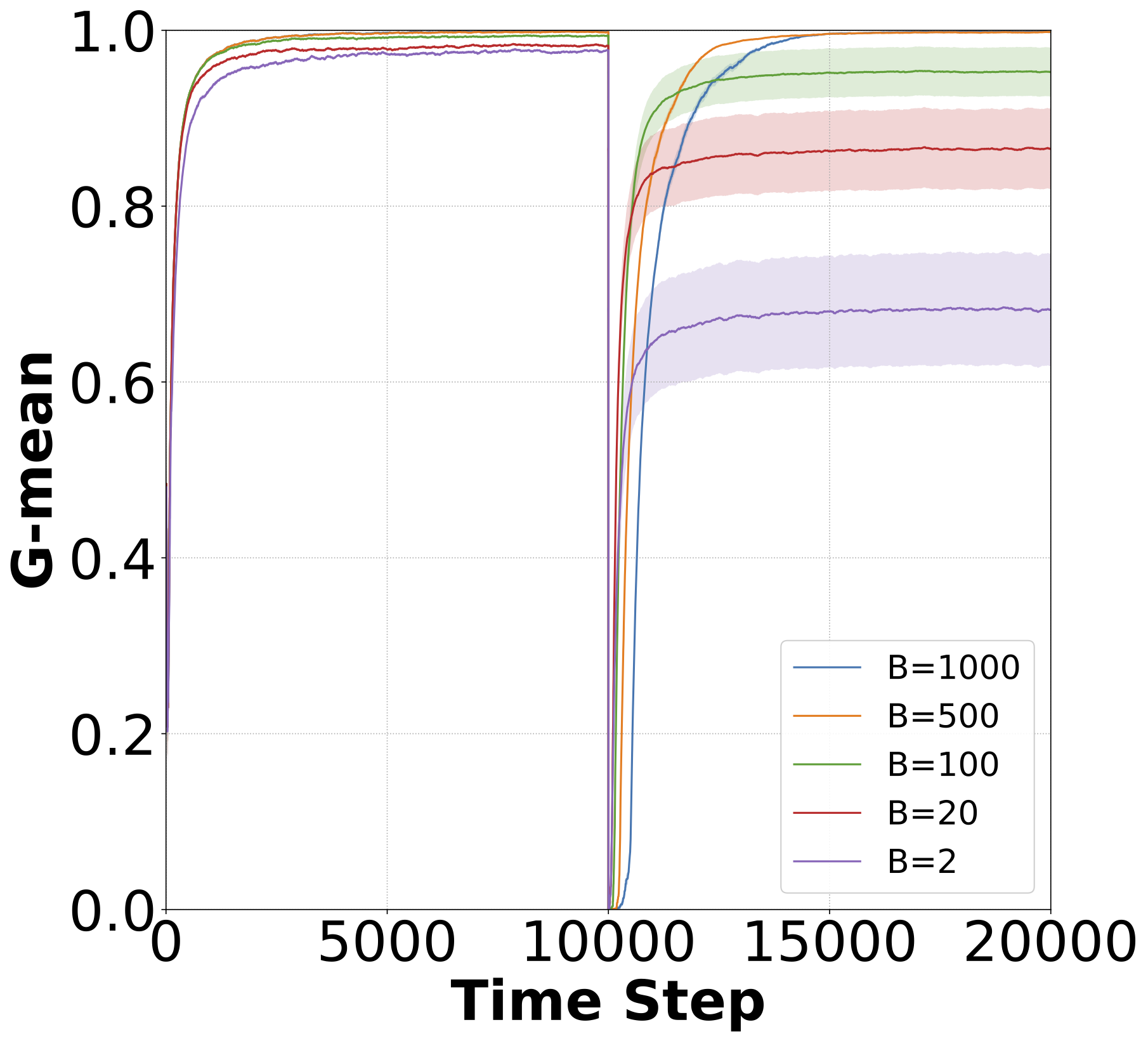}%
		\label{fig:analysis_sea_pp10_posterior_areba}}
	\subfloat[$CI=1\%$]{\includegraphics[scale=0.11]{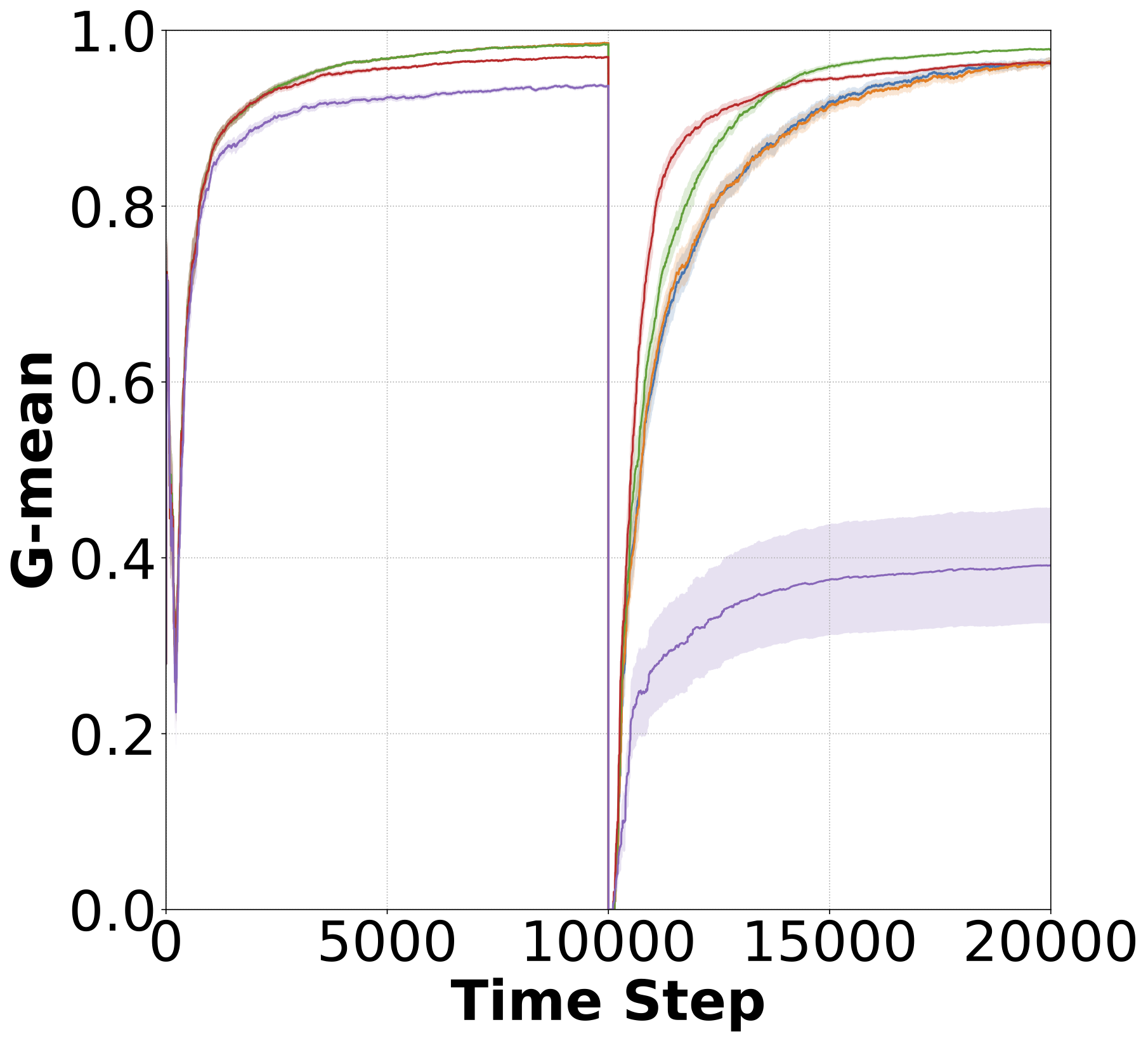}%
		\label{fig:analysis_sea_pp1_posterior_areba}}
	\subfloat[$CI=0.1\%$]{\includegraphics[scale=0.11]{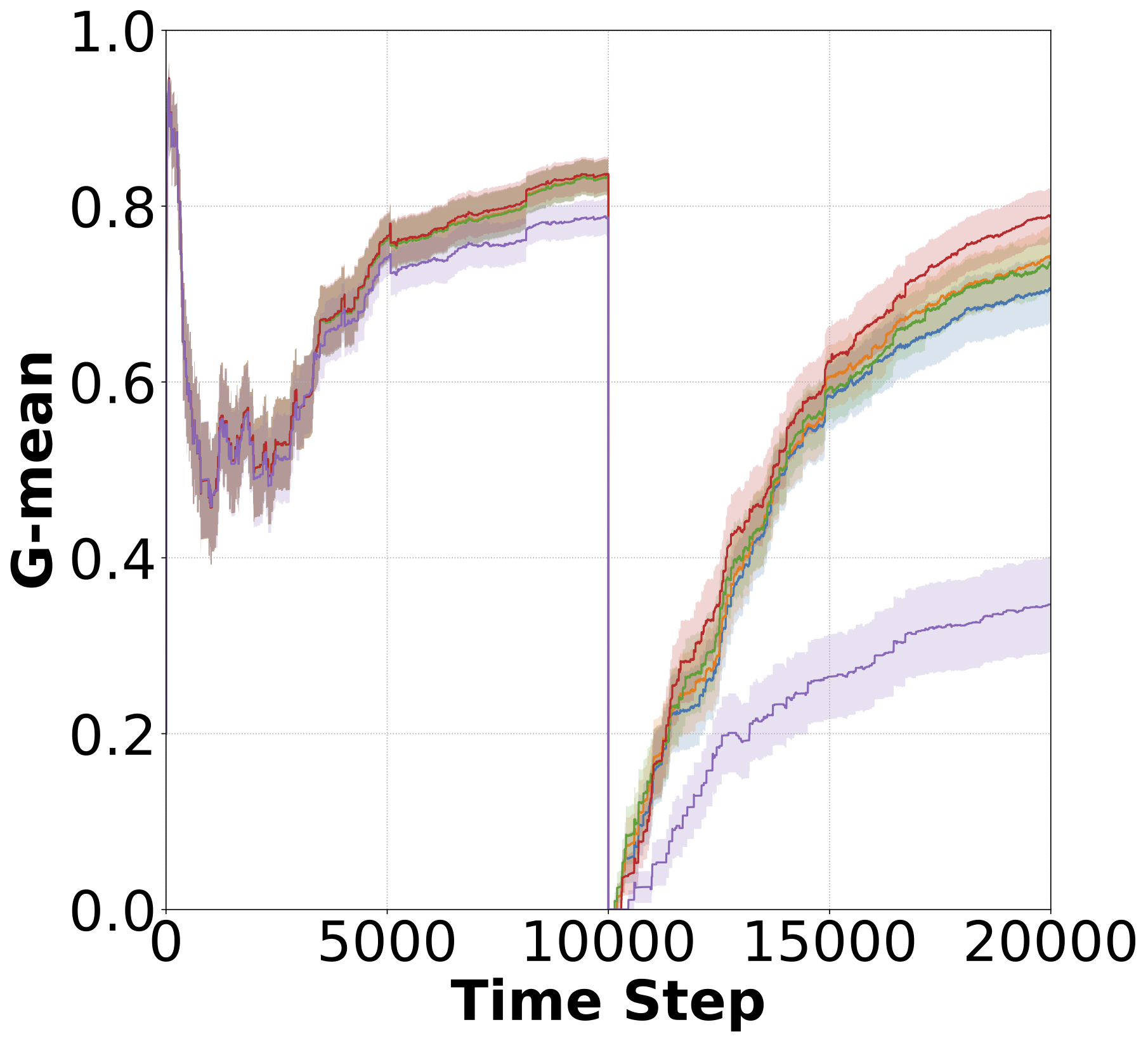}%
		\label{fig:analysis_sea_pp0_posterior_areba}}
	\subfloat[\textit{Pre-drift}]{\includegraphics[scale=0.11]{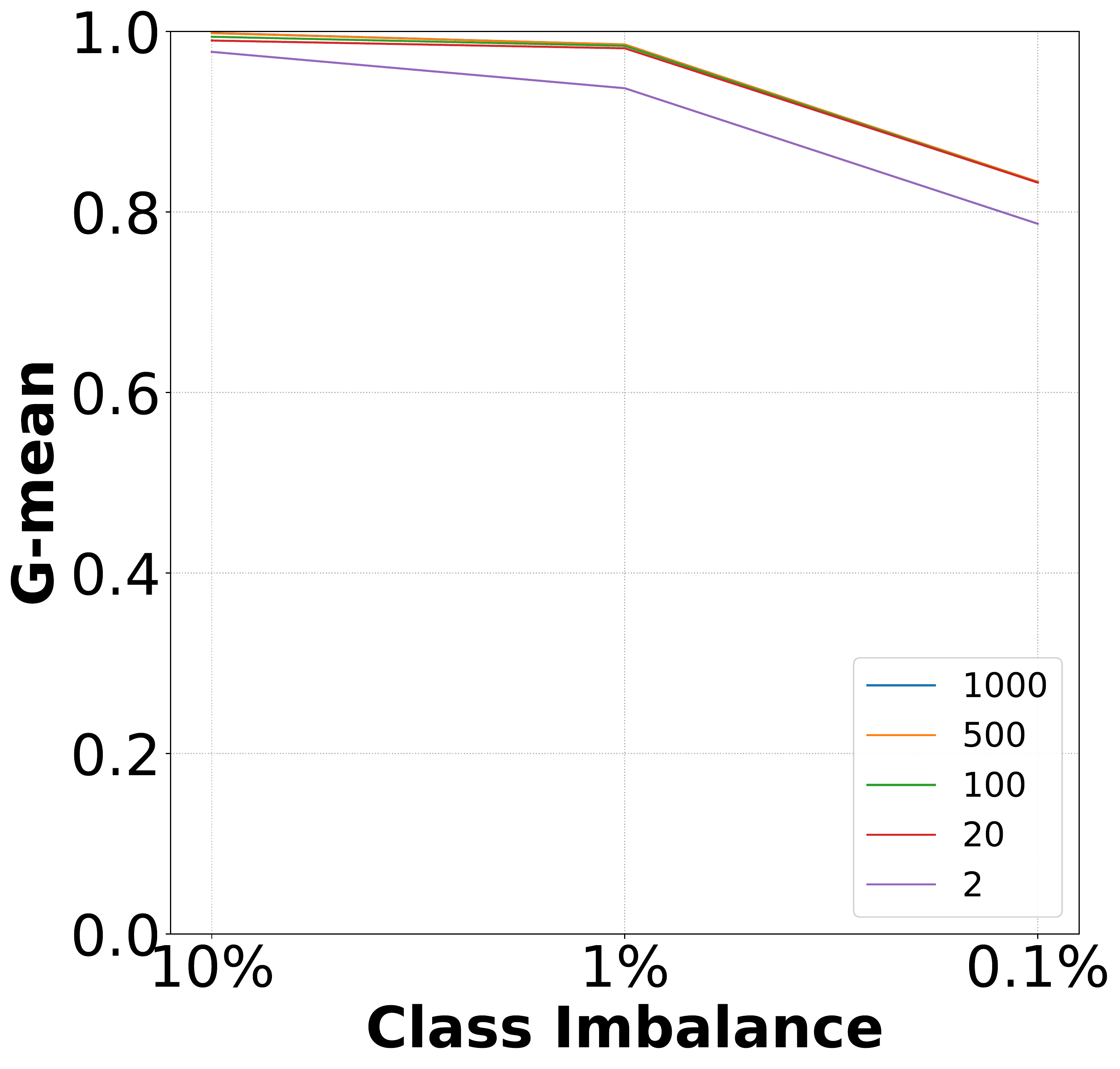}%
		\label{fig:sea_pre_drift}}
	\subfloat[\textit{Post-drift}]{\includegraphics[scale=0.11]{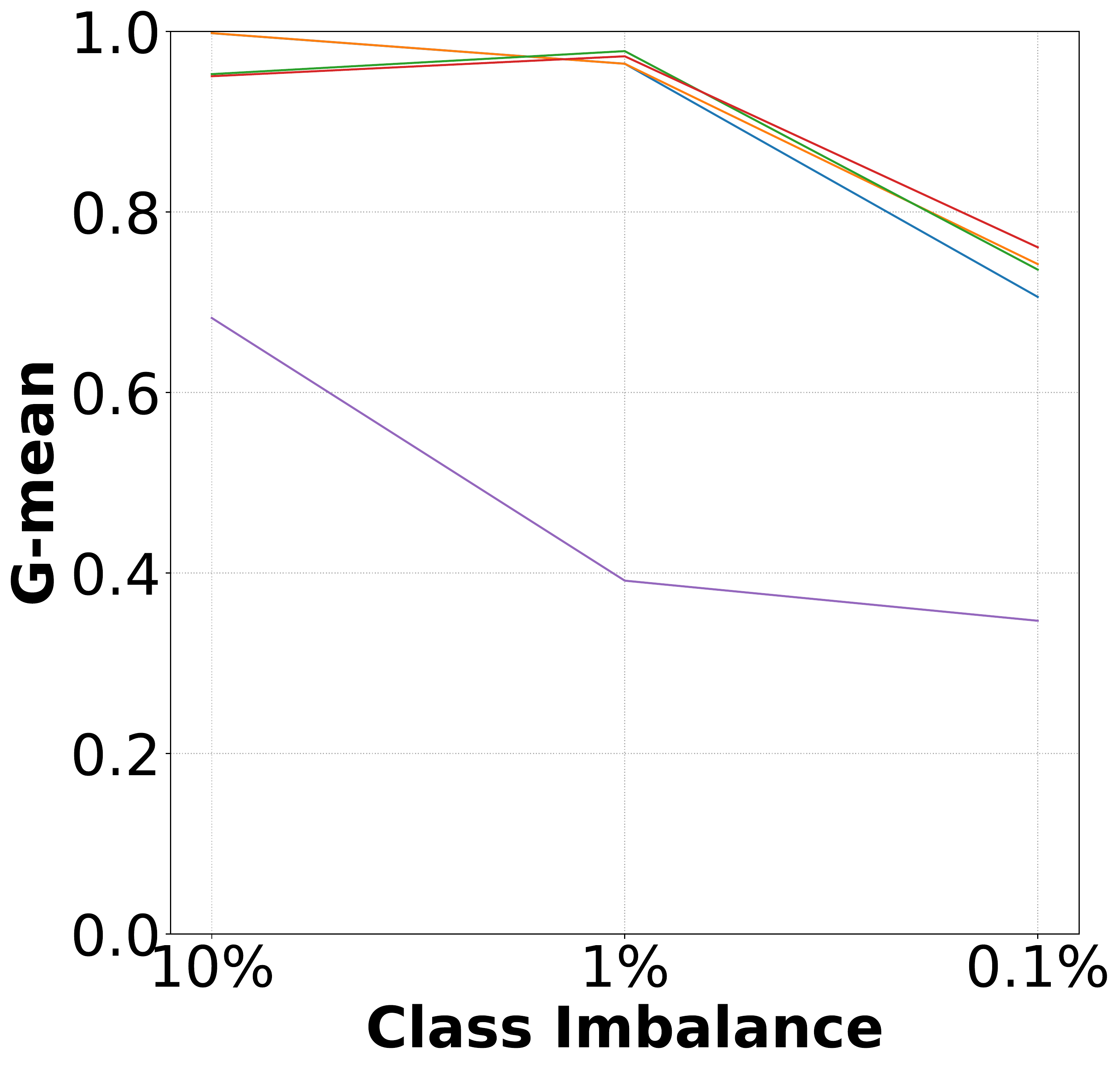}%
		\label{fig:sea_post_drift}}
	
	\caption{Role of the memory size ($B$) for \textit{AREBA} in the \textit{Sea} dataset with posterior probability drift and various class imbalance ($CI$) rates. The last two figures depict the final performance before ($t=9999$) and after ($t=20000$) the drift.} 
	\label{fig:analysis_sea}
\end{figure*}

Figure~\ref{fig:analysis_circle} shows the results for the \textit{Circle} dataset under posterior drift. Figures~\ref{fig:analysis_circle_pp10_posterior_areba} - \ref{fig:analysis_circle_pp0_posterior_areba} depict \textit{AREBA}'s learning curves for various memory sizes with class imbalance of $CI = 10\%, 1\%, 0.1\%$ respectively. Figures~\ref{fig:circle_pre_drift} and \ref{fig:circle_post_drift} depict the final performances before and after the drift respectively.

We start with the case of mild imbalance (Fig.~\ref{fig:analysis_circle_pp10_posterior_areba}). For the part of the curves before the drift, we conclude that the higher the value of $B$ the better, however, after a (dataset-specific) threshold, the improvement is negligible. This has been studied in Section~\ref{sec:role_areba} (Fig~\ref{fig:analysis_circle_pp10_areba}), therefore, from now on, we turn our attention to the part of the learning curves after the drift.

For the learning speed, \textit{AREBA} with $B=500, 1000$ is slower than $B=20,100$ which in turn is slower than $B=2$. For the final performance, \textit{AREBA} with $B=2$ significantly outperforms the rest. We observe that the smaller the value of $B$ the better the results. This is also clear in Fig.~\ref{fig:circle_post_drift}. This is expected as immediately after a drift, a larger value of $B$ means that more outdated examples exist in the queues.

Surprisingly, this no longer holds when imbalance becomes severe or extreme (Figs.~\ref{fig:analysis_circle_pp1_posterior_areba} and \ref{fig:analysis_circle_pp0_posterior_areba}) where all \textit{AREBA} versions yield the same final performance. This is attributed to two reasons. Firstly, a small value of $B$ is suitable in case of outdated examples with mild imbalance (Fig.~\ref{fig:analysis_circle_pp10_posterior_areba}). Secondly, in case of severe imbalance without drift, a large value of $B$ is suitable (Fig.~\ref{fig:analysis_circle_pp1_areba}). In the presence of both drift and severe imbalance, the aforementioned are in conflict with each other and, interestingly, it appears that imbalance becomes the key issue when it is severe rather than drift.

Overall, an important trade-off exists. In stationary settings, the higher the value of $B$ the better the results, as shown in Fig~\ref{fig:analysis_circle_pp10_areba}. In nonstationary settings, the smaller the value of $B$ the better the performance, as shown in Fig~\ref{fig:analysis_circle_pp10_posterior_areba}. The effect of this trade-off appears to be fading away in the presence of severe or extreme imbalance, as shown in Fig~\ref{fig:analysis_circle_pp1_posterior_areba} - \ref{fig:analysis_circle_pp0_posterior_areba}. The importance of this trade-off is high and should be carefully assessed, as the optimal value of $B$ depends on the dataset.

For the \textit{Sine} dataset the posterior drift is defined as follows:
\begin{equation}\label{eq:posterior_sine}
\begin{aligned}
p(y=1 | x\textnormal{ }below\textnormal{ }curve) = 1.0 & \longrightarrow 0.0\\
p(y=1 | x\textnormal{ }above\textnormal{ }curve) = 0.0 & \longrightarrow 1.0\\
p(y=0 | x\textnormal{ }above\textnormal{ }curve) = 1.0 & \longrightarrow 0.0\\
p(y=0 | x\textnormal{ }below\textnormal{ }curve) = 0.0 & \longrightarrow 1.0\\
\end{aligned}
\end{equation}

Figure~\ref{fig:analysis_sine} shows the results for the \textit{Sine} dataset under posterior drift. Figures~\ref{fig:analysis_sine_pp10_posterior_areba} - \ref{fig:analysis_sine_pp0_posterior_areba} depict \textit{AREBA}'s learning curves while Figures~\ref{fig:sine_pre_drift} and \ref{fig:sine_post_drift} depict the final performances. For mild imbalance (Fig.~\ref{fig:analysis_sine_pp10_posterior_areba}), the learning speed of \textit{AREBA} with $B=500, 1000$ is slower than $B=20,100$. In case of severe imbalance (Fig.~\ref{fig:analysis_sine_pp1_posterior_areba}), we can conclude that the key issue becomes the imbalance rather than the drift. Notably, \textit{AREBA} with $B=1000$ outperforms $B=2$ and equalises the performance of $B=100, 500$ despite containing significantly more outdated examples in its queues. In Fig.~\ref{fig:sine_post_drift}, the best final performance when imbalance is mild ($CI=10\%$) occurs when $B=100$. The best final performance when imbalance is severe ($CI=1\%$) occurs when $B=20$. Under conditions of mild imbalance (Fig~\ref{fig:analysis_sine_pp10_posterior_areba}), the aforementioned trade-off is not as clear as it was for the \textit{Circle} dataset (Fig~\ref{fig:analysis_circle_pp10_posterior_areba}). Therefore, we note that the need for some experimentation to find a suitable choice of $B$ becomes even more important.

For the \textit{Sea} dataset the posterior drift is defined in Eq.~\ref{eq:posterior_sea}. Figure~\ref{fig:analysis_sea} shows the results for the \textit{Sea} dataset under posterior drift. Figures~\ref{fig:analysis_sea_pp10_posterior_areba} - \ref{fig:analysis_sea_pp0_posterior_areba} depict \textit{AREBA}'s learning curves while Figures~\ref{fig:sea_pre_drift} and \ref{fig:sea_post_drift} depict the final performances.
\begin{equation}\label{eq:posterior_sea}
\begin{aligned}
p(y=1 | x_1 + x_2 \leq 7) = 1.0 & \longrightarrow 0.0\\
p(y=1 | x_1 + x_2 > 7) = 0.0 & \longrightarrow 1.0\\
p(y=0 | x_1 + x_2 > 7) = 1.0 & \longrightarrow 0.0\\
p(y=0 | x_1 + x_2\leq 7) = 0.0 & \longrightarrow 1.0\\
\end{aligned}
\end{equation}

Notice that the experiment runs for 20000 steps to inspect \textit{AREBA}'s long-term behaviour. We start with the case of mild imbalance (Fig.~\ref{fig:analysis_sea_pp10_posterior_areba}). While immediately after the drift, \textit{AREBA} with $B= 500, 1000$ appear to be learning slower, given additional time, the two eventually outperform the rest. This is expected because after a long time without any recurring drift, the data can be considered as stationary and, hence, we reach the same conclusion as previously (Section~\ref{sec:role_areba}, Fig.~\ref{fig:analysis_circle_pp10_areba}).

For severe imbalance (Fig.~\ref{fig:analysis_sea_pp1_posterior_areba}), we conclude that the key issue becomes the imbalance rather than the drift. Notably, \textit{AREBA} with $B=1000$ outperforms $B=2$ and equalises $B=20, 500$ despite containing significantly more outdated examples in its queues. The best final performance when imbalance is severe occurs when $B=100$. This is also shown in Fig.~\ref{fig:sea_post_drift}. For the case of extreme class imbalance i.e. $CI=0.1\%$, all \textit{AREBA} versions (except when $B=2$) perform similarly, although, $B=20$ has a slight advantage.

To sum up, the following important remarks can be made:
\begin{itemize}
	\item For mild class imbalance, the general trend is that the lower the value of the memory size $B$, the better \textit{AREBA} performs. This is attributed to the fact that a smaller amount of outdated examples are contained in the queues. The optimal value of $B$ depends on the dataset.
	\item When severe imbalance exists, it appears to be the key problem rather than drift. Surprisingly, \textit{AREBA} with large values of $B$ has been shown to perform similarly or even better than \textit{AREBA} with lower values. Interestingly, this is somewhat in contradiction to the previous point as the former contains more outdated examples in the queues.
	\item The previous are useful guidelines to help with the selection of the memory size parameter. However, some experimentation is still necessary to obtain the best value of $B$. We discuss the role of $B$ further in Section~\ref{sec:discussion}.
\end{itemize}

\section{Comparative Study}\label{sec:study}

\subsection{Stationary data}\label{sec:stationary}
\begin{figure}[t!]
	\centering
	
	\subfloat[$G\textnormal{-}mean$]{\includegraphics[scale=0.10]{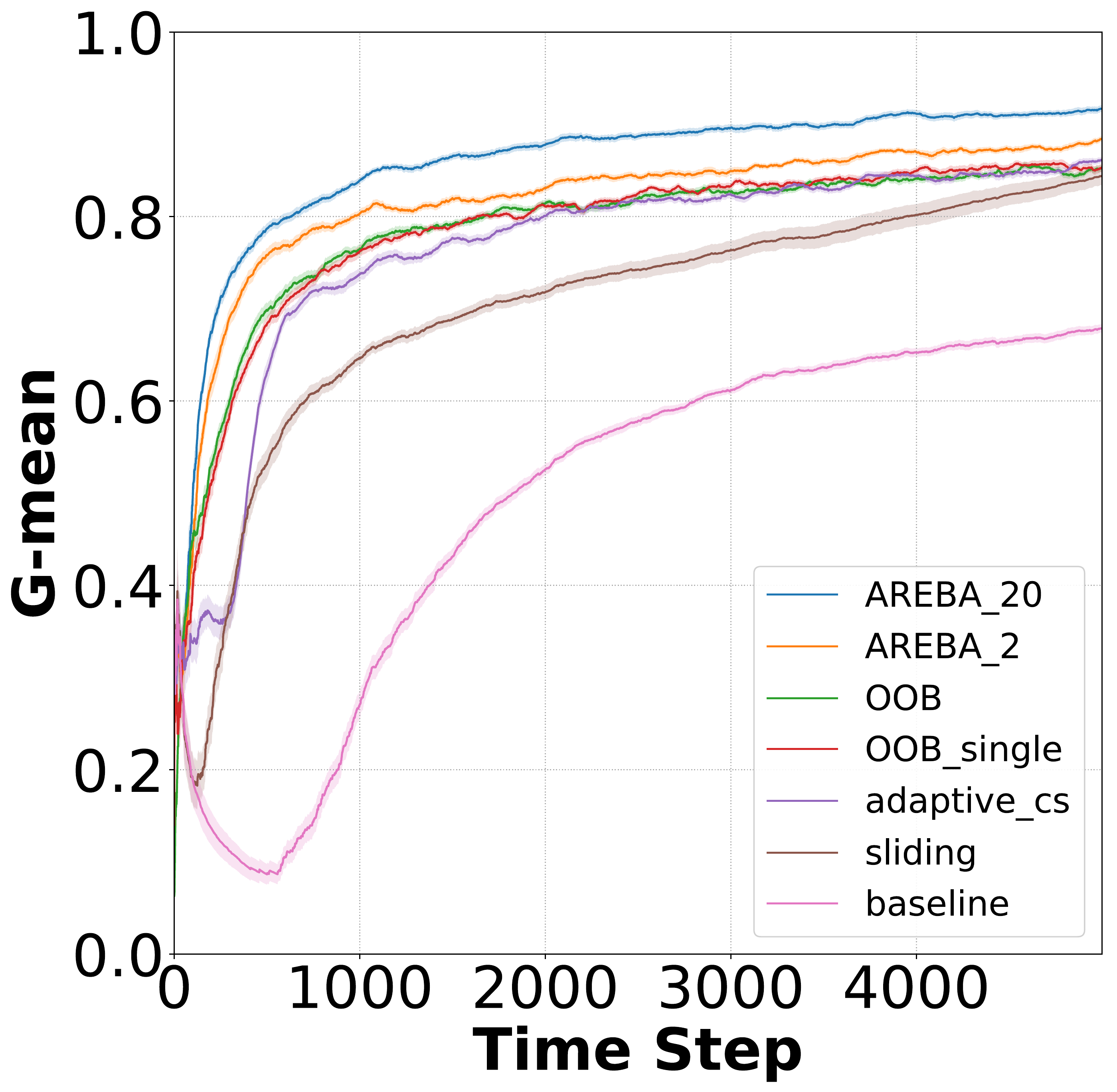}%
		\label{fig:stationary_sine_pp10_gmeans}}
	\subfloat[$Acc^+$]{\includegraphics[scale=0.10]{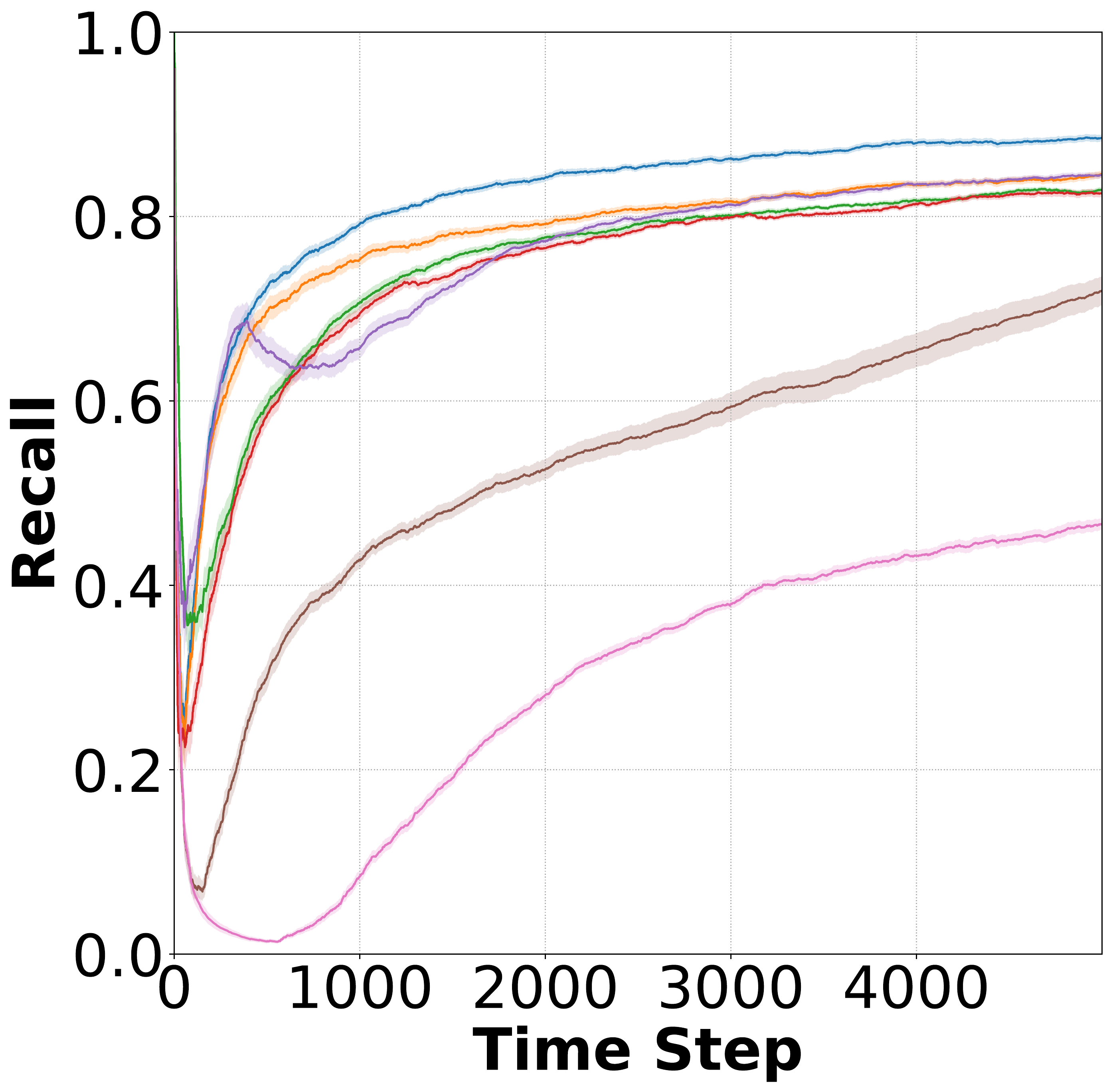}%
		\label{fig:stationary_sine_pp10_recalls}}
	\subfloat[$Acc^-$]{\includegraphics[scale=0.10]{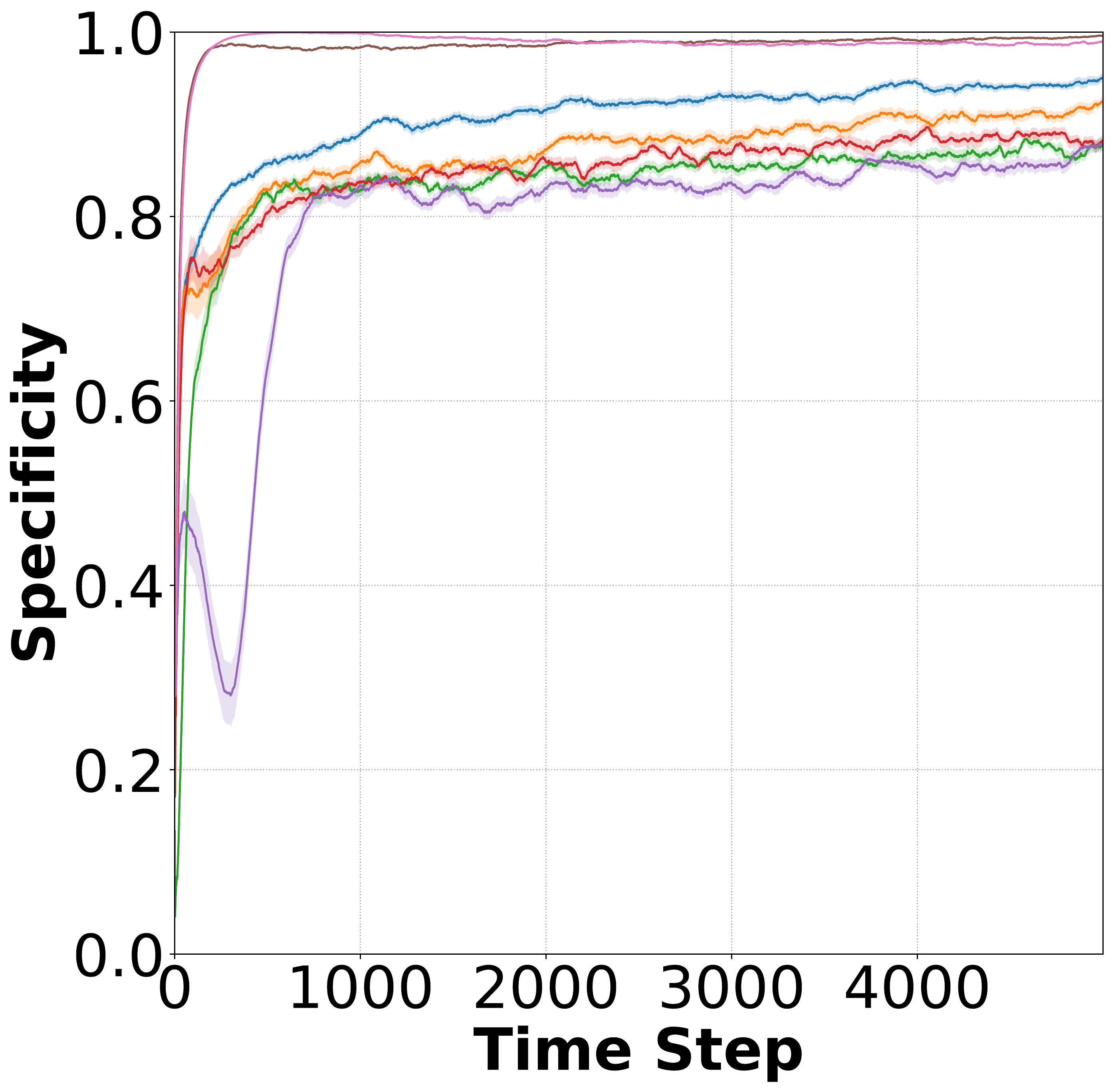}%
		\label{fig:stationary_sine_pp10_specificities}}
	
	\caption{Comparative study on \textit{Sine} (stationary) with mild imbalance 10\%.}
	\label{fig:stationary_sine_pp10}
\end{figure}

\begin{figure}[t!]
	\centering
	
	\subfloat[$G\textnormal{-}mean$]{\includegraphics[scale=0.10]{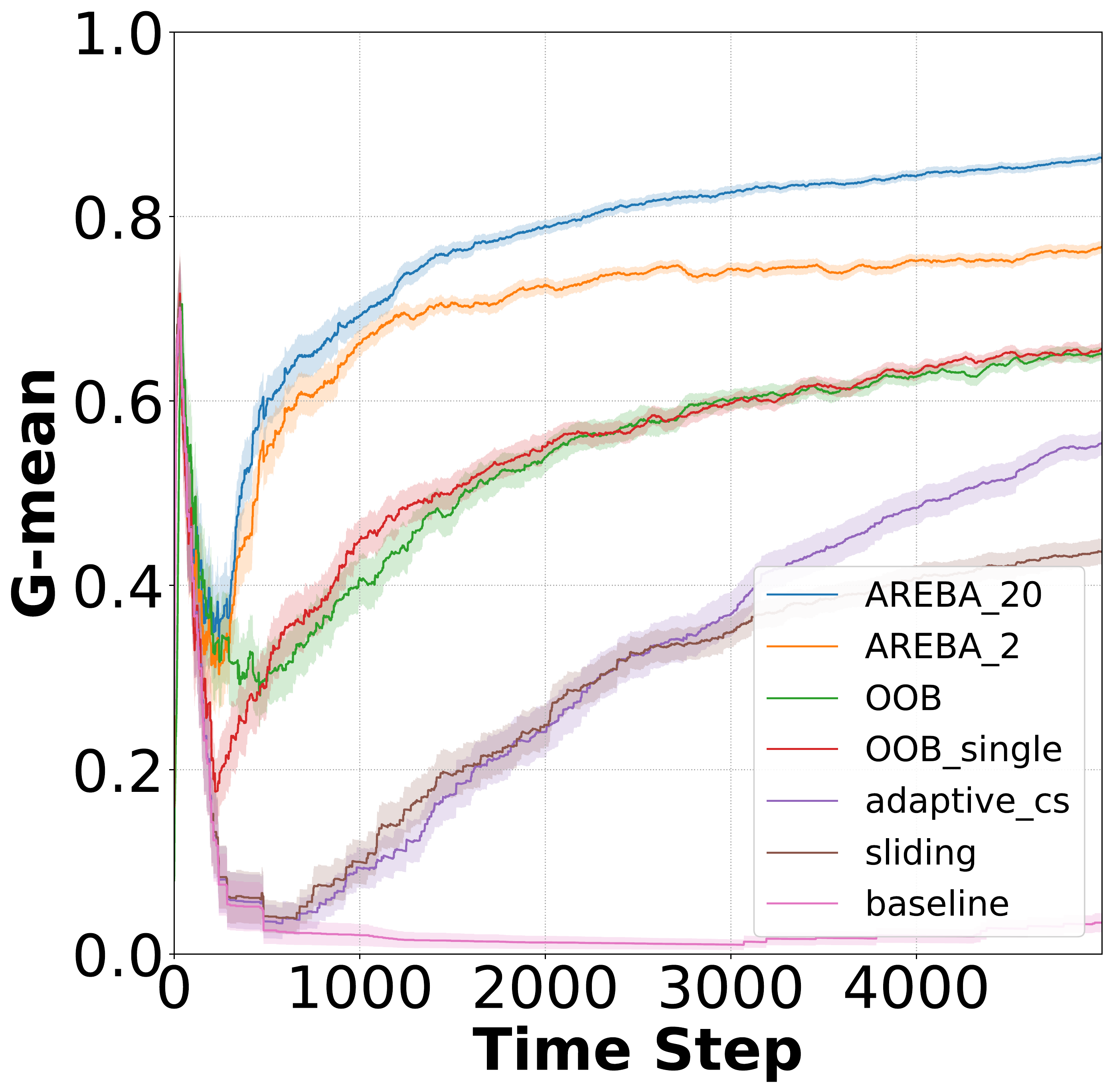}%
		\label{fig:stationary_sine_pp1_gmeans}}
	\subfloat[$Acc^+$]{\includegraphics[scale=0.10]{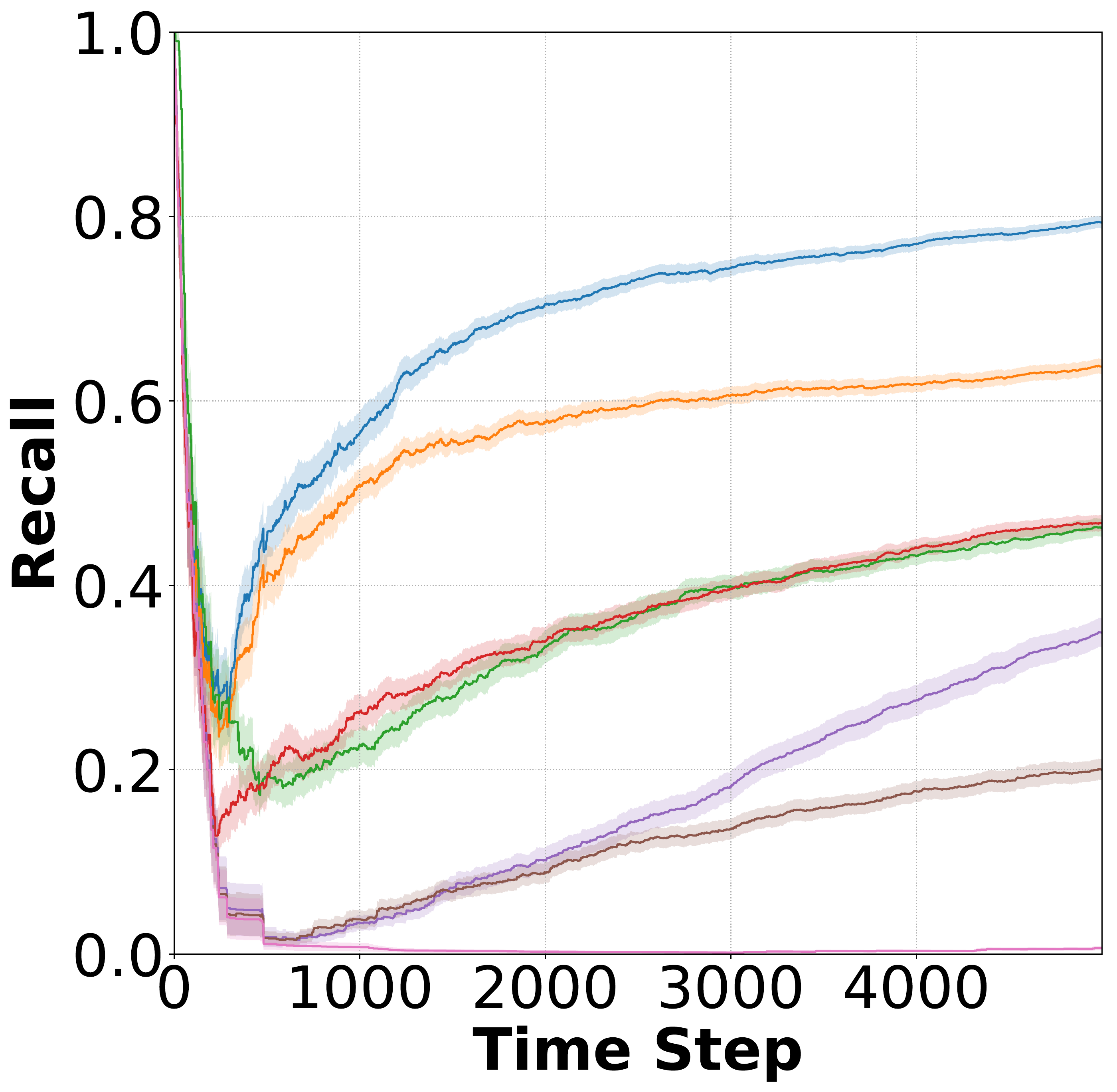}%
		\label{fig:stationary_sine_pp1_recalls}}
	\subfloat[$Acc^-$]{\includegraphics[scale=0.10]{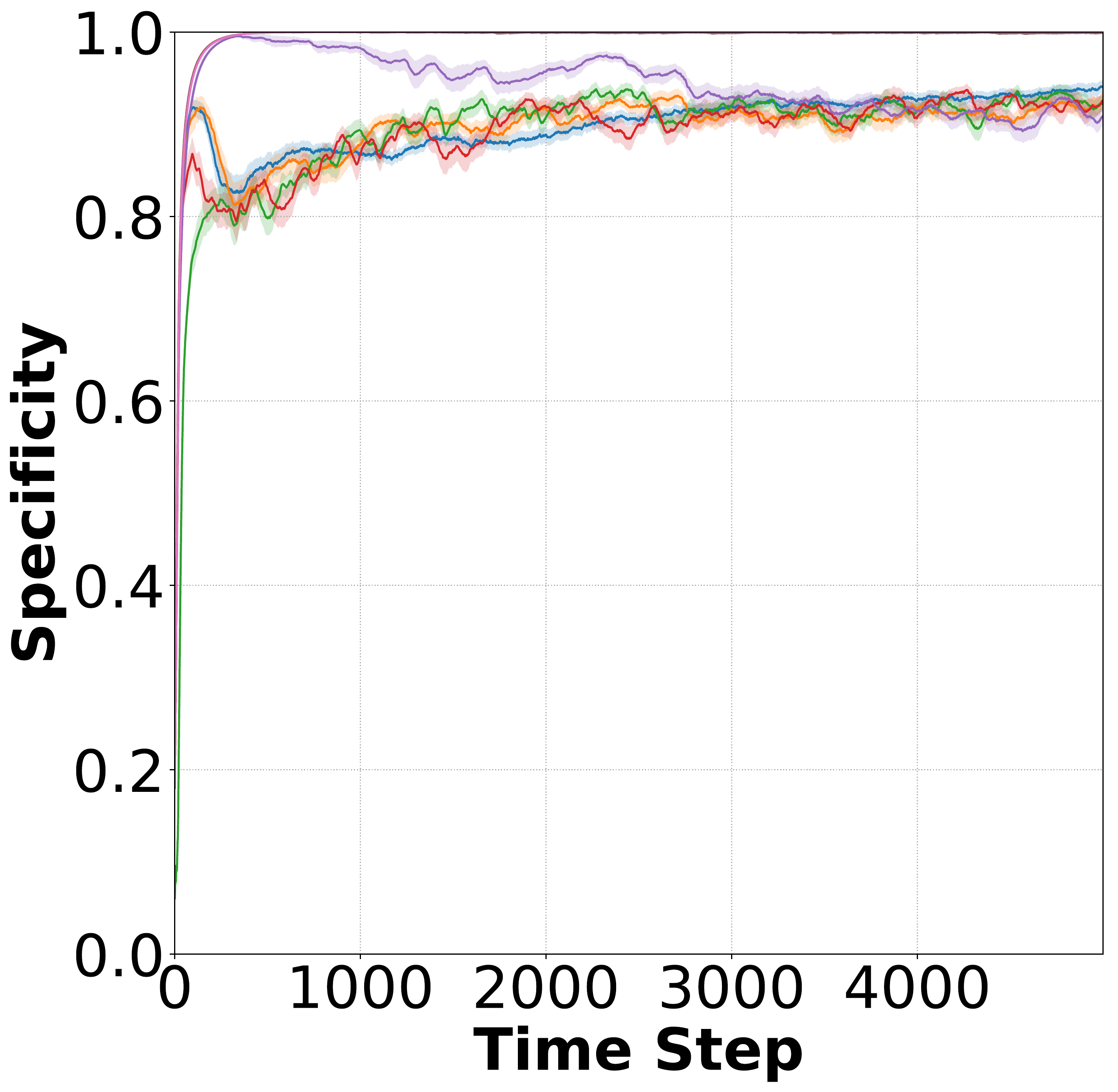}%
		\label{fig:stationary_sine_pp1_specificities}}
	
	\caption{Comparative study on \textit{Sine} (stationary) with severe imbalance 1\%.}
	\label{fig:stationary_sine_pp1}
\end{figure}

\begin{figure}[t!]
	\centering
	
	\subfloat[$G\textnormal{-}mean$]{\includegraphics[scale=0.10]{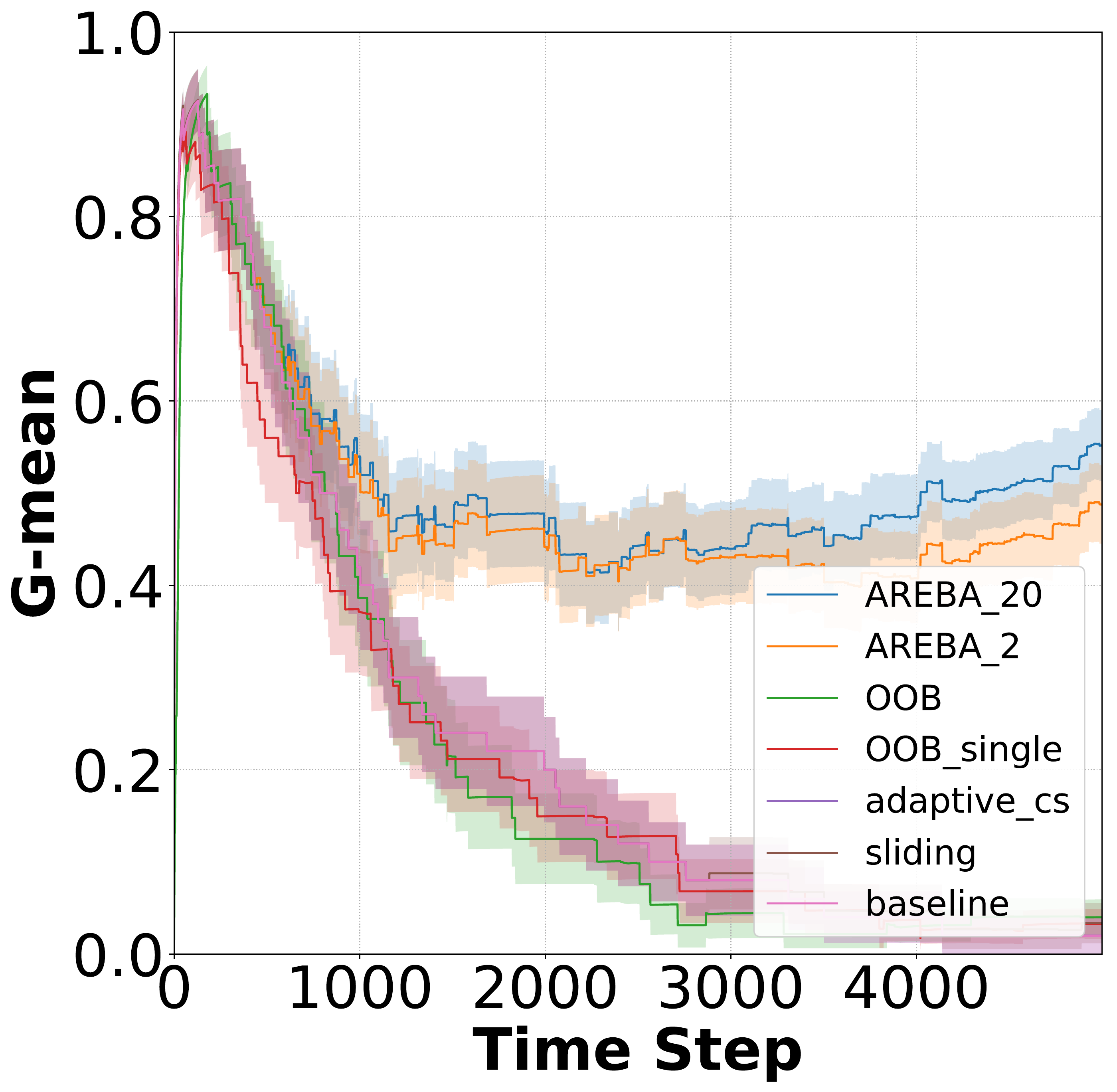}%
		\label{fig:stationary_sine_pp0_gmeans}}
	\subfloat[$Acc^+$]{\includegraphics[scale=0.10]{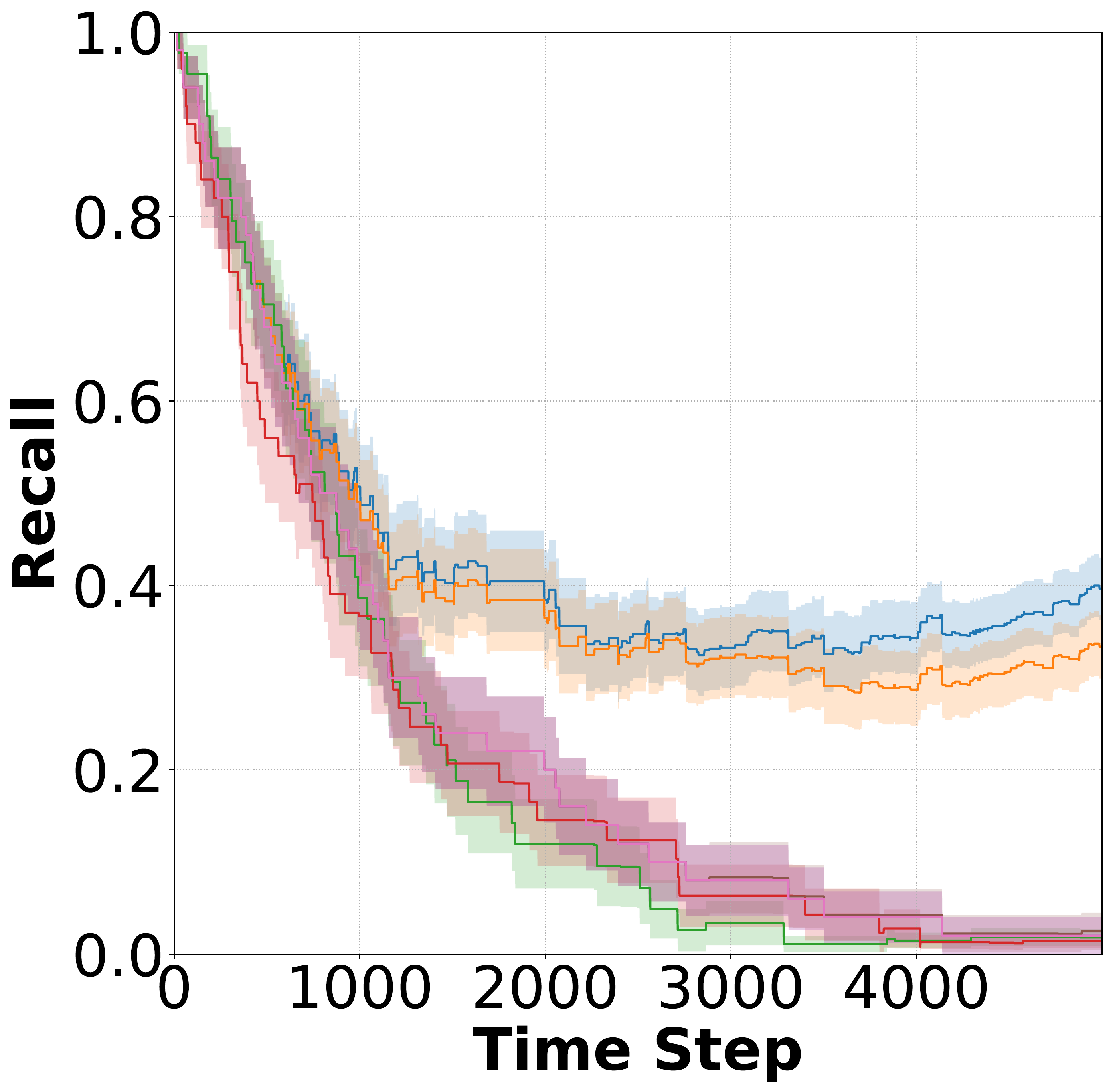}%
		\label{fig:stationary_sine_pp0_recalls}}
	\subfloat[$Acc^-$]{\includegraphics[scale=0.10]{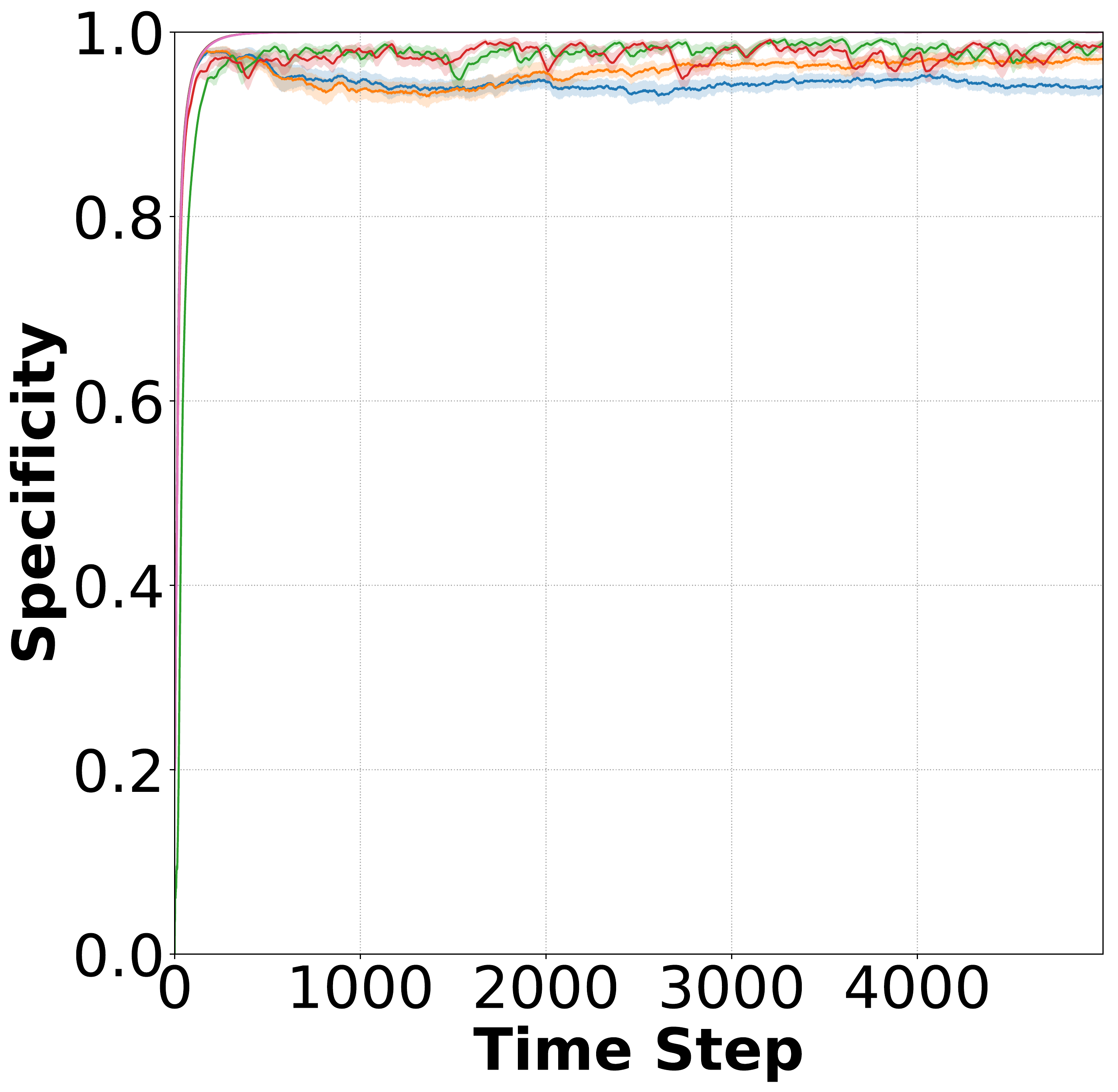}%
		\label{fig:stationary_sine_pp0_specificities}}
	
	\caption{Comparative study on \textit{Sine} (stationary) with extreme imbalance 0.1\%.}
	\label{fig:stationary_sine_pp0}
\end{figure}

We describe our work on stationary synthetic data. Figs~\ref{fig:stationary_sine_pp10} -~\ref{fig:stationary_sine_pp0} show the results for the \textit{Sine} dataset with imbalance of $CI = 10\%, 1\%, 0.1\%$ respectively. For completeness, we also present the prequential recall and specificity. Based on the analysis in Section~\ref{sec:analysis}, we choose $B=20$ for the \textit{Sine} dataset.

 In Fig.~\ref{fig:stationary_sine_pp10_gmeans}, the best performance is achieved by \textit{AREBA\_20}. The rest reach a similar performance with the exception of the \textit{Baseline}. Both \textit{AREBA} versions learn faster with \textit{OOB / OOB\_single} being the second best. Noteworthy, the 20 classifier ensemble (\textit{OOB}) performs similarly to the single classifier \textit{OOB\_single}. While this may seem surprising at first, it is in fact consistent with the results of their authors  \cite{wang2015resampling}, where they have concluded that \textit{resampling}, and not \textit{ensembling}, is the reason behind the effectiveness of the approach.

In Fig.~\ref{fig:stationary_sine_pp1_gmeans} where severe imbalance exists the two \textit{AREBA} versions significantly outperform the rest. Under extreme imbalance (Fig.~\ref{fig:stationary_sine_pp0_gmeans}) \textit{AREBA} performs more than ten times better than the rest. \textit{AREBA}'s effectiveness is attributed to the following. While all algorithms perform well on the majority class (Figs~\ref{fig:stationary_sine_pp1_specificities} and \ref{fig:stationary_sine_pp0_specificities}), \textit{AREBA} has a superior performance on the minority class (Figs~\ref{fig:stationary_sine_pp1_recalls} and \ref{fig:stationary_sine_pp0_recalls}). This means that the rest of the algorithms find it difficult to classify correctly positive examples. Recall that the G-mean is high when both $Acc^+$ and $Acc^-$ are high and when their difference is small.

The reason for \textit{AREBA}'s advantage on minority class examples is due to the concept of maintaining \textit{separate} and \textit{balanced} queues for each class. Let us first consider the incremental learning algorithms \textit{Baseline}, \textit{Adaptive\_CS}, \textit{OOB\_single} and \textit{OOB}. These algorithms use only the most recent arriving example which they later discard. As a result, under severe or extreme imbalance, they don't experience many minority class examples. Let us now consider \textit{Sliding} which is a memory-based algorithm as it implements a single sliding window. Under severe or extreme imbalance, this method may still suffer from the same problem if only a small number of minority class examples are found in its sliding window. The problem can be alleviated with a larger window size, however, it wouldn't be able to rapidly react to concept drift. In contrast, the proposed \textit{AREBA} can afford to have a small window size and, at the same time, experience a sufficient number of minority class examples. \textit{AREBA}, like all methods, depends on the classifier being able to forget old knowledge quickly enough to react to drift. Therefore, as discussed in Section~\ref{sec:role_B}, the optimal value of $B$ is application-dependent and tuning is required. To sum up, these remarks are made:
\begin{itemize}
	\item \textit{AREBA\_20} outperforms all algorithms in all imbalance scenarios while \textit{AREBA\_2} is the second best. Under extreme class imbalance, \textit{AREBA} has been shown to perform ten times better than state-of-the-art algorithms.
		
	\item Interestingly, resampling methods (\textit{AREBA}, \textit{OOB\_single}) seem to better handle online imbalance learning than cost-sensitive learning methods (\textit{Adaptive\_CS}).
	
	\item As the imbalance becomes severe, the performance of all algorithms declines significantly. \textit{AREBA} has been shown to be affected less seriously, thus being more robust to it.
	
	\item Methods without a mechanism to handle class imbalance in online learning perform poorly (\textit{Baseline, Sliding}).
\end{itemize}

\subsection{Nonstationary data}\label{sec:nonstationary}
We describe our work on nonstationary synthetic data. To examine each drift type independently, we devise the following experiments on the \textit{Sine} dataset based on \cite{wang2018systematic}. In all experiments, the imbalance is set to $CI = 1\%$ i.e. $p(y=1)=0.01$.

\begin{figure}[t!]
	\centering
	
	\subfloat[prior drift]{\includegraphics[scale=0.10]{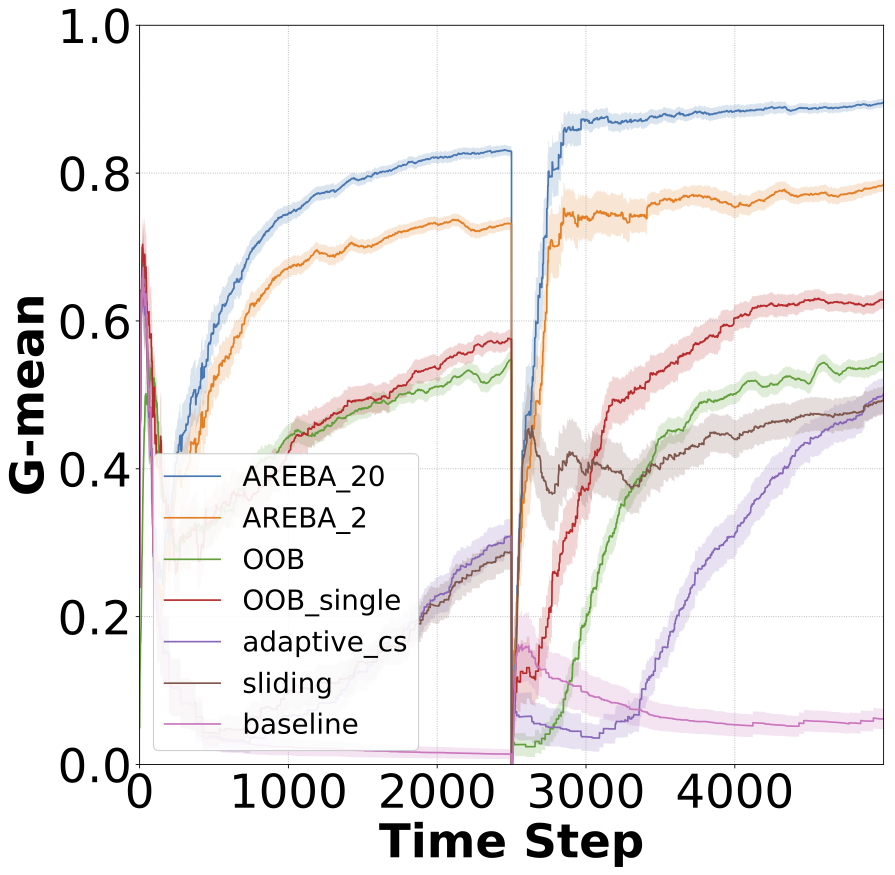}%
		\label{fig:sine_pp1_prior}}
	\subfloat[likelihood drift]{\includegraphics[scale=0.10]{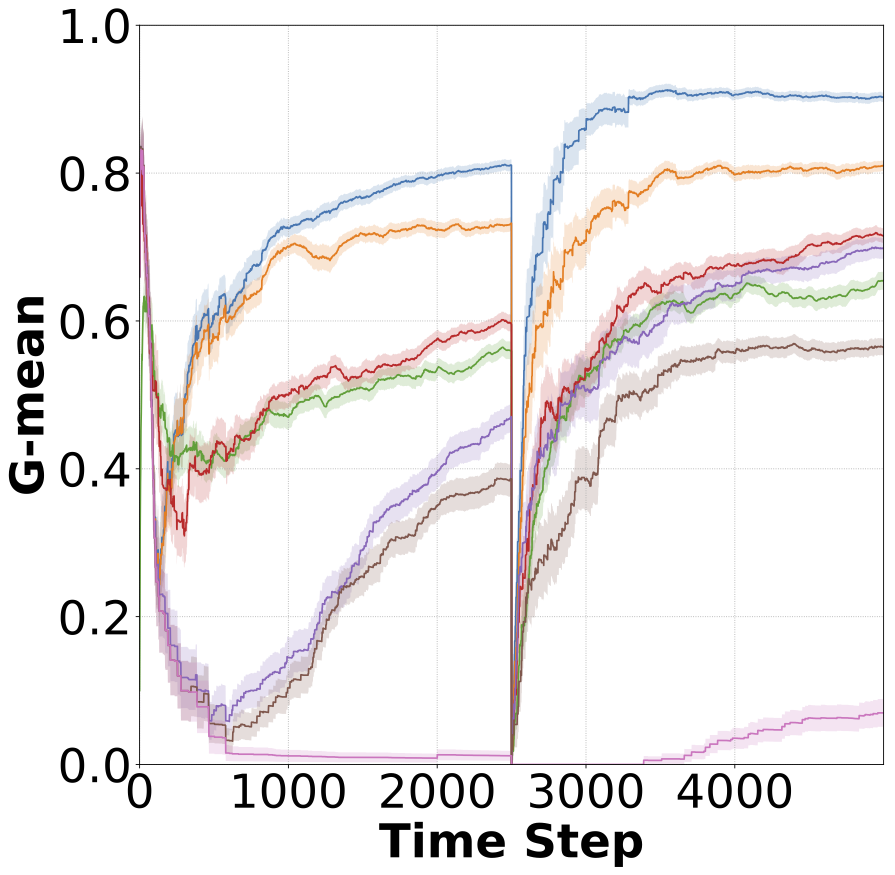}%
		\label{fig:sine_pp1_likelihood}}
	\subfloat[posterior drift]{\includegraphics[scale=0.10]{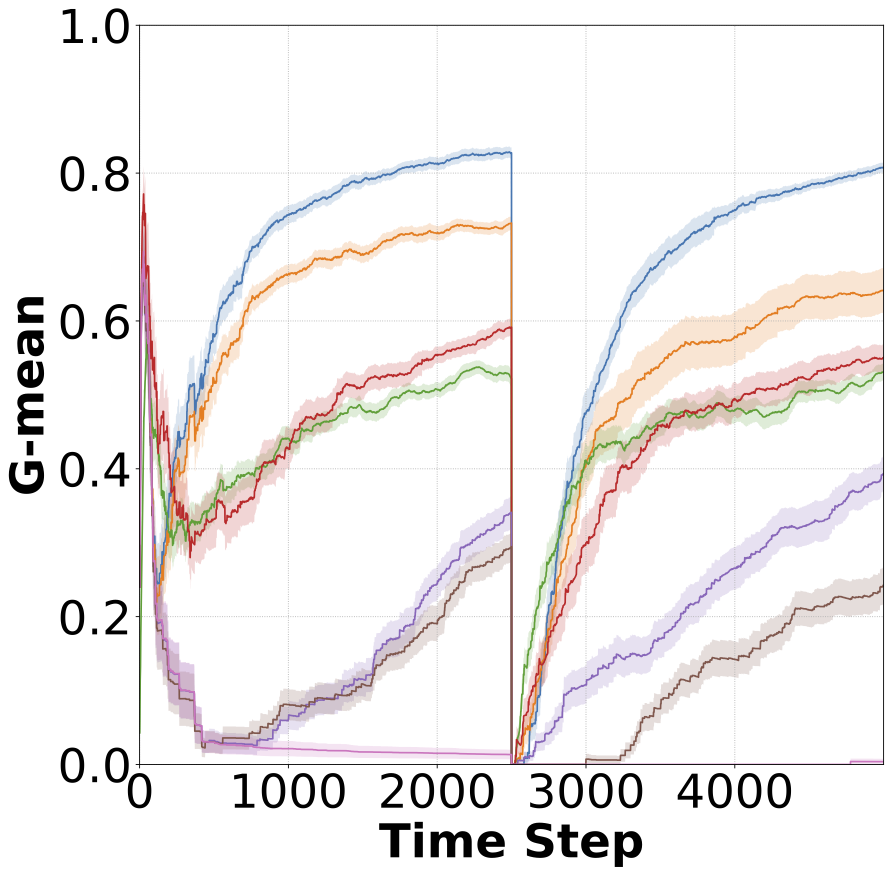}%
		\label{fig:sine_pp1_posterior}}
	
	\caption{Comparative study on \textit{Sine} with imbalance of 1\% and different drift types. \textit{AREBA} is robust to drift, can fully recover and significantly outperforms other state-of-the-art methods as it had been the case before the drift occurred.}
\end{figure}

A drift in prior probability $p(y)$ occurs as shown in Eq~(\ref{eq:prior_exp}):
\begin{equation}\label{eq:prior_exp}
\begin{aligned}
p(y=1) = 0.01 & \longrightarrow 0.99\\
p(y=0) = 0.99 & \longrightarrow 0.01
\end{aligned}
\end{equation}
A drift in likelihood $p(x|y)$ occurs as shown in Eq~(\ref{eq:likelihood_exp}):
\begin{equation}\label{eq:likelihood_exp}
\begin{aligned}
p(x_1 < 0.6 | y=0) = 0.9 & \longrightarrow 0.1\\
p(x_1 \geq 0.6 | y=0) = 0.1 & \longrightarrow 0.9
\end{aligned}
\end{equation}
A posterior probability drift occurs as shown in Eq~(\ref{eq:posterior_sine}). Figures~\ref{fig:sine_pp1_prior} -~\ref{fig:sine_pp1_posterior} show a comparative study on the \textit{Sine} dataset with a prior, class-conditional and posterior probability drift respectively. Overall, the proposed \textit{AREBA\_20} outperforms the rest in all cases while \textit{AREBA\_2} is the second best.

The first point to note is about virtual drift; recall that this drift type does not alter the \textit{true} decision boundary but it can affect the \textit{learnt} one. Generally, the state-of-the-art (\textit{AREBA}, \textit{OOB\_single}, \textit{Adaptive\_CS}) algorithms not only are robust to virtual drift, but an overall slight improvement is observed (Figures \ref{fig:sine_pp1_prior} and \ref{fig:sine_pp1_likelihood}). This is because more feature space is revealed to the algorithm after the drift occurs, for instance, in Eq~(\ref{eq:likelihood_exp}) more examples with $x_1 \geq 0.6$ will be observed.

The second point is about posterior drift. In this case the performance of all algorithms decline significantly after the drift (Fig. \ref{fig:sine_pp1_posterior}). Recall from Eq~(\ref{eq:posterior_sine}) that the way we defined posterior drift is by performing a ``concept swap''. Therefore, all algorithms should start re-learning the new concept.

The third point is about \textit{OOb\_single}. It outperforms its ensemble version when drift is virtual (Figs.~\ref{fig:sine_pp1_prior} and \ref{fig:sine_pp1_likelihood}). After a posterior drift occurs, \textit{OOB} is significantly better than \textit{OOB\_single} but they obtain a similar final performance. Again, this is in alignment with \cite{wang2015resampling} as in a number of occasions, the single classifier has outperformed its ensemble version. From now on, we will consider only \textit{OOB\_single}. To sum up, the following important remarks can be made:
\begin{itemize}
	\item \textit{AREBA\_20} outperforms all algorithms in all drift scenarios while \textit{AREBA\_2} is the second best.
	
	\item A drift in $p(y|x)$ is the most severe type of data alteration and this clearly reflects on all algorithms' performance.
	
	\item Algorithms with a mechanism to address drift (e.g. \textit{Sliding}) but without one to address imbalance perform poorly.
	
	\item In settings where there is solely drift (no imbalance), \textit{AREBA\_B} and \textit{Sliding} significantly outperform all algorithms; they are almost identical when $W=2B$. \textit{AREBA\_2} and \textit{Baseline} jointly follow. \textit{Adaptive\_CS} and \textit{OOB\_single} are outperformed by the rest (the figures for these results are not included).
\end{itemize}

\begin{figure}[t!]
	\centering
	
	\subfloat[prior drift]{\includegraphics[scale=0.10]{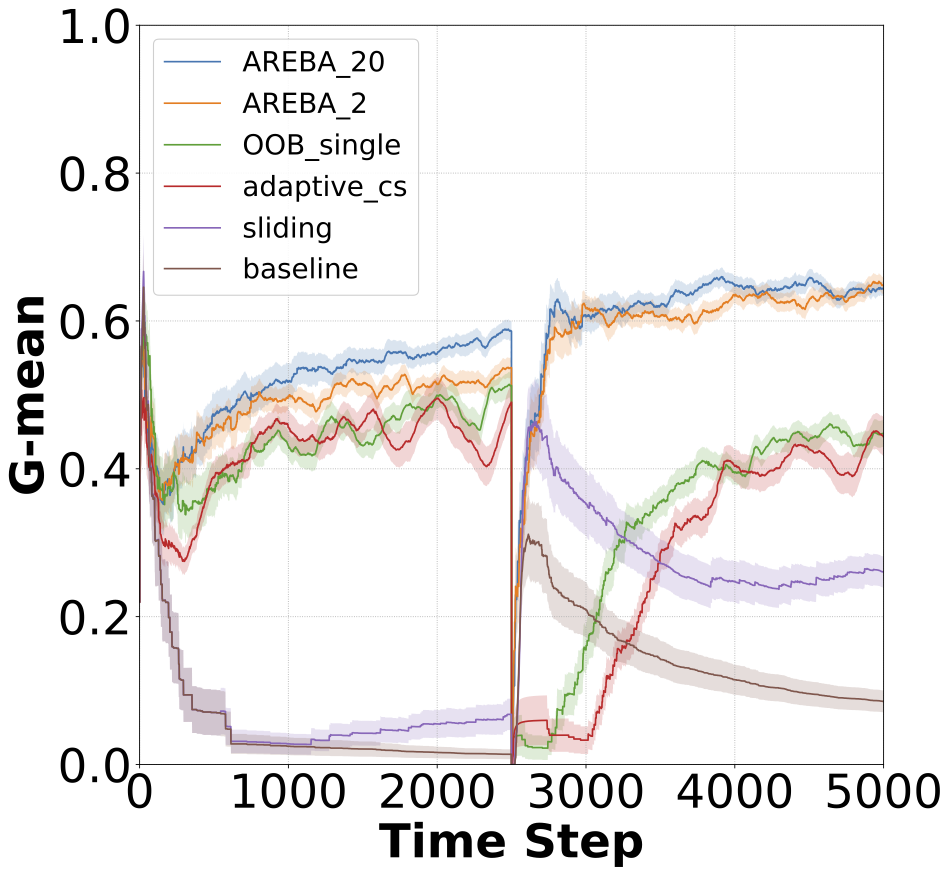}%
		\label{fig:sine_pp1_prior_noise}}
	\subfloat[likelihood drift]{\includegraphics[scale=0.10]{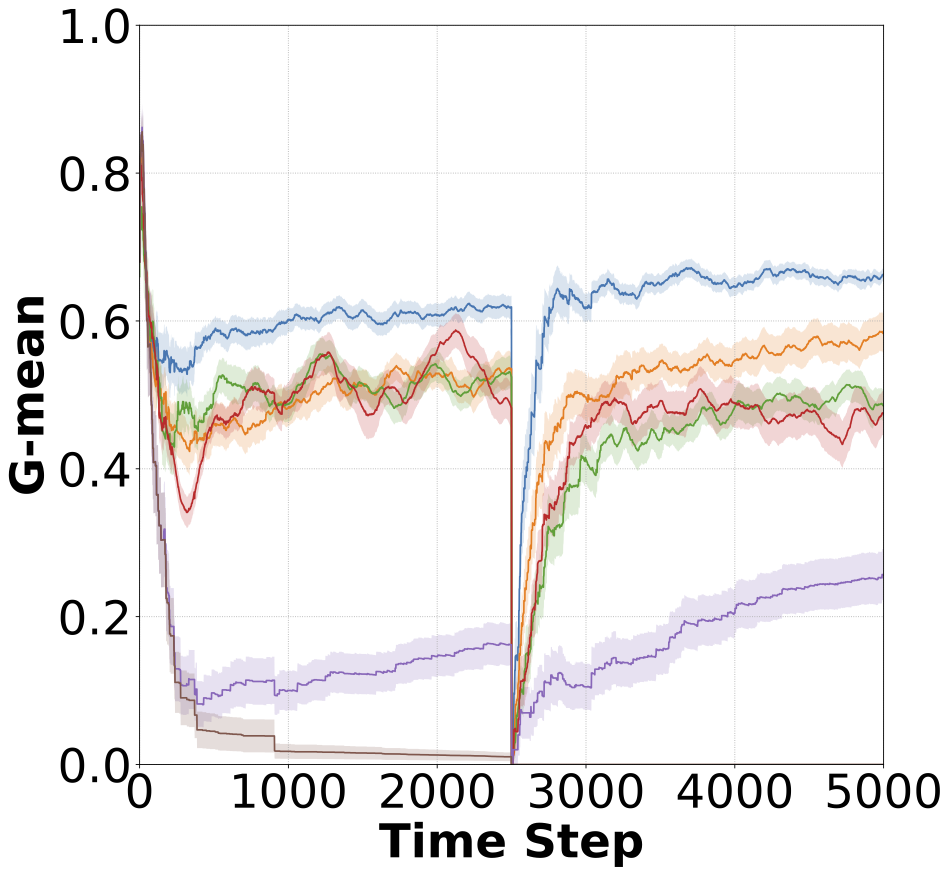}%
		\label{fig:sine_pp1_likelihood_noise}}
	\subfloat[posterior drift]{\includegraphics[scale=0.10]{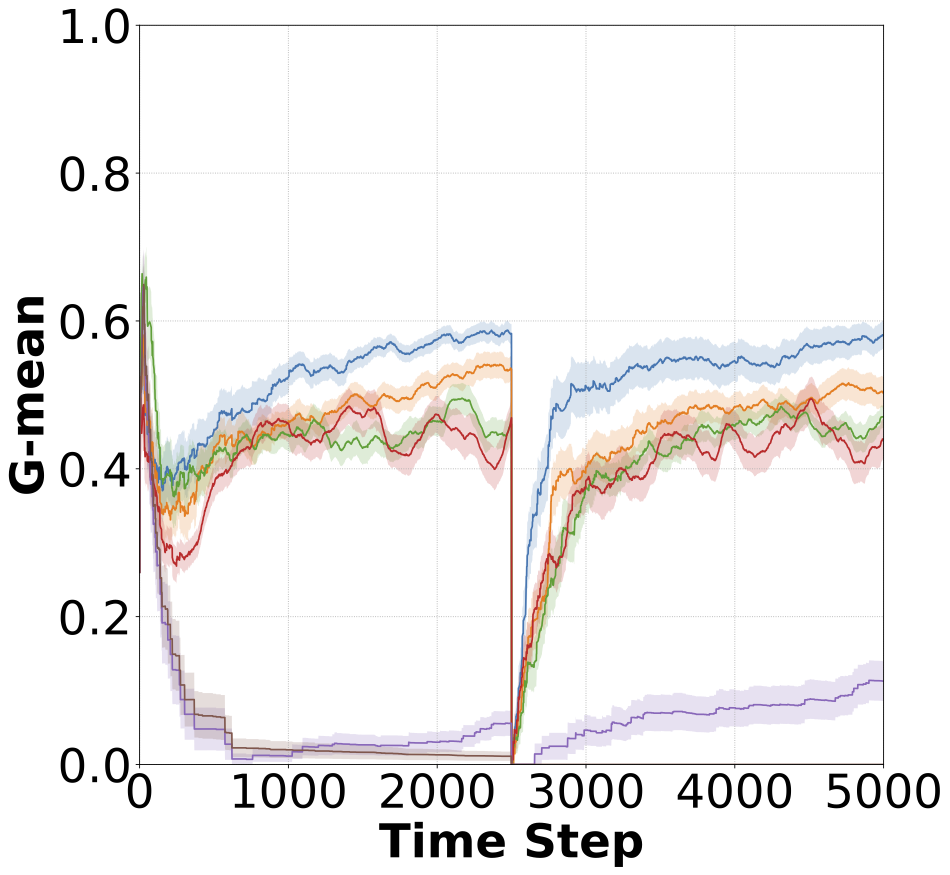}%
		\label{fig:sine_pp1_posterior_noise}}
	
	\caption{Comparative study on the \textit{Sine} dataset with class imbalance of $CI=1\%$ and $10\%$ noise in the class labels.}
\end{figure}

\begin{figure*}[t!]
	\centering
	
	\subfloat[\textit{Cervical Cancer}]{\includegraphics[scale=0.11]{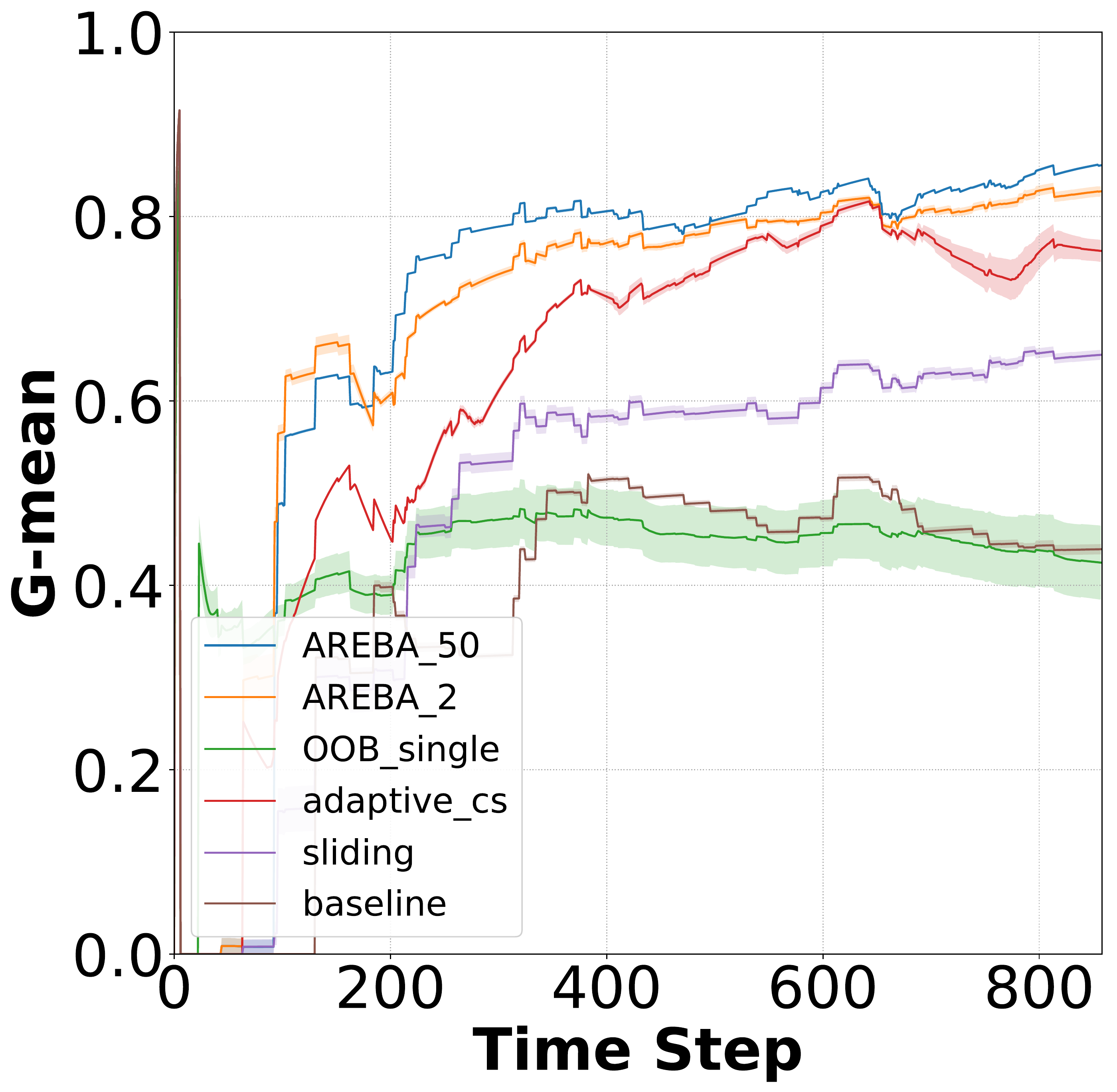}%
		\label{fig:real_cervical_cancer}}
	\subfloat[\textit{Fraud}]{\includegraphics[scale=0.11]{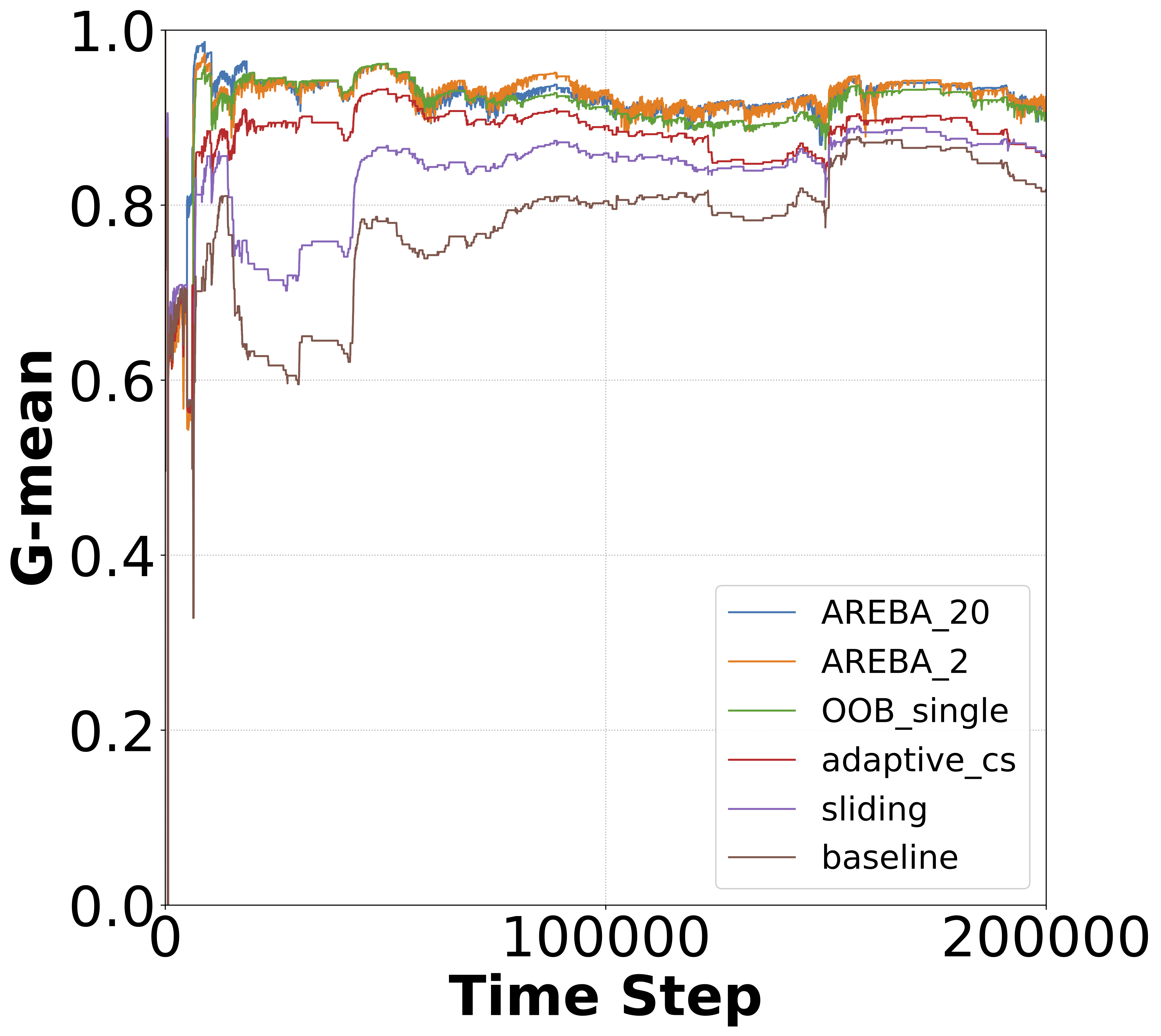}%
		\label{fig:real_fraud}}
	\subfloat[\textit{Credit Score}]{\includegraphics[scale=0.11]{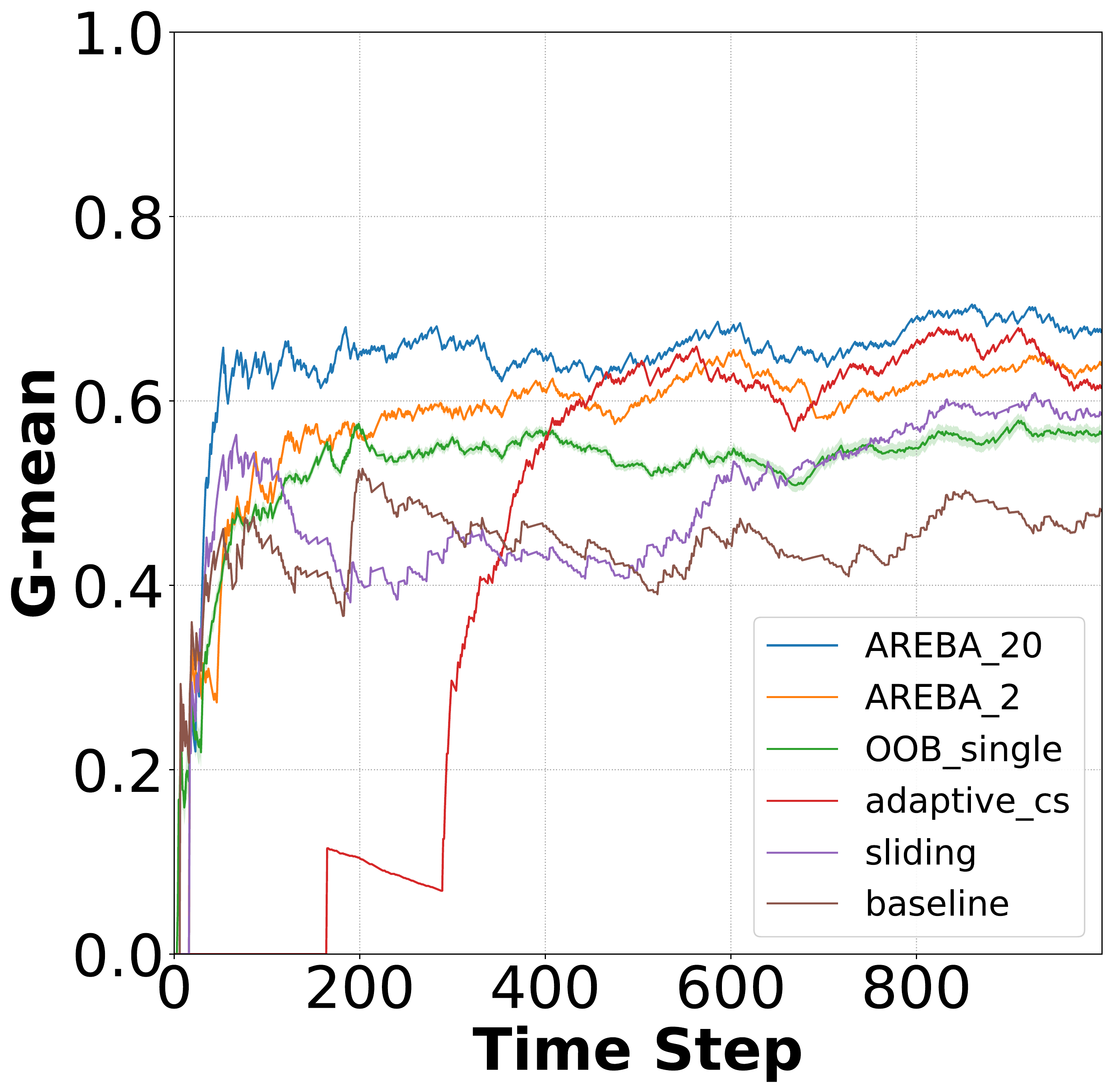}%
		\label{fig:real_credit_score}}
	\subfloat[\textit{MNIST}]{\includegraphics[scale=0.11]{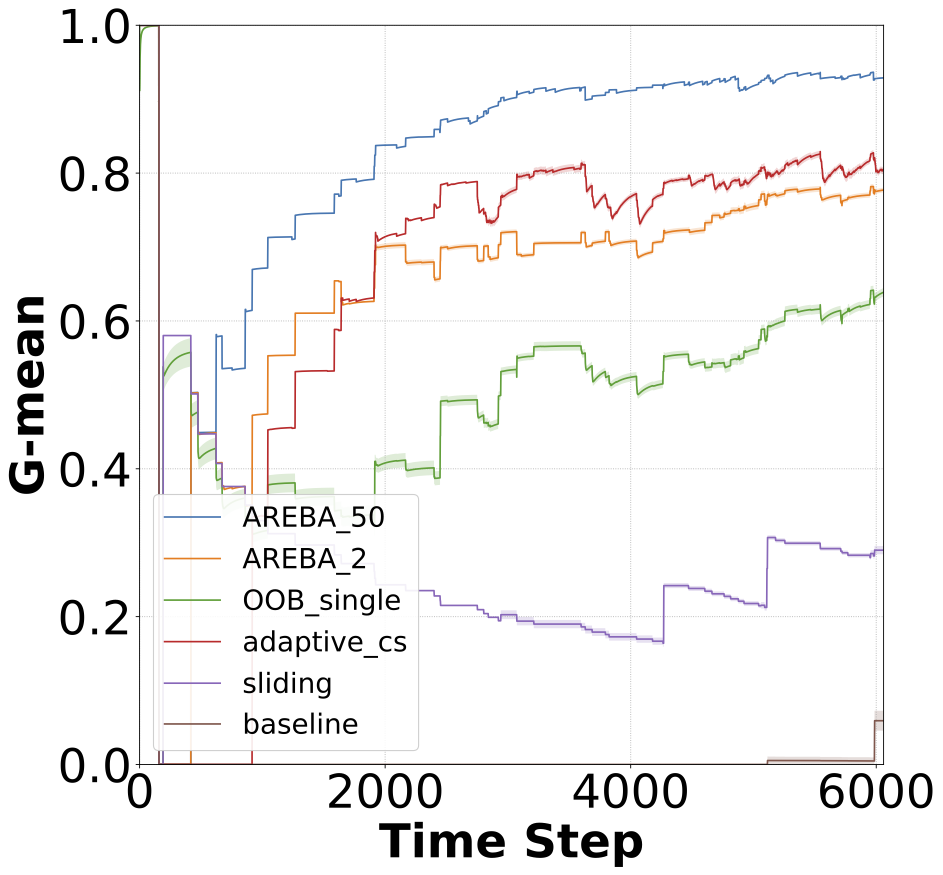}%
		\label{fig:real_mnist}}
	\subfloat[\textit{Forest Type}]{\includegraphics[scale=0.11]{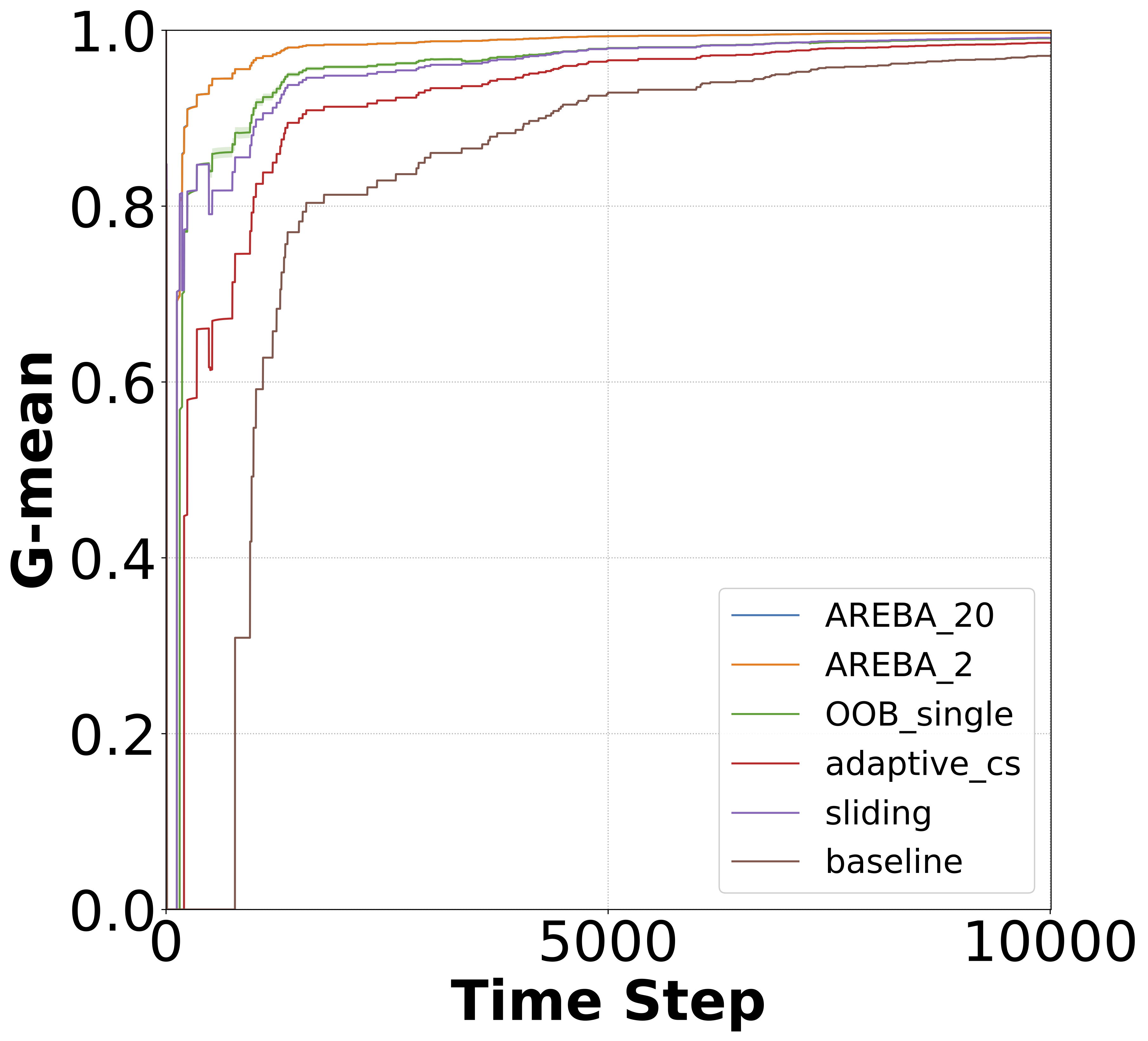}%
		\label{fig:real_forest}}

	\caption{Comparative study on the real-world datasets. \textit{AREBA} is the best (or joint best) performing algorithm.}
\end{figure*}

\begin{table*}[h!]
	\renewcommand{\arraystretch}{1.3}
	\caption{Final Performance (G-mean) on the real-world datasets displaying the mean and standard deviation}
	\label{tab:real}
	\centering
	\begin{tabular}{| c || c || c || c || c || c |}
		\hline
		\bfseries Algorithm & \bfseries Cervical Cancer & \bfseries Fraud & \bfseries Credit Score & \bfseries MNIST & \bfseries Forest Cover Type\\
		\hline\hline
		\textit{AREBA\_20/50} & \bfseries 0.8555 (0.0104) & 0.8900 (0.0038) & \bfseries 0.6746 (0.0052) & \bfseries 0.9289 (0.0047) & 1.0000 (0.0000) \\
		\hline
		\textit{AREBA\_2} & \bfseries 0.8272 (0.0396) & \bfseries 0.8985 (0.0000) & 0.6392 (0.0095) & 0.7771 (0.0174) & 1.0000 (0.0000) \\
		\hline
		\textit{OOB\_single} & 0.4246 (0.2842) & 0.8968 (0.0019) & 0.5645 (0.061) & 0.6388 (0.0313) & 1.0000 (0.0000) \\
		\hline
		\textit{Adaptive\_CS} & 0.7627 (0.0848) & 0.8564 (0.0000) & 0.6141 (0.0023) & 0.8051 (0.0252) & 1.0000 (0.0000) \\
		\hline
		\textit{Sliding} & 0.6499 (0.0383) & 0.8538 (0.0000) & 0.5851 (0.0024) & 0.2899 (0.0305) & 1.0000 (0.0000) \\
		\hline
		\textit{Baseline} & 0.4391 (0.0394) & 0.8233 (0.0000) & 0.4807 (0.0158) & 0.0592 (0.0737) & 1.0000 (0.0000) \\
		\hline
	\end{tabular}
\end{table*}

\subsection{Data with noisy class labels}\label{sec:noise}
We study each algorithm under conditions where the class label (i.e. ground truth) is incorrectly provided. The study aims at emulating realistic scenarios that can be encountered in deployed settings. One such scenario is when an expert is unable to provide the label (violation of the \textit{label availability} assumption). Another scenario is when the label can be provided, but not always timely i.e. before the arrival of the next example (violation of the \textit{no verification latency} assumption). In these scenarios, an \textit{estimated} label could be provided by an automated mechanism which may or may not be true.

To model this behaviour, we specify a probability threshold by which the true label \textit{cannot} be provided.  The degree to which the aforementioned assumptions are violated in our experiments i.e. the probability threshold is $10\%$.  Specifically, with a probability of $10\%$ we revert the true class label of the arriving example, therefore, each algorithm is not always trained using the ground truth. We repeat the experiments of Figures~\ref{fig:sine_pp1_prior} -~\ref{fig:sine_pp1_posterior} with the added noise and present the results in Figures~\ref{fig:sine_pp1_prior_noise} -~\ref{fig:sine_pp1_posterior_noise}. Similar results for the other synthetic datasets were obtained and are included in the supplementary materials. These important remarks can be made:
\begin{itemize}
	\item In the presence of noise in the class labels, the performance of all algorithms declines substantially. 
	\item \textit{AREBA\_20} outperforms the rest in all drift scenarios.
	\item Compared to \textit{AREBA\_2}, the performance gap is no longer big; as $B$ gets large, more noisy examples are included in the queues. Interestingly, however, $B=2$ does not yield the best results. Hence, smaller values of $B$ are suggested but still some experimentation is necessary.
	\item Interestingly, the performance of \textit{AREBA\_20} in all drift scenarios with $10\%$ noise and $1\%$ (severe) imbalance is $G\textnormal{-}mean \simeq 0.6$, which is close to the performance obtained by the proposed algorithm in the case of $0.1\%$ (extreme) imbalance without drift (Figure~\ref{fig:stationary_sine_pp0_gmeans}).
\end{itemize}

\subsection{Real-world data}\label{sec:real}
We describe now our work on real-world data. To examine the learning speed we present the learning curves in Figures~\ref{fig:real_cervical_cancer} -~\ref{fig:real_forest}. The final mean performances are shown in Table~\ref{tab:real}; the best performing algorithm based on ANOVA and its post-hoc tests (Section~\ref{sec:evaluation}) is shown in bold font which denotes statistical significance over the others, and the standard deviation is shown in brackets. The tables with the p-values are found in the supplementary materials. Following the guidelines derived from our analysis in Section~\ref{sec:role_areba} we use \textit{AREBA\_20} in three datasets and \textit{AREBA\_50} in the other two.

In Fig~\ref{fig:real_cervical_cancer}, \textit{AREBA\_50} achieves $G\text{-}mean = 0.8$ at $t \approx 350$ while \textit{Adaptive\_CS} obtains this score at $t \approx 650$. The learning speed is of utmost importance as, in practise, this means \textit{Adaptive\_CS} would observe about 300 more biopsies to equalise \textit{AREBA}. In Fig~\ref{fig:real_credit_score}, \textit{AREBA\_20} achieves $G\text{-}mean = 0.6$ at $t \approx 50$, while \textit{Adaptive\_CS} obtains this score at $t \approx 450$. In Fig~\ref{fig:real_forest} all algorithms equalise \textit{AREBA\_20} at $t \approx 10000$. In Fig~\ref{fig:real_fraud}, \textit{AREBA} and \textit{OOB\_single} behave similarly.

On Table~\ref{tab:real} and \textit{Cervical\_Cancer}, \textit{AREBA\_20/50} is joint first with \textit{AREBA\_2} and outperforms the second \textit{Adaptive\_CS} by more than $9\%$. In \textit{Credit\_Score}, \textit{AREBA\_20/50} outperforms the second \textit{AREBA\_2} by $3.5\%$ and the third \textit{Adaptive\_CS} by $6\%$. In \textit{MNIST}, \textit{AREBA\_20/50} achieves a superior performance, outperforming the second \textit{Adaptive\_CS} by more than $12\%$ and the third \textit{AREBA\_2} by more than $15\%$. In \textit{Fraud}, \textit{AREBA\_2} outperforms the rest but the improvement over \textit{AREBA\_20/50} and \textit{OOB\_single} is less than $1\%$. In \textit{Forest Cover Type}, all algorithms obtain the same final performance.  Surprisingly, \textit{Adaptive\_CS} performs better overall than \textit{OOB\_single} while the opposite hold true in the synthetic datasets. To sum up, these important remarks can be made:
\begin{itemize}
	\item \textit{AREBA} outperforms other algorithms in all real datasets. These are cases where it would have an impact in practise.
	\item In regard to the choice of memory size $B$, the results are consistent with those observed in our studies with synthetic data. Specifically, \textit{AREBA\_2} either performs similarly to \textit{AREBA\_20} or closely but worse.
\end{itemize}

\section{Discussion and Conclusion}\label{sec:discussion}
We discuss below important aspects of \textit{AREBA}, its advantages and limitations, and directions for future work.

\textbf{Dual nature}. \textit{AREBA}'s effectiveness is attributed to a few important characteristics. By maintaining \textit{separate} and \textit{balanced} queues for each class helps to address the imbalance problem. Propagating past examples in the most recent training set is viewed as a form of oversampling. The fact that examples are carried over a series of steps allows the classifier to ``remember'' old concepts. Also, to address the drift challenge, the classifier needs to also be able to ``forget'' old concepts. This is achieved by \textit{AREBA}'s memory-based nature i.e. by bounding the length of queues, these are essentially behaving like sliding windows. Hence, \textit{AREBA} can cope with both imbalance and drift. We have shown that the proposed synergy is seamless, and it significantly outperforms algorithms that belong to resampling or memory-based methods solely, and even algorithms that belong to other types e.g. cost-sensitive methods. Lastly, recall that no drift detector can perform satisfactory under any situation. When domain expertise can foresee the drift's nature, \textit{AREBA} can work in cooperation with change detection-based methods to complement each other.

\textbf{\textit{QBR} vs \textit{AREBA}}. \textit{QBR} was first introduced in our preliminary study \cite{malialis2018queue} that ran experiments only on synthetic datasets, and examined only a limited range of imbalance rates and a single drift type. In this work, we conducted an extensive experimental work and stress tested \textit{QBR} under conditions of severe imbalance and different drift types. We have identified, discussed and analysed \textit{QBR}'s limitations and under which conditions these occur. It turns out that \textit{QBR}'s limitations arise from the fact that the imbalance problem persists in the queues (as described with reference to Fig.~\ref{fig:qbr}). To overcome this, the paper proposes \textit{AREBA} with two major design changes over \textit{QBR}. Firstly, it allows each queue to be of size $B$ (rather than $\frac{B}{2}$). Secondly, the queues remain balanced throughout the time using a dynamic mechanism (adaptive rebalancing). These changes allow \textit{AREBA} to be very effective when dealing with online learning tasks under drift and imbalance.

\textbf{Role of the memory size.} $B$'s role is of great importance and goes beyond that of just controlling the maximum queue lengths. It controls the ``level'' of \textit{incremental} learning. The smaller its value the closer to being incremental, specifically, when $B=2$ \textit{AREBA} becomes \textit{near-incremental}. Finding a suitable value is dataset-specific and requires some trial-and-error. To reduce this tedious process, we have provided in Section~\ref{sec:role_B} some guidelines to help us determine $B$.

\textbf{Choice of classifier.} \textit{AREBA} does not impose any restrictions on the selection of the classifier. Some classifiers, however, are more suitable than others for online learning. Our scope was on neural networks which have been shown to work well (e.g. \cite{wang2014cost, wang2018systematic}). Hoeffding Trees have also been shown to work well (e.g. \cite{brzezinski2013reacting}). Future work will apply \textit{AREBA} with trees to examine if the observed improvement can generalise.

\textbf{Verification latency.} The learning framework used is suitable for \textit{human-in-the-loop} learning and assumes that no verification latency exists. Involving humans, however, may cause delays in receiving the labels. In practise, to avoid or reduce potential delays, would require mechanisms to collect labels in an automatic manner. We have shown in Section~\ref{sec:noise} that \textit{AREBA} maintains its dual nature benefits and still outperforms other state-of-the-art algorithms in conditions where the assumption is violated. Future work will relax this assumption and examine other paradigms, such as, active learning \cite{malialis2020data}.

To conclude, we introduced the novel \textit{Adaptive REBAlancing (AREBA)} algorithm to addresses the problem of class imbalance in nonstationary environments. We provided new interesting insights towards the \textit{joint problem} of imbalance and concept drift. Our study compared \textit{AREBA} to other four baseline and state-of-the-art algorithms and showed that it significantly outperforms them in the vast majority of compared settings.

% or


%\appendices
%% you can choose not to have a title for an appendix
%% if you want by leaving the argument blank
%\section{}

\section*{Acknowledgment}
This work has been supported by the EU's Horizon 2020 research and innovation programme under grant agreements No 867433 (Fault-Learning) and No 739551 (KIOS CoE), and from the Republic of Cyprus through the Directorate General for European Programmes, Coordination and Development.

\ifCLASSOPTIONcaptionsoff
  \newpage
\fi



\bibliographystyle{IEEEtran}
\bibliography{areba}
%\begin{thebibliography}{1}
%\end{thebibliography}


%\newpage

\begin{IEEEbiography}[{\includegraphics[width=1in,height=1.25in,clip,keepaspectratio]{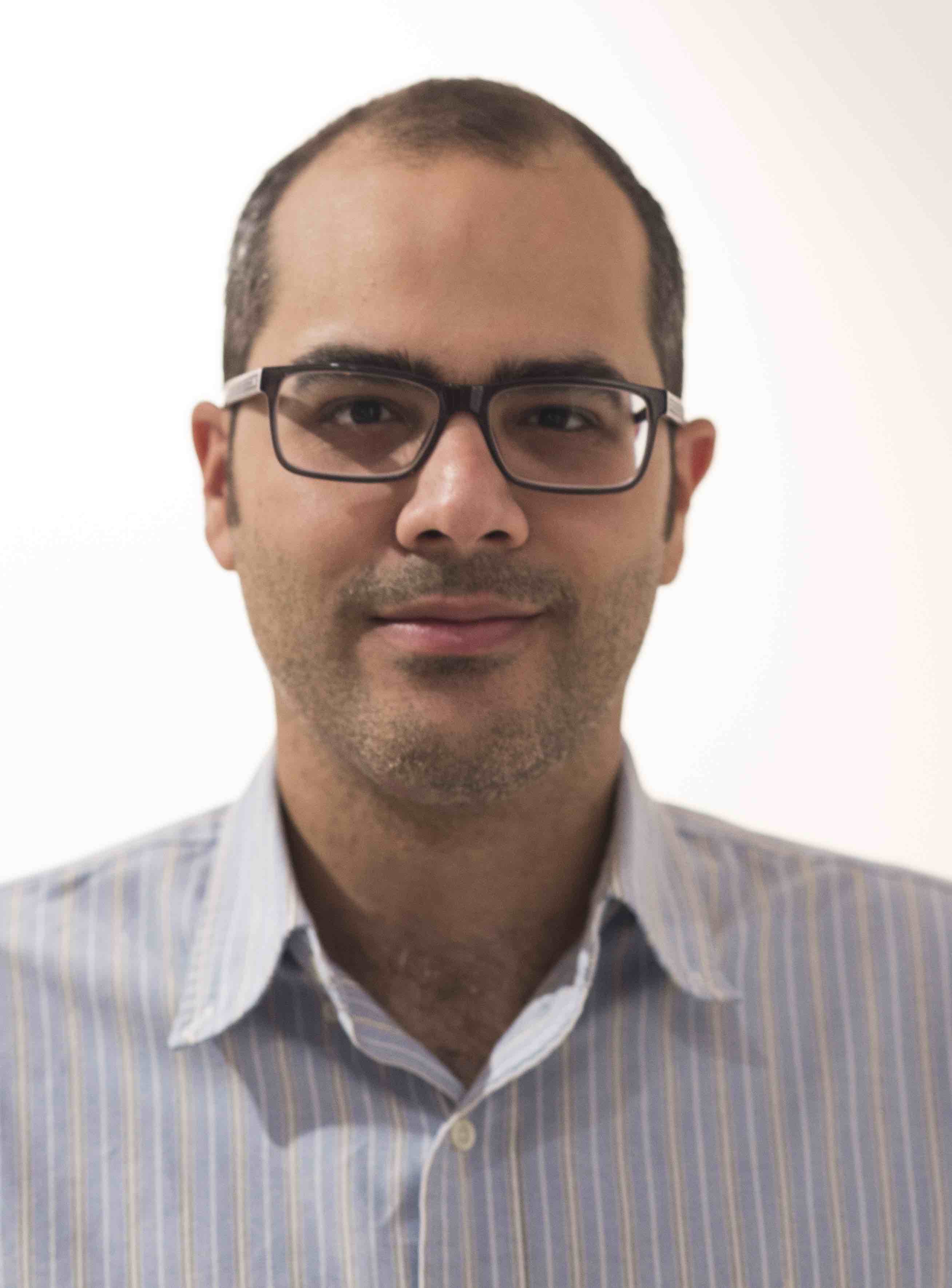}}]%
	{Kleanthis Malialis} is a Marie Skłodowska-Curie Widening Fellow in the KIOS Center of Excellence (KIOS CoE) at the University of Cyprus (UCY) interested in online machine learning. Previously, he was a Research Associate in the KIOS CoE. Prior joining UCY, Dr. Malialis was working at The Telegraph as a Data Scientist where his focus was on building predictive models using machine learning algorithms. He was a Research Associate in the Department of Computer Science at University College London (UCL), as part of an Innovate UK Knowledge Transfer Partnership between UCL and a data analytics start-up. He subsequently joined the start-up as a Data Scientist. He holds a PhD degree from the Department of Computer Science at the University of York, UK with a focus on multiagent systems and reinforcement learning. His PhD was funded by an EPSRC DTA Scholarship. He obtained his MEng degree in Computer Systems and Software Engineering from the same department.
\end{IEEEbiography}

\begin{IEEEbiography}[{\includegraphics[width=1in,height=1.25in,clip,keepaspectratio]{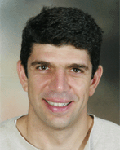}}]%
	{Christos G. Panayiotou} has received a B.Sc. and a Ph.D. degree in Electrical and Computer Engineering from the University of Massachusetts at Amherst, in 1994 and 1999 respectively. He also received an MBA from the Isenberg School of Management, at the aforementioned university in 1999. Currently he is a Professor at the Electrical and Computer Engineering Department at UCY and he serves as the Deputy Director of the KIOS CoE for which he is also a founding member. His research interests include distributed control systems, wireless, ad hoc and sensor networks, computer communication networks, quality of service (QoS) provisioning, optimization and control of discrete-event systems, resource allocation, simulation, transportation networks and manufacturing systems. He is a senior member of the IEEE and also a reviewer for various conferences and journals, and he has served in the organizing and program committees of various international conferences.
\end{IEEEbiography}

\begin{IEEEbiography}[{\includegraphics[width=1in,height=1.25in,clip,keepaspectratio]{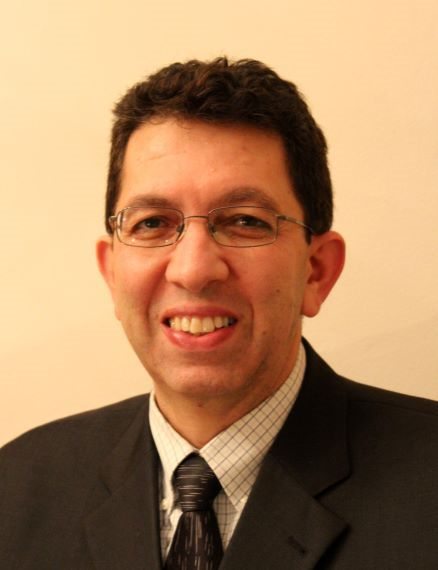}}]%
	{Marios M. Polycarpou} is a Professor of Electrical and Computer Engineering and the Director of the KIOS Research and Innovation Center of Excellence at the University of Cyprus. He received the B.A degree in Computer Science and the B.Sc. in Electrical Engineering, both from Rice University, USA in 1987, and the M.S. and Ph.D. degrees in Electrical Engineering from the University of Southern California, in 1989 and 1992 respectively. His teaching and research interests are in intelligent systems and networks, adaptive and cooperative control systems, computational intelligence, fault diagnosis and distributed agents. He has published more than 300 articles in refereed journals, edited books and refereed conference proceedings, and co-authored 7 books. He is also the holder of 6 patents. Prof. Polycarpou is a Fellow of IEEE and IFAC. He is the recipient of the 2016 IEEE Neural Networks Pioneer Award. He received with his co-authors the 2014 Best Paper Award for the journal Building and Environment (Elsevier). Prof. Polycarpou served as the President of the IEEE Computational Intelligence Society (2012-2013), and as the Editor-in-Chief of the IEEE Transactions on Neural Networks and Learning Systems (2004-2010). He is currently the President of the European Control Association (EUCA). He has participated in more than 60 research projects/grants, funded by several agencies and industry in Europe and the United States, including the prestigious European Research Council (ERC) Advanced Grant.
\end{IEEEbiography}


%\vfill


\end{document}